\documentclass[letterpaper]{article} 
\usepackage{aaai2027}  
\usepackage[hyphens]{url}  
\usepackage{graphicx} 
\usepackage{natbib}  
\usepackage{caption} 
\usepackage{amsmath}
\usepackage{algorithm}
\usepackage{algorithmic}

\usepackage{newfloat}
\usepackage{listings}
\DeclareCaptionStyle{ruled}{labelfont=normalfont,labelsep=colon,strut=off} 
\lstdefinelanguage{text}{}
\lstdefinelanguage{json}{}
\floatstyle{ruled}
\newfloat{listing}{tb}{lst}{}
\floatname{listing}{Listing}

\definecolor{promptbg}{RGB}{247,248,250}
\definecolor{promptrule}{RGB}{176,181,188}
\lstdefinestyle{prompt}{
  basicstyle={\fontsize{7.5}{9.2}\selectfont\ttfamily},
  showstringspaces=false,
  breaklines=true,
  breakatwhitespace=false,
  columns=fullflexible,
  keepspaces=true,
  backgroundcolor=\color{promptfill},
  frame=none,
  framesep=0pt,
  xleftmargin=0pt,
  xrightmargin=0pt,
  numbers=none,
  aboveskip=0pt,
  belowskip=0pt
}

\usepackage{booktabs}
\usepackage{tabularx}
\usepackage{array}
\usepackage{needspace}
\usepackage[fencedCode,codeSpans,underscores=false]{markdown}

\usepackage[most]{tcolorbox}
\newcommand{\promptmdheading}[1]{\par\medskip\noindent\textbf{#1}\par\smallskip}
\newcommand{\promptfencedcode}[2]{\lstinputlisting[style=prompt,language=#2]{#1}}
\newcommand{\promptverbatiminput}[1]{\lstinputlisting[style=prompt]{#1}}
\markdownSetup{
  tightLists=false,
  rendererPrototypes={
    headingOne={\promptmdheading{#1}},
    headingTwo={\promptmdheading{#1}},
    headingThree={\promptmdheading{#1}},
    headingFour={\promptmdheading{#1}},
    headingFive={\promptmdheading{#1}},
    headingSix={\promptmdheading{#1}}
  },
  renderers={
    inputFencedCode={\promptfencedcode{#1}{#2}},
    inputVerbatim={\promptverbatiminput{#1}}
  }
}
\definecolor{prompttitle}{HTML}{5B86B3}
\definecolor{promptframe}{HTML}{A8C7E3}
\definecolor{promptfill}{HTML}{F2F8FD}

\newenvironment{promptbox}[1]{%
  \begin{tcolorbox}[
    colback=promptfill,
    colframe=promptframe,
    boxrule=0.4pt,
    arc=2pt,
    left=6pt,
    right=6pt,
    top=6pt,
    bottom=6pt,
    title=#1,
    fonttitle=\bfseries\small,
    coltitle=white,
    colbacktitle=prompttitle,
    boxed title style={arc=2pt, boxrule=0pt},
    before skip=6pt,
    after skip=8pt,
    breakable]
  \small\raggedright\sloppy\setlength{\parskip}{2pt}
}{%
  \end{tcolorbox}
}
\newenvironment{promptlisting}[1]{%
  \begin{tcolorbox}[
    colback=promptfill,
    colframe=promptframe,
    boxrule=0.4pt,
    arc=2pt,
    left=6pt,
    right=6pt,
    top=6pt,
    bottom=6pt,
    title=#1,
    fonttitle=\bfseries\small,
    coltitle=white,
    colbacktitle=prompttitle,
    boxed title style={arc=2pt, boxrule=0pt},
    before skip=6pt,
    after skip=8pt,
    breakable]
  \small\raggedright\sloppy\setlength{\parskip}{2pt}
}{%
  \end{tcolorbox}
}

\newcolumntype{Y}{>{\raggedright\arraybackslash}X}
\newcolumntype{L}[1]{>{\raggedright\arraybackslash}p{#1}}

\usepackage{xspace}
\usepackage{tikz}
\usepackage{pgfplots}
\pgfplotsset{compat=1.18}
\definecolor{mfblue}{HTML}{1F5FBF}
\definecolor{mfgreen}{HTML}{139B6E}
\definecolor{mforange}{HTML}{EF8A18}
\definecolor{mfpurple}{HTML}{6652B8}
\definecolor{mfred}{HTML}{D94A3A}
\definecolor{mfcyan}{HTML}{2B9CCF}
\definecolor{mfgray}{HTML}{7A8798}

\title{MemFuse: Multi-Source Memory Fusion from Fragmented Observations}
\nocopyright
\author{
    Chao Li\textsuperscript{\rm 1},
    Yuanfa Li\textsuperscript{\rm 1},
    Wenhao Wu\textsuperscript{\rm 2}\thanks{Work done during internship.},
    Xule Liu\textsuperscript{\rm 1},
    Zhi Wang\textsuperscript{\rm 2},
    Kun Shao\textsuperscript{\rm 1}\corresponding
}
\affiliations{
    \textsuperscript{\rm 1}Xiaomi Inc.\\
    \textsuperscript{\rm 2}Nanjing University\\
    lichao75@xiaomi.com, liyuanfa@xiaomi.com, wenhaowu@smail.nju.edu.cn,\\
    liuxule@xiaomi.com, zhiwang@nju.edu.cn, shaokun@xiaomi.com
}

\newcommand{\MemFuse}{MemFuse\xspace}
\newcommand{\MemFuseBench}{MemFuseBench\xspace}

\newcommand{\fusednode}{\texttt{FusedNode}\xspace}

\begin{document}

\maketitle

\begin{abstract}
Long-term memory is essential for agents that operate across extended interactions, yet existing memory systems and benchmarks predominantly focus on single-source textual histories. In realistic settings, however, relevant information is often fragmented across applications and devices, as well as across users and time, requiring agents to integrate dispersed observations into coherent episodic memories while preserving their source provenance. To address these gaps, we introduce \textbf{\MemFuseBench}, a benchmark for \emph{multi-source memory fusion}. \MemFuseBench is built with a Scene-to-Sensor pipeline that synthesizes controllable scenarios into source-tagged observations, evidence-grounded questions, and adversarial distractors. It enables systematic evaluation of temporal reasoning, cross-source evidence fusion, and robustness to noise. We further propose \textbf{\MemFuse}, a structured memory system that preserves source-level evidence in event-layer atomic memory and organizes related atomic events into cluster-layer fused memory within a causal fusion graph. During retrieval, \MemFuse retrieves and organizes related evidence fragments while maintaining traceability to original source events. Experiments on \MemFuseBench show that \MemFuse achieves the best overall performance among the evaluated memory systems under all three LLM settings and consistently improves performance on questions requiring cross-source evidence fusion.
\end{abstract}

\noindent\textbf{Code} --- \url{https://github.com/Darwin-Agent/Mi-Memory/tree/master/MemFuse}

\section{Introduction}

Long-term memory systems have received growing attention as agents are expected to maintain useful context across extended interactions. Existing systems store past interactions as records, summaries, or structured memories for retrieval \citep{memgpt2023,mem0,amem2025,agentmemoryage2025,evermemos2026,higmem2026,hmem2026,hu2026xmemory}, and work well when relevant context forms a coherent history. As user expectations grow, however, agents need to remember not only what users explicitly tell them, but also useful observations from devices, applications, and other users. In this setting, the same underlying episode may then be represented by fragmented events from different origins, making it necessary to integrate complementary observations without losing their sources. We call this memory-level problem \emph{multi-source memory fusion}: retrieving and integrating distributed semantic events while preserving event-layer provenance.

Existing memory benchmarks mainly evaluate conversational recall, temporal updates, or long-context reasoning over interaction histories \citep{locomo2024,longmemeval2025,membench2025,evermembench2026,groupmembench2026}. Recent benchmarks consider heterogeneous digital traces and multimodal evidence \citep{lifebench2026,smmbench2026}, but they do not specifically test whether memory systems can link fragmented, source-tagged events into traceable evidence for fusion-oriented questions. It therefore remains difficult to test whether a system can recover complementary observations from different origins.

\begin{figure}[t]
\centering
\includegraphics[width=0.95\columnwidth]{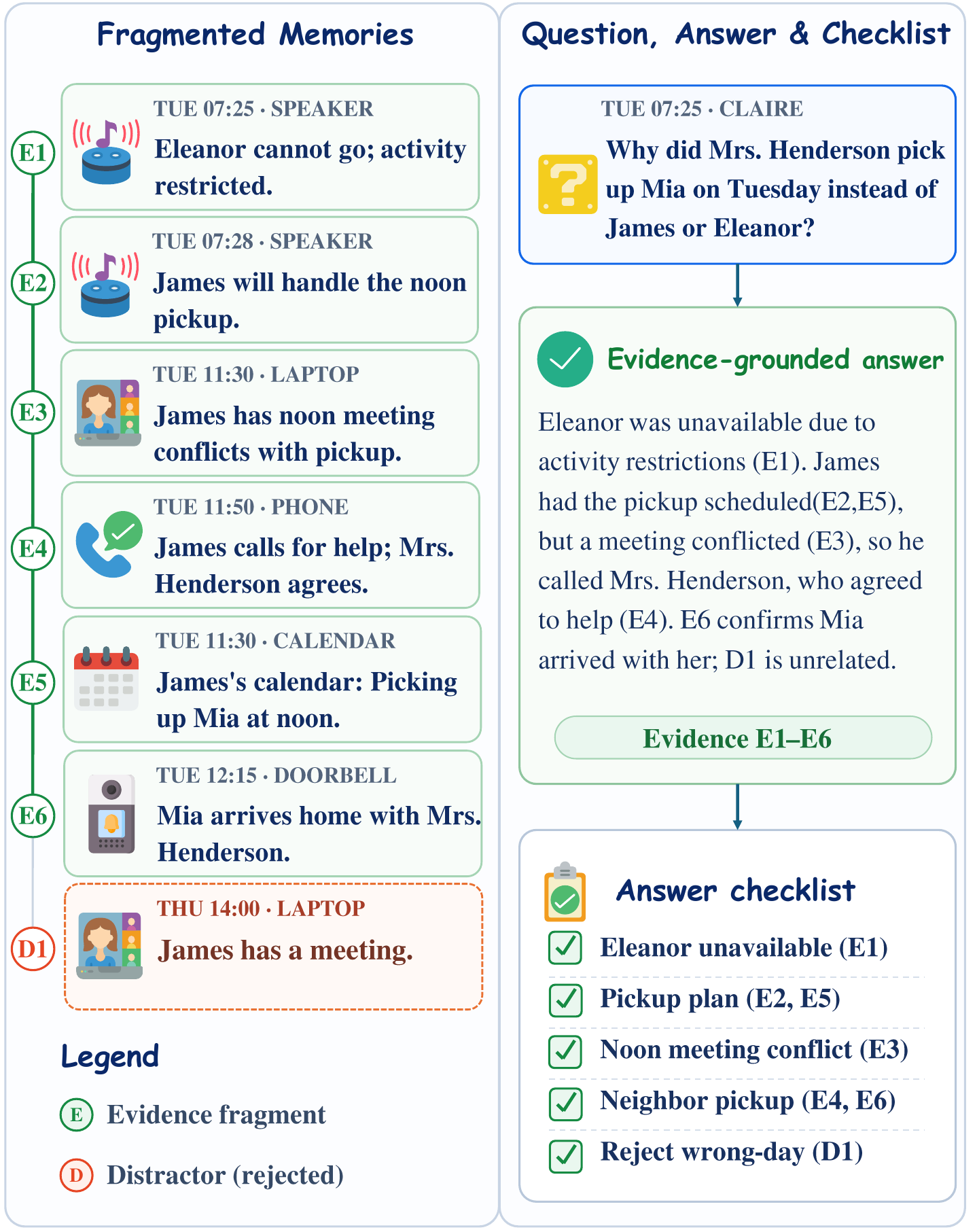}
\caption{A \MemFuseBench instance requiring fusion across a household conversation, work calendar, and phone call, while rejecting a plausible distractor on the wrong day.}
\label{fig:intro-example}
\end{figure}

To address these challenges, we first introduce \textbf{\MemFuseBench}, a benchmark for multi-source memory fusion. Starting from controllable scenarios, its Scene-to-Sensor pipeline generates source-tagged observations, evidence-grounded questions, and adversarial distractors. Figure~\ref{fig:intro-example} illustrates a question requiring evidence from multiple origins while ignoring a plausible distractor. The benchmark contains 357 questions over 7,823 events across six diagnostic categories.

To further support multi-source memory fusion, we propose \textbf{\MemFuse}, a graph-structured memory system that preserves event-layer memory as the evidence layer and organizes related events into cluster-layer memory within a causal fusion graph. At retrieval time, \MemFuse uses agentic search to recover both compact cluster memories and their traceable source events.

In summary, our contributions are: (1) identifying multi-source memory fusion as a memory-level research problem over fragmented, source-tagged events; (2) introducing \MemFuseBench, a fusion-centric benchmark for this setting, featuring six diagnostic categories and answer checklists, supported by a scalable LLM-based synthesis and reviewer-corrector validation pipeline; (3) proposing \MemFuse, a structured memory system that preserves source-level evidence in event-layer atomic memory and fuses related events into cluster-layer memory within a causal fusion graph; and (4) validating \MemFuseBench and \MemFuse through experiments across multiple models, where \MemFuse achieves the best Overall score among the evaluated retrieval and memory systems under all three LLM settings.

\section{Related Work}
\paragraph{Long-Term Agent Memory.}
Agent-memory systems store past interactions as records, summaries, linked notes, or hierarchical memory units for later retrieval \citep{generativeagents2023,memgpt2023,memorybank2024,thinkinmemory2023,longmem2023,mem0,amem2025,llmagentsurvey2025,agentmemoryage2025,evermemos2026,higmem2026,hmem2026}. Recent work also introduces structured or graph-based memory to connect related experiences and support multi-step retrieval \citep{compassmem2026,structmem2026,graphmemorysurvey2026,hu2026xmemory}. These systems improve organization over flat logs, but they mainly assume conversational or interaction histories rather than fragmented observations across different sources.

\paragraph{Memory Benchmarks.}
LoCoMo, LongMemEval, and MemBench evaluate conversational recall, temporal updates, abstention, and long-context reasoning \citep{locomo2024,longmemeval2025,membench2025}. EverMemBench and GroupMemBench extend this line to long-term interactive memory and multi-party conversations \citep{evermembench2026,groupmembench2026}. LifeBench and SMMBench move closer to realistic settings by using heterogeneous digital traces or independently originated multimodal evidence \citep{lifebench2026,smmbench2026}. They broaden the input sources for memory evaluation, but do not center the task design on linking, fusing, and tracing distributed event fragments.

\paragraph{Multi-Source memory Fusion.}
Event-centric resources and lifelogging systems organize heterogeneous records into event structures for temporal or commonsense reasoning \citep{eventkg2018,aser2020,lifelogging2014}. Multisensor fusion studies how signals or decisions from multiple sensors are registered, estimated, or combined \citep{multisensorfusion2013}. These research directions are related to multi-source evidence, but they are not concerned with agent memory: none addresses how a long-term memory system should preserve atomic provenance while grouping and retrieving fragmented observations for downstream question answering.

\section{MemFuseBench}
Most memory benchmarks evaluate recall within a single conversational history.
Recent ones broaden the input to heterogeneous or multimodal traces, but none isolates the core difficulty of reasoning over evidence fragmented across devices, applications, users, and time, where no single record suffices.
Evaluating this demands two properties existing benchmarks lack: source-level evidence annotations that expose provenance, and questions whose answers genuinely depend on stitching fragments from more than one source.
We therefore build \MemFuseBench, pairing source-tagged event streams with evidence-grounded questions answerable only by fusing fragmented atomic events across sources, alongside adversarial distractors that penalize shortcut retrieval.

\subsection{Dataset Construction}
Each \MemFuseBench instance contains source-specific event streams and a question with its reference answer and an answer checklist. We construct these instances with a Scene-to-Sensor pipeline that turns controllable latent scenarios into source-specific observations, evidence-grounded questions, and adversarial non-evidence events. Reviewer-corrector validation checks and repairs intermediate artifacts for consistency, evidence support, and answer preservation under injected distractors.

\begin{figure*}[t]
\centering
\includegraphics[width=0.9\textwidth]{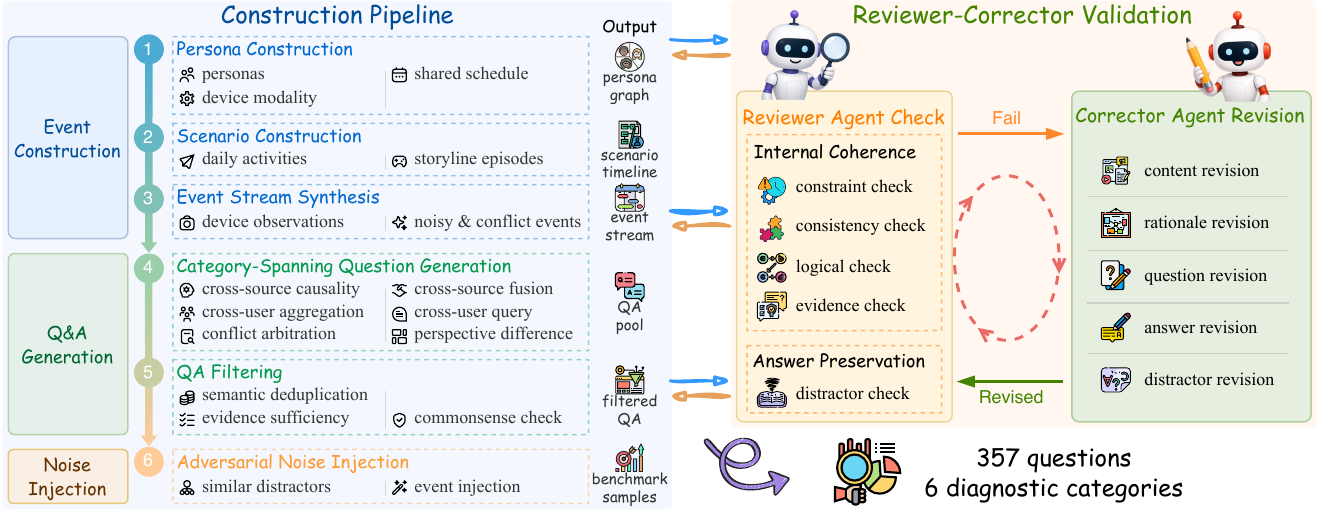}
\caption{Overall \MemFuseBench construction framework. The left panel summarizes the \textbf{construction pipeline}, whose six stages are organized into three phases: event construction, QA generation, and noise injection. The right panel shows \textbf{reviewer-corrector validation}, where candidate samples are checked, revised, and iterated until the final accepted sample is produced.}
\label{fig:memfusebench-pipeline}
\end{figure*}

\paragraph{Construction Pipeline.}
The core construction method is \textbf{Scene-to-Sensor} generation, which builds each instance by top-down narrowing rather than directly generating question-answer pairs.
It first fixes a high-level ground truth---stable personas and a causally linked scenario timeline (\emph{what happens in the world})---then projects each latent activity into concrete, timestamped, source-specific event streams (\emph{what each device observes}), from which the evidence-grounded questions are derived.
Figure~\ref{fig:memfusebench-pipeline} summarizes the framework, which is organized into six stages:

\begin{enumerate}
    \item \textbf{Persona Construction.} Sample stable personas and their shared context---schedules, relationships, and multi-device inventories with observable modalities---into a persona-source graph, which constrains all later stages.
    \item \textbf{Scenario Construction.} Organize the personas' daily activities into causally linked storyline events grouped by episode. This timeline is the latent ground truth, later refracted into source-specific events.
    \item \textbf{Event Stream Synthesis.} Render the scenario into source-specific events, interleaving routine, periodic, noise, and conflict events so each source gives only a partial view. This yields a timestamped event stream.
    \item \textbf{Category-Spanning Question Generation.} Derive questions spanning the six diagnostic categories in Table~\ref{tab:benchmark-stats}, producing a QA pool of questions, reference answers, answer checklists.
    \item \textbf{QA Filtering.} Filter the pool to remove semantic duplicates, commonsense shortcuts, and items whose labeled evidence cannot support the reference answer, retaining only answerable, evidence-supported, non-duplicative items.
    \item \textbf{Adversarial Noise Injection.} Inject semantically similar distractors to complicate retrieval while keeping the gold evidence set auditable and unchanged, so finalized samples contain both gold evidence and plausible non-evidence distractors.
\end{enumerate}

\paragraph{Validation Protocol.}
As the pipeline builds each instance stage by stage in a top-down manner, an inconsistent persona-source graph can propagate errors to every later artifact.
To prevent this, \MemFuseBench applies \textbf{reviewer-corrector validation} to every intermediate artifact before it enters the next stage: a \emph{reviewer} flags structural, semantic, and consistency issues with stage-specific prompts, a \emph{corrector} repairs them,  and this review-correction process iterates until the reviewer finds no remaining issues.

The criteria are stage-specific and focus on two targets.
For the first four stages, validation targets \emph{internal coherence}: the persona-source graph, scenario timeline, event stream, and QA pool must each be self-consistent and aligned with the artifacts upstream.
For Adversarial Noise Injection, validation instead targets \emph{answer preservation}: injected distractors must complicate retrieval without altering the gold evidence set or reference answers.

We further conduct a model-guided verification pass: given the full context, GPT-5.5, Claude Opus 4.6, and Gemini 3.1 Pro each independently answer and evaluate every question, and the samples on which they disagree are manually inspected and corrected. This procedure results in at least one revision for approximately 20\% of the 357 QA instances.

\subsection{Dataset Analysis}
This pipeline yields \MemFuseBench with six scenario datasets, each paired with evidence-grounded questions spanning six diagnostic categories. On average, a scenario contains 1,303.8 atomic events and 110.6k tokens under the Gemini 3.1 Flash Lite tokenizer. Table~\ref{tab:benchmark-stats} summarizes the resulting 357 questions; each question requires evidence from 9.4 distinct events on average.

\begin{table*}[t]
\centering
\small
\setlength{\tabcolsep}{6pt}
\begin{tabular}{lrrp{2.5in}}
\toprule
Category & \#Q & Avg. Events & Description \\
\midrule
Cross-source causality (Causal) & 63 & 8.5 & Infer causal relations across source observations. \\
Cross-source fusion (Fusion) & 71 & 15.2 & Combine partial observations across sources. \\
Cross-user aggregation (User Agg.) & 52 & 11.4 & Aggregate evidence across users. \\
Cross-user query (User Query) & 61 & 7.1 & Answer queries about another user's events. \\
Conflict arbitration (Conflict) & 70 & 2.8 & Resolve conflicting source reports. \\
Perspective difference (Perspective) & 40 & 13.3 & Account for different user or source perspectives. \\
\midrule
Total & 357 & 9.4 & Full benchmark release. \\
\bottomrule
\end{tabular}
\caption{Per-category statistics of the current \MemFuseBench release. Short labels in parentheses are used in result tables. \#Q is the number of questions; Avg. Events is the average number of distinct evidence events required per question.}
\label{tab:benchmark-stats}
\end{table*}

\section{MemFuse}
\begin{figure*}[t]
\centering
\includegraphics[width=0.9\textwidth]{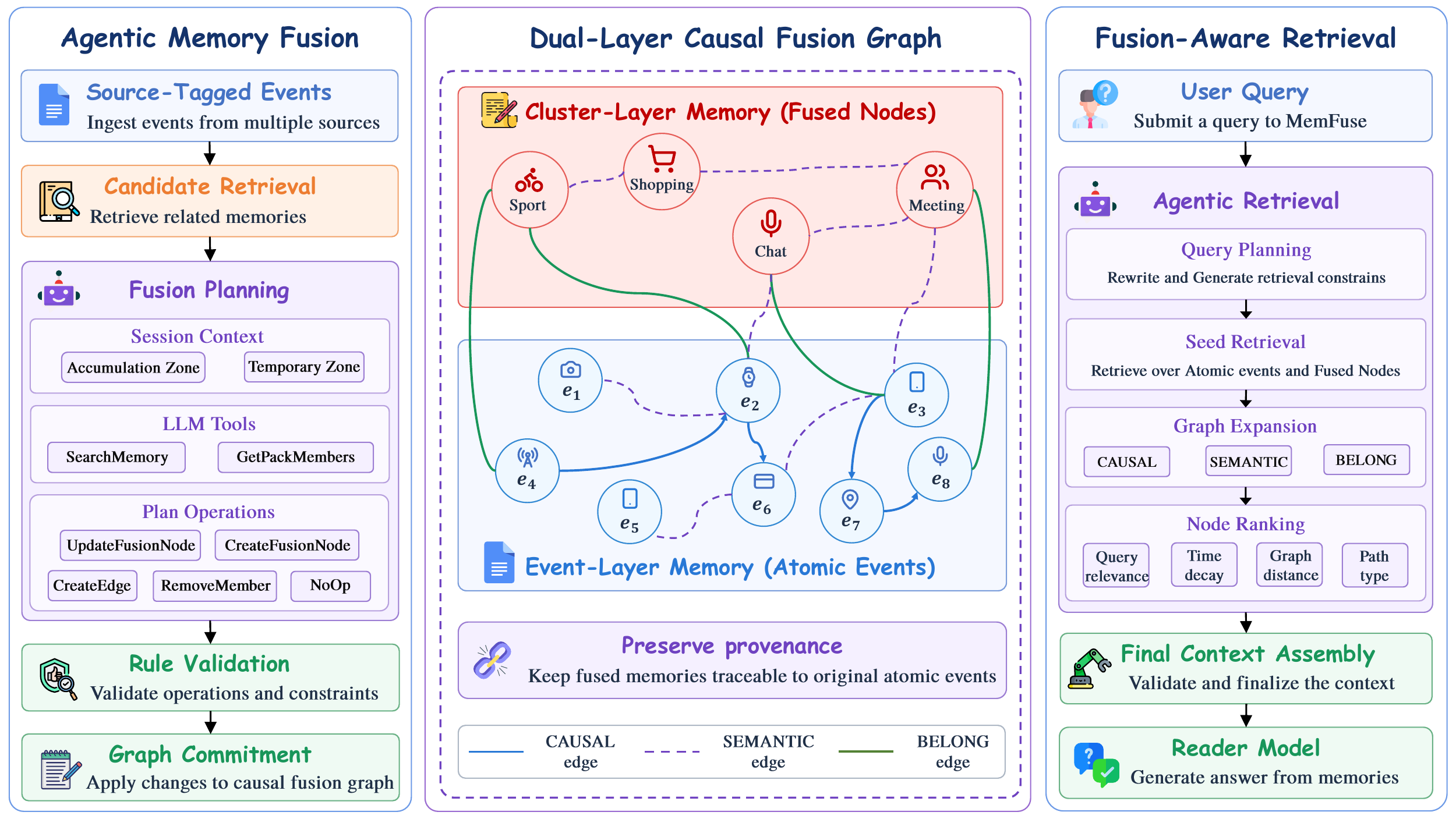}
\caption{Overall \MemFuse framework. Left: online construction from source-tagged events through agentic fusion. Middle: a dual-layer causal fusion graph connecting atomic events and fused memories. Right: fusion-aware retrieval that returns traceable event-layer evidence for grounded answering.}
\label{fig:memfuse-framework}

\end{figure*}

We target more realistic settings where relevant evidence is fragmented across devices, applications, users, and time, requiring a memory system that preserves the provenance of each observation while grouping and reasoning over related fragments.
\MemFuse separates provenance preservation from memory aggregation: event-layer atomic memory preserves source evidence, cluster-layer fused memory summarizes related evidence, and the causal fusion graph connects the two memory layers.
Built on this structure, \MemFuse stores incoming atomic events, groups related events into fused memory, proposes validated fusion operations to construct the graph, and performs fusion-aware retrieval for evidence-grounded answer generation. Figure~\ref{fig:memfuse-framework} gives the overall architecture.

\subsection{Preliminaries}
Given a stream of normalized, source-tagged atomic events $\mathcal{E}=\{e_i\}_{i=1}^{n}$, \MemFuse organizes memory in two layers:
\begin{itemize}
\item \textbf{Event-layer atomic memory} $\mathcal{M}_E$: each event is stored as an immutable and indexed memory that retains its provenance for later grounding.
\item \textbf{Cluster-layer fused memory} $\mathcal{M}_V$: related events are grouped into compact retrieval units, each a \fusednode $v\in\mathcal{V}$ defined by a member set $\mu(v)\subseteq\mathcal{E}$, a fused summary $y_v$ as retrieval entry point, and back-pointers that keep $v$ grounded in its source events.
\end{itemize}
The overall memory state is $\mathcal{M}=\{\mathcal{M}_E,\mathcal{M}_V,\mathcal{G}\}$, where $\mathcal{G}$ is the causal fusion graph connecting the two layers. Given a query $q$, \MemFuse retrieves a relevant context $\mathcal{C}_q$ from $\mathcal{M}$, and the task is formalized as $a = \mathrm{Assistant}(\mathcal{C}_q, q)$.

\subsection{Agentic Memory Fusion}
\label{sec:agentic_memory_fusion}
\MemFuse builds the dual-layer memory online through an agentic fusion pipeline with four stages: \emph{candidate retrieval}, \emph{fusion planning}, \emph{rule validation}, and \emph{graph commitment}. For each incoming event, \MemFuse stores it in the atomic memory, retrieves related atomic events and fused memories as candidates, lets a fusion agent gather further evidence and propose how to fuse the event, and validates and commits the accepted operations to the dual-layer causal fusion graph described next. We describe each stage below.

\paragraph{Candidate Retrieval.}
Given an incoming atomic event $e_i$, \MemFuse retrieves a candidate set $\mathcal{C}_i=\mathcal{C}^{\mathrm{atom}}_i\cup\mathcal{C}^{\mathrm{fused}}_i$ of potentially related atomic events and fused memories. This bounds the initial decision scope to a small set, avoiding a full memory scan while still exposing complementary evidence across sources.

\paragraph{Agentic Fusion Planning.}
A fusion agent then decides how to fuse $e_i$ in two steps: it first gathers related evidence through agentic information-seeking, and then proposes the corresponding fusion operations. To provide sufficient context for fusion decisions, \MemFuse maintains a \emph{session context} $\mathcal{S}_i=(\mathcal{Z}^{\mathrm{acc}}_i,\mathcal{Z}^{\mathrm{tmp}}_i)$ with two complementary memory zones: the accumulation zone $\mathcal{Z}^{\mathrm{acc}}_i$, a sliding window over previously processed events, and the temporary zone $\mathcal{Z}^{\mathrm{tmp}}_i$, containing the current event $e_i$ and its candidate set $\mathcal{C}_i$. Together, $\mathcal{S}_i$ provides compact access to prior context and local evidence.

Conditioned on $\mathcal{S}_i$, the fusion agent performs a bounded information-seeking trajectory, represented by the tool-call history
\[
\tau_i^{(t)}
=
\left(
(u_i^{(1)}, r_i^{(1)}),\ldots,(u_i^{(t)}, r_i^{(t)})
\right),
\qquad
u_i^{(t)} \in \mathcal{U}_{\mathrm{tools}},
\]
where $u_i^{(t)}$ is the selected tool and $r_i^{(t)}$ its returned result. The tool set $\mathcal{U}_{\mathrm{tools}}$ contains \textsc{SearchMemory}, which expands the candidate set with newly retrieved atomic events and fused memories, and \textsc{GetPackMembers}, which returns the member atomic events of a candidate fused memory. Starting from $\tau_i^{(0)}=\emptyset$, at each turn the agent picks the next tool $u_i^{(t)}=\mathrm{LLM}_{\theta}(\mathcal{S}_i,\tau_i^{(t-1)})$ and executes it to obtain $r_i^{(t)}=\mathrm{Exec}(u_i^{(t)})$, iteratively accumulating evidence across sources.

Once sufficient evidence has been gathered, the agent proposes how $e_i$ should be fused into the cluster-layer memory by terminating the trajectory with a call to \textsc{SubmitFusionPlan}, yielding the fusion plan $\mathcal{P}_i$:
\[
\mathcal{P}_i
=
\textsc{SubmitFusionPlan}
\left(
\mathcal{S}_i,\tau_i^{(J_i)}
\right)
=
\left(
o_i^{(1)},\ldots,o_i^{(L_i)}
\right),
\]
where $J_i$ is the number of tool turns before submission, $L_i$ the number of operations, and each $o_i \in \mathcal{O}$. The operation set $\mathcal{O}$ comprises \textsc{CreateEdge}, \textsc{CreateFusionNode}, \textsc{UpdateFusionNode}, \textsc{RemoveMember}, and \textsc{NoOp}, which respectively create an edge, create or update a fused node, remove a member from fused node, or keep the event atomic.

\paragraph{Rule Validation and Graph Commitment.}
Before the proposed operations are applied, a rule-based validator checks the consistency and validity of each operation against the memory storage structure. If any constraint is violated, the whole plan is rejected rather than partially committed, preventing malformed outputs from corrupting persistent memory. The accepted operations are then applied to the dual-layer causal fusion graph, materializing the fusion of $e_i$ into the cluster-layer memory, which we detail next.

\subsection{Dual-Layer Causal Fusion Graph}
\label{sec:causal_fusion_graph}

Beyond grouping related atomic events into \fusednode objects, \MemFuse connects memories with typed relations that support cross-source reasoning. These relations form a dual-layer causal fusion graph, which serves as the persistent structure for committing validated fusion operations. We first define the graph and its relation types, then describe the commitment steps.

\paragraph{Graph Definition.}
With $\mathcal{N}=\mathcal{E}\,\dot{\cup}\,\mathcal{V}$ the disjoint union of atomic-event nodes $\mathcal{E}$ and \fusednode objects $\mathcal{V}$, the graph is defined as
$
\mathcal{G}
=
\left(
\mathcal{N},
\mathcal{R}_{\textsc{Belong}},
\mathcal{R}_{\textsc{Causal}},
\mathcal{R}_{\textsc{Semantic}}
\right),
$
with $\mathcal{R}_{\textsc{Belong}} \subseteq \mathcal{E}\times\mathcal{V}$, $\mathcal{R}_{\textsc{Causal}} \subseteq \mathcal{E}\times\mathcal{E}$, and $\mathcal{R}_{\textsc{Semantic}} \subseteq \mathcal{N}\times\mathcal{N}$.

\paragraph{Typed Relations.}
A \textsc{Belong} edge $(e,v)$ marks event $e$ as a member of \fusednode $v$, giving the member set $\mu(v)=\{e\in\mathcal{E} | (e,v)\in\mathcal{R}_{\textsc{Belong}}\}$. Fused summaries thus serve only as retrieval abstractions, while answers stay grounded in the source-tagged events recovered through $\mu(v)$. A directed \textsc{Causal} edge $(e_i,e_j)$ links atomic events in causal order, stored with direction but traversable either way. A \textsc{Semantic} edge connects semantically related nodes---atomic, fused, or both---and carries a similarity score $\omega_S(x_i,x_j)\in[0,1]$ for thresholded retrieval.

\paragraph{Graph Construction.}
Committing the accepted operations $\mathcal{P}_i$ materializes these edges: newly created or updated \fusednode objects induce their \textsc{Belong} edges by membership and are embedded and indexed for retrieval; \textsc{Causal} edges from \textsc{CreateEdge} operations are committed after endpoint and type validation; and a \textsc{Semantic} edge is added online whenever an inserted node's embedding similarity to an existing node exceeds $\rho_S$ ($0.8$ in our experiments). These typed relations form the traversal structure used by fusion-aware retrieval.

\subsection{Fusion-Aware Retrieval}
\label{sec:fusion_aware_retrieval}

Given a query, \MemFuse returns a memory set through an agentic retrieval loop. At each step, a retrieval agent issues a search call that returns a top-$k$ set of event candidates, and it keeps searching and accumulating until it judges the collected evidence sufficient. It then selects the final top-$k$ events from the accumulated candidates to form the returned context $\mathcal{C}_q$. The agentic retrieval loop combines query planning, seed retrieval, graph expansion, ranking, and assembly, which we describe below.

\paragraph{Query Planning and Seed Retrieval.}
Given a query $q$ with timestamp $t_q$ and requester identity $u_q$, \MemFuse issues a search call with query-planning parameters $p_q$, which include a rewritten query and optional retrieval constraints such as temporal or source filters. Conditioned on $p_q$, \MemFuse performs dense (vector) and sparse (BM25) retrieval over the shared index of atomic events and fused memories, plus temporal retrieval when needed. The ranked lists are combined by reciprocal rank fusion into a seed score $r_{\mathrm{seed}}(x |  q,p_q)$, and the seed set $\mathcal{S}_q$ keeps the top-$K_{\mathrm{seed}}$ candidates with positive scores.

\paragraph{Typed Graph Expansion.}
Since direct retrieval may recover only one fragment of the required evidence, \MemFuse expands the seed nodes into a larger candidate set $\mathcal{X}_q$ through the \textsc{Belong}, \textsc{Causal}, and \textsc{Semantic} relations of the causal fusion graph: $\mathcal{X}_q=\operatorname{Expand}_{\mathcal{G}}\left(\mathcal{S}_q;p_q\right)$, with a policy that depends on the seed type.


For an atomic seed, \MemFuse traverses nearby bidirectional \textsc{Causal} edges and high-confidence \textsc{Semantic} edges, and follows \textsc{Belong} edges from the seed or an atomic neighbor within one causal hop to expose the corresponding member events. For a fused seed, \MemFuse first follows reverse \textsc{Belong} edges to its member events and then applies the same bounded causal and semantic expansion from those members. In both cases, traversal is capped by the hop limit, the semantic threshold, and a budget for fused-node expansion.

For each candidate $x\in\mathcal{X}_q$, \MemFuse retains an expansion trace $\xi_q(x)=(s_x,\pi_x,d_x)$ recording the originating seed $s_x\in\mathcal{S}_q$, the typed relation path $\pi_x$, and its distance $d_x$.

\paragraph{Candidate Ranking and Evidence Construction.}
The expanded candidates are reranked using their query relevance, temporal consistency, and graph-expansion traces:
\[
\widehat{\mathcal{X}}_q
=
\operatorname{TopK}_{K_{\mathrm{rank}}}
\left(
\mathcal{X}_q;
s(x |  q,p_q,\xi_q(x))
\right).
\]
The final score $s(x | q,p_q,\xi_q(x))$ combines cosine similarity to the query, time decay, graph-hop decay, the RRF seed score when available, a path-type prior from $\xi_q(x)$, and a date-match boost. As a result, at comparable relevance, candidates reached through short membership or causal paths are favored over distant semantic neighbors.

Since a fused node serves as a retrieval and expansion unit, \MemFuse projects each ranked candidate to atomic events via $\operatorname{Ev}(x)$: the event itself if $x\in\mathcal{E}$, or its member set $\mu(x)$ if $x\in\mathcal{V}$. It then deduplicates and truncates the result to the top-$k$ events for the current search.

\paragraph{Final Context Assembly.}
Once the agentic loop terminates, \MemFuse selects the final top-$k$ events from the accumulated search results, forming the context $\mathcal{C}_q$. To keep this budget fixed, \MemFuse backfills any unfilled slots from the earlier search history and truncates any excess to the final top-$k$ events. The resulting context $\mathcal{C}_q$ is finally sent to the reader model, which generates the answer.

\section{Experiments}
Our experiments characterize performance on fragmented, source-tagged event streams in \MemFuseBench and evaluate \MemFuse as an end-to-end memory system. We ask four questions: (i) how well existing systems handle fragmented multi-source memory under top-$k$ access relative to the long context reference, (ii) how \MemFuse compares with retrieval and memory baselines, (iii) which diagnostic categories expose the largest gaps, and (iv) how removing each \MemFuse component affects answer quality.

\subsection{Experiment Setup}
\subsubsection{Baselines.}
We compare \MemFuse against Long context prompting, naive RAG, and three existing memory systems---Mem0~\citep{mem0}, A-MEM~\citep{amem2025}, and EverMemOS~\citep{evermemos2026}.

\subsubsection{Implementation.}
We evaluate each system under three LLM settings: Qwen3-30B-A3B, GPT-4.1 Mini, and Gemini 3.1 Flash Lite, using the same LLM within each setting for method-specific LLM calls and answer generation. All systems ingest the same event stream and question set. Long context prompting ingests the entire event stream. Naive RAG retrieves nearest-neighbor events by embedding similarity. Mem0, A-MEM, and EverMemOS use their native memory interfaces to ingest the same event stream and return candidate memories. We use BGE-M3 embeddings and compare top-$k$ systems under the same top-$20$ item budget. For \MemFuse, fused nodes expand candidate discovery; the resulting candidates are projected to atomic events, deduplicated, and truncated to the final 20-event context.

\subsubsection{Evaluation Metrics.}
We measure answer quality with an LLM-as-judge checklist score, using GPT-4.1 Mini as the judge for all evaluations.
The judge marks each answer checklist item as covered or not, and the score is the fraction covered; the full prompt is provided in the supplement. We report an Overall score as the mean across all questions, together with per-category scores averaged within each diagnostic category in Table~\ref{tab:benchmark-stats}.

\subsection{Main Results}
Table~\ref{tab:main-results} reports answer checklist scores and token usage for all systems across the three LLM settings.
Based on these results, we aim to answer the following questions:

\begin{table*}[t]
\centering
\scriptsize
\renewcommand{\arraystretch}{0.88}
\setlength{\tabcolsep}{8pt}
\begin{tabular}{@{}llccccccccc@{}}
\toprule
\multicolumn{2}{c}{} & \multicolumn{7}{c}{Answer Checklist Score} & \multicolumn{2}{c}{Token Usage (M)} \\
\cmidrule(lr){3-9}\cmidrule(lr){10-11}
System & Setting & Overall & Causal & Fusion & User Agg. & User Query & Conflict & Perspective & Ingest & Inference \\
\midrule
\multicolumn{11}{c}{\textit{Qwen3-30B-A3B}} \\
Long context & all & 0.4424 & 0.4521 & 0.3737 & 0.3575 & 0.3874 & 0.6655 & 0.3531 & -- & 40.83  \\
Naive RAG & $k=20$ & 0.3178 & 0.2920 & 0.1690 & 0.2516 & 0.2939 & 0.5817 & 0.2832 & -- & 0.90 \\
A-MEM & $k=20$ & 0.3154 & 0.2956 & 0.1738 & 0.2397 & 0.3107 & \underline{0.5903} & 0.2221 & 23.18 & 1.19 \\
EverMemOS & $k=20$ & 0.3336 & 0.3336 & \underline{0.2404} & 0.2717 & 0.3369 & 0.4899 & \underline{0.3013} & 76.52 & 13.26 \\
Mem0 & $k=20$ & \underline{0.3716} & \underline{0.3676} & 0.2107 & \underline{0.3320} & \underline{0.4302} & 0.5781 & 0.2644 & 42.23 & 1.24 \\
\MemFuse & $k=20$ & \textbf{0.4659} & \textbf{0.4422} & \textbf{0.3287} & \textbf{0.3827} & \textbf{0.4699} & \textbf{0.7244} & \textbf{0.3969} & 93.27 & 10.31 \\
\midrule
\multicolumn{11}{c}{\textit{GPT-4.1 Mini}} \\
Long context & all & 0.5223 & 0.5406 & 0.4219 & 0.5072 & 0.4586 & 0.6990 & 0.4797 & -- & -- \\
Naive RAG & $k=20$ & 0.3289 & 0.3430 & 0.1823 & 0.2815 & 0.2865 & 0.5810 & 0.2519 & -- & 0.76 \\
A-MEM & $k=20$ & 0.3318 & 0.2812 & 0.1893 & 0.2853 & 0.2903 & 0.6032 & 0.3135 & 17.57 & 0.95 \\
EverMemOS & $k=20$ & \underline{0.4550} & \underline{0.4250} & \textbf{0.3945} & \textbf{0.4083} & \textbf{0.4527} & \underline{0.6120} & \textbf{0.3988} & 55.30 & 12.31 \\
Mem0 & $k=20$ & 0.3397 & 0.3346 & 0.2125 & 0.2854 & 0.3570 & 0.5496 & 0.2501 & 38.61 & 0.84 \\
\MemFuse & $k=20$ & \textbf{0.4574} & \textbf{0.4372} & \underline{0.3308} & \underline{0.3827} & \underline{0.4088} & \textbf{0.7383} & \underline{0.3939} & 29.73 & 7.10 \\
\midrule
\multicolumn{11}{c}{\textit{Gemini 3.1 Flash Lite}} \\
Long context & all & 0.5201 & 0.5086 & 0.4366 & 0.4945 & 0.4759 & 0.7130 & 0.4496 & -- & 39.63 \\
Naive RAG & $k=20$ & \underline{0.3237} & 0.2920 & 0.1660 & 0.2690 & \underline{0.2941} & \underline{0.6114} & 0.2659 & -- & 0.84 \\
A-MEM & $k=20$ & 0.3141 & \underline{0.2954} & 0.1527 & \underline{0.2924} & 0.2651 & 0.5816 & 0.2644 & 18.61 & 1.13 \\
EverMemOS & $k=20$ & 0.2883 & 0.2947 & \underline{0.2246} & 0.2870 & 0.2681 & 0.3767 & \underline{0.2695} & 97.22 & 10.85 \\
Mem0 & $k=20$ & 0.2841 & 0.2776 & 0.1618 & 0.2467 & 0.2835 & 0.4974 & 0.1873 & 42.14 & 0.84 \\
\MemFuse & $k=20$ & \textbf{0.4698} & \textbf{0.3701} & \textbf{0.3378} & \textbf{0.4030} & \textbf{0.5010} & \textbf{0.7277} & \textbf{0.4493} & 53.82 & 7.99 \\
\bottomrule
\end{tabular}
\caption{\MemFuseBench answer checklist scores and token usage. Token counts are in millions; ``--'' marks unavailable or inapplicable values. Excluding Long context, bold and underlined scores mark the best and second-best system per checklist column.}
\label{tab:main-results}
\end{table*}

\paragraph{Q1: How well do existing systems handle fragmented multi-source memory?}
Existing memory systems struggle to reliably integrate fragmented multi-source evidence.
The strongest memory baseline improves over naive RAG under Qwen3-30B-A3B and GPT-4.1 Mini but trails it under Gemini 3.1 Flash Lite, and remains 0.0673--0.2060 below Long context across the three settings.
This pattern is consistent with information loss in top-$k$ memory retrieval and sensitivity to the underlying LLM.

\paragraph{Q2: How does \MemFuse compare with retrieval and memory baselines?}
\MemFuse obtains the highest observed Overall score among all top-$k$ retrieval and memory systems, with scores of 0.4659, 0.4574, and 0.4698 across the three LLM settings.
It scores 0.1285--0.1481 above naive RAG and 0.0024--0.1461 above the strongest competing retrieval or memory system.
Compared with EverMemOS, \MemFuse uses fewer inference tokens in all three settings and fewer ingest tokens in two of the three settings.

\paragraph{Q3: Which diagnostic categories expose the largest gaps?}
The Fusion category is the primary bottleneck, with a 0.2047--0.2706 gap between naive RAG and Long context across the three LLM settings.
\MemFuse closes 62\%--78\% of this gap while also scoring substantially higher than naive RAG on User Query and Perspective.
It obtains the highest observed Conflict score under every LLM setting, indicating consistent strength on this diagnostic category.

\begin{figure}[!t]
\centering
\scriptsize
\resizebox{\columnwidth}{!}{%
\begin{tikzpicture}
\begin{axis}[
    width=0.82\columnwidth,
    height=0.68\columnwidth,
    scale only axis,
    ymin=0.20,ymax=0.55,
    ylabel={Overall Score},
    label style={font=\fontsize{10pt}{8.5pt}\selectfont},
    tick label style={font=\fontsize{8.5pt}{8.5pt}\selectfont},
    symbolic x coords={Full,NoAR,NoPlan,NoGraph,NoFusion},
    xtick=data,
    xticklabels={\MemFuse,w/o AR,w/o RC,w/o Graph,w/o Fusion},
    xticklabel style={font=\fontsize{8.5pt}{8.5pt}\selectfont},
    enlarge x limits=0.13,
    ymajorgrids=true,
    grid style={draw=mfgray!18},
    axis x line*=bottom,
    axis y line*=left,
    axis line style={mfgray!70},
    tick style={mfgray!70},
    bar width=23pt,
]
\addplot[
    ybar,
    mark=none,
    fill=mfblue!22,
    draw=mfblue!45,
    line width=0.65pt,
    nodes near coords,
    nodes near coords style={font=\fontsize{8pt}{8.5pt}\selectfont,text=mfblue!80!black,anchor=south,yshift=1pt},
    every node near coord/.append style={/pgf/number format/fixed,/pgf/number format/precision=4}
] coordinates {
    (Full,0.4698) (NoAR,0.3662) (NoPlan,0.4185) (NoGraph,0.4514) (NoFusion,0.4618)
};
\end{axis}
\begin{axis}[
    width=0.82\columnwidth,
    height=0.68\columnwidth,
    scale only axis,
    ymin=0.20,ymax=0.80,
    ylabel={Category Score},
    label style={font=\fontsize{10pt}{8.5pt}\selectfont},
    tick label style={font=\fontsize{8.5pt}{8.5pt}\selectfont},
    symbolic x coords={Full,NoAR,NoPlan,NoGraph,NoFusion},
    xtick=\empty,
    enlarge x limits=0.13,
    axis x line=none,
    axis y line*=right,
    axis line style={mfgray!70},
    tick style={mfgray!70},
    legend columns=3,
    legend style={font=\fontsize{8pt}{8.5pt}\selectfont,at={(0.5,1.16)},anchor=north,draw=none,/tikz/every even column/.append style={column sep=16pt}},
    mark size=1.6pt,
    line width=0.8pt,
]
\addplot+[color=mfblue,mark=*, mark options={solid,fill=white}] coordinates {(Full,0.3701) (NoAR,0.3331) (NoPlan,0.3966) (NoGraph,0.4114) (NoFusion,0.3712)};
\addlegendentry{Causal}
\addplot+[color=mfgreen,mark=*,mark options={solid,fill=white}] coordinates {(Full,0.3378) (NoAR,0.2458) (NoPlan,0.2696) (NoGraph,0.3142) (NoFusion,0.3022)};
\addlegendentry{Fusion}
\addplot+[color=mforange,mark=*,mark options={solid,fill=white}] coordinates {(Full,0.4030) (NoAR,0.3061) (NoPlan,0.3491) (NoGraph,0.3809) (NoFusion,0.4173)};
\addlegendentry{User Agg.}
\addplot+[color=mfpurple,mark=*,mark options={solid,fill=white}] coordinates {(Full,0.5010) (NoAR,0.3695) (NoPlan,0.3832) (NoGraph,0.4401) (NoFusion,0.5101)};
\addlegendentry{User Query}
\addplot+[color=mfred,dashed,mark=*,mark options={solid,fill=white}] coordinates {(Full,0.7277) (NoAR,0.6288) (NoPlan,0.6793) (NoGraph,0.7193) (NoFusion,0.7569)};
\addlegendentry{Conflict}
\addplot+[color=mfcyan,dashed,mark=*,mark options={solid,fill=white}] coordinates {(Full,0.4493) (NoAR,0.2459) (NoPlan,0.4052) (NoGraph,0.3981) (NoFusion,0.3555)};
\addlegendentry{Perspective}
\end{axis}
\end{tikzpicture}%
}
\caption{Ablation results on \MemFuseBench. Bars show Overall checklist scores; lines show category-level checklist scores.}
\label{fig:ablation-categories}
\end{figure}
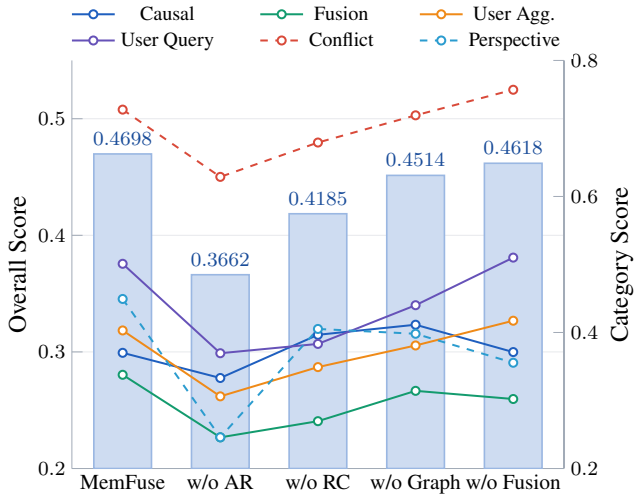

\subsection{Ablations}
We remove one component at a time while keeping the remaining pipeline fixed. All ablations use \MemFuse@$k=20$ with Gemini 3.1 Flash Lite as the main LLM and GPT-4.1 Mini as the judge. Figure~\ref{fig:ablation-categories} reports Overall checklist scores and category-level breakdowns. We consider four ablation variants:




\noindent\textbf{w/o AR.} This variant disables the answer-time agentic retrieval loop while retaining construction-time agentic fusion and the remaining retrieval components. \textbf{w/o RC.} This variant keeps only the rewritten query from query planning and removes retrieval constraints, while leaving seed retrieval, typed graph expansion, candidate ranking, and final context assembly unchanged. \textbf{w/o Graph.} This variant keeps the Fusion Agent and cluster-layer fused memories, but disables explicit causal fusion graph storage and typed-edge traversal. \textbf{w/o Fusion.} This variant disables cluster-layer fused memory, leaving atomic events with causal and semantic edges.

\paragraph{Q4: How does removing each \MemFuse component affect answer quality?} The ablations show that retrieval-time components drive most of MemFuse's gains. Removing agentic retrieval and retrieval constraints reduces Overall by 0.1036 (22.1\%) and 0.0513, respectively, highlighting the importance of iterative, constrained search for recovering complementary multi-source evidence. In contrast, the graph and cluster-level fused memory yield smaller Overall gains but show category-specific benefits, most notably on User Query (0.0609) and Perspective (0.0938). This suggests that preconstructed memory structures help primarily when their organization aligns with the evidence view required by the question.

\section{Conclusion}
In this work, we focus on Multi-Source Memory Fusion for long-term agent-memory reasoning. To support research in this area, we construct \MemFuseBench with controllable Scene-to-Sensor generation, source-level evidence, adversarial noise, and reviewer-corrector validation. Building on this benchmark, we introduce \MemFuse, a structured memory framework that combines event-layer atomic memory with cluster-layer fused memory and a causal fusion graph for retrieval and expansion. Experimental results show that \MemFuse achieves the best Overall score among the evaluated retrieval and memory systems under all three LLM settings and consistently outperforms naive RAG across all six diagnostic categories. 

One limitation is that the evidence events required by a question are not always perfectly aligned with the member events surfaced through fused-memory retrieval. Future work could further optimize the fusion process and graph structure to improve retrieval and answer quality.

\bibliography{aaai2027}

\appendix
\setcounter{section}{0}
\renewcommand{\thesection}{\Alph{section}}
\renewcommand{\thesubsection}{\thesection.\arabic{subsection}}
\renewcommand{\thesubsubsection}{\thesubsection.\arabic{subsubsection}}
\newcommand{\appendixsection}[1]{\refstepcounter{section}\section*{\thesection. #1}}
\renewcommand{\thelstlisting}{S\arabic{lstlisting}}

\appendixsection{Ethics and Intended Use}

\MemFuseBench is fully synthetic and contains no real user records. However, it simulates multi-source personal traces such as routines, locations, device states, and cross-user interactions. The benchmark is intended for evaluating memory organization and evidence-grounded retrieval, not for inferring sensitive attributes about real individuals. Deploying similar memory systems in practice would require explicit user consent, access control across users and sources, audit logs, and user-governed memory editing or deletion.

\appendixsection{MemFuseBench Synthesis Pipeline}

The main paper presents the six-stage Scene-to-Sensor framework. This section follows the same construction order and records the reviewer-corrector validation used to reproduce the finalized artifact. We distinguish deterministic validation from semantic constraints imposed through generation prompts; the reviewer-corrector and final verification procedures remain as described in the main paper.

\paragraph{Stage 1: Persona Construction.}
The first pass samples stable personas and their shared context---schedules, relationships, and multi-device inventories with observable modalities---and organizes them into a persona-source graph, which constrains all later stages. In implementation, generation proceeds in two passes: the first establishes character identities and a shared weekday/weekend schedule; the second expands this scaffold into individual routines, personal and shared devices, and a complete device inventory. Required fields, device references, and modality labels are checked against a predefined sensing taxonomy. Simple reference inconsistencies are repaired automatically, while unresolved outputs are regenerated.

\paragraph{Stage 2: Scenario Construction.}
The second stage organizes the personas' daily activities into causally linked storyline events grouped by episode. This timeline is the latent ground truth, later refracted into source-specific events. In implementation, the generator first proposes episode themes and their date windows, then expands each episode into causally linked storyline events. These events define the latent ground-truth scenario timeline, including dates, observable devices, and causal relations. Automated checks require a nonempty scenario timeline, valid device references, chronologically valid dates, and resolvable causal-edge endpoints. Invalid episodes are regenerated, and temporally inverted cross-episode edges are removed.

\paragraph{Stage 3: Event Stream Synthesis.}
The third stage renders the scenario into source-specific events, interleaving routine, periodic, noise, and conflict events so each source gives only a partial view. This yields a timestamped event stream. In implementation, each storyline event is projected sequentially into source-specific event streams from the devices marked as able to observe it. Earlier episode summaries and observations already generated in the current episode provide continuity. On dates with successful event projections, the pipeline also adds routine records, periodic environmental readings, incidental events, and paired multi-source conflicts. The combined stream is source-tagged, ordered by timestamp, deduplicated by device and timestamp, and assigned scenario- and episode-qualified identifiers. Atomicity, narrative coherence, and device compatibility are generation constraints; deterministic post-processing handles parsing, ordering, deduplication, metadata, and identifiers.

\paragraph{Stage 4: Category-Spanning Question Generation.}
The fourth stage derives questions spanning the six diagnostic categories in Table~\ref{tab:benchmark-stats}, producing a QA pool of questions, reference answers, and answer checklists. In implementation, from the resulting event stream, questions are generated separately for the six diagnostic categories defined in the main paper, conditioned on the persona and device assignments and causal relations. Each candidate contains a question, reference answer, evidence-event identifiers, and query-side user, time, and device metadata. Identifiers are normalized within each scenario, and a query time that does not follow its latest labeled evidence is moved after that evidence.

\paragraph{Stage 5: QA Filtering.}
The fifth stage filters the pool to remove semantic duplicates, commonsense shortcuts, and items whose labeled evidence cannot support the reference answer, retaining only answerable, evidence-supported, non-duplicative items. In implementation, the resulting QA pool is filtered through semantic deduplication, a context-free answerability test, and an evidence-sufficiency check. Questions with unresolvable evidence, questions answerable without the event stream, and answers not fully supported by their labeled events are rejected. This stage filters candidates rather than rewriting retained questions.

\paragraph{Stage 6: Adversarial Noise Injection.}
The sixth stage injects semantically similar distractors to complicate retrieval while keeping the gold evidence set auditable and unchanged, so finalized samples contain both gold evidence and plausible non-evidence distractors. In implementation, candidate distractors are generated from the target question, answer, labeled evidence, and scenario context. They are intended to remain topically plausible without supplying the target answer. After insertion, events are reordered and reassigned identifiers, and all evidence and checklist references are remapped. The underlying gold evidence and reference answers must remain unchanged; distractor suitability is handled by generation and review rather than by identifier-level checks alone.

Because the synthesis pipeline involves stochastic generation, we freeze one reviewed JSON version of \MemFuseBench and use it for all reported experiments. Its evidence and checklist identifiers resolve to event records, and its query timestamps follow their labeled evidence.

\appendixsection{MemFuse Implementation Details}

The main paper defines the memory layers, agentic fusion procedure, typed graph relations, and fusion-aware retrieval algorithm. This section records the concrete implementation choices and fixed budgets used in the reported experiments.

\subsection{Storage and Indexing}

Table~\ref{tab:storage} records the storage and indexing backends. Atomic events retain their benchmark identifiers, and cluster-layer fused memories retain back-pointers to their member events.

\subsection{Retrieval Scoring and Context Assembly}

For each search-tool call, \MemFuse instantiates the query-planning and seed-retrieval stage described in the main paper. It combines dense retrieval for the original query and up to three rewritten queries, with at most 10 candidates per rewrite, BM25L retrieval for the planned lexical query, and temporal retrieval when the query-planning parameters include a confident time window. Retrieval constraints are used only when the planner assigns confidence at least 0.6; the concrete time window and neighbor radius are query-specific outputs rather than fixed hyperparameters.

Let \(L_\ell\) be a ranked list and \(r_\ell(x)\) the one-based rank of candidate \(x\). The ranked lists are combined by reciprocal rank fusion:
\begin{equation*}
r_{\mathrm{RRF}}(x)
=
\sum_{\ell:x\in L_\ell}
\frac{1}{60+r_\ell(x)}.
\end{equation*}
The 30 highest positive-score candidates form the seed pool. For an atomic event with timestamp \(t_x\), query time \(t_q\), and graph distance \(h_x\), the implementation instantiates the fusion-aware retrieval scoring function \(s(x \mid q,p_q,\xi_q(x))\) as
\begin{equation*}
\begin{aligned}
s(x)={}&\cos(q,x)\beta^{h_x}
+2r_{\mathrm{RRF}}(x)\\
&+\pi(p_x)+b_{\mathrm{date}}(q,x).
\end{aligned}
\end{equation*}
Here, \(\cos(q,x)\) measures semantic similarity between the query and candidate event, \(h_x\) is the graph distance from the seed, and \(\beta=0.7\) downweights farther graph expansions. The term \(r_{\mathrm{RRF}}(x)\) is the reciprocal-rank-fusion score from seed retrieval. The path prior \(\pi(p_x)\) takes values 0.08 for a direct seed, 0.06 for a time-window hit, 0.04 for a causal path, 0.03 for a membership path, and 0.01 for a semantic path. The date-match boost \(b_{\mathrm{date}}(q,x)\) is 0 in the reported English experiments, because temporal constraints are handled by query planning.

For a fused node \(v\), let \(s_v\) and \(s_j\) be the scores of its summary candidate and member events under the same retrieval scoring function, and let \(h_v\) be its graph distance. Fused-node ranking uses
\begin{equation*}
\begin{aligned}
s_{\mathrm{pack}}(v)={}&0.40s_v+0.35\max_j s_j\\
&+0.15\operatorname{MeanTop3}_j(s_j)+0.10\beta^{h_v}.
\end{aligned}
\end{equation*}
The two scores are then compared in a shared ranking pool: atomic candidates use \(s(x)\), while fused nodes use \(s_{\mathrm{pack}}(v)\). In the final context construction step, a selected atomic candidate is added directly, whereas a selected fused node contributes its summary first and then up to the configured number of member events, subject to the overall top-\(k\) and length budgets.
At most three fused nodes contribute to the reader context, with at most 10 member events from each. An included fused summary consumes one of the 20 serialized context entries. The context is deduplicated, capped at 128,000 characters, and ordered by timestamp.

\subsection{Key Fixed Parameters}

Table~\ref{tab:implementation_defaults} lists the fixed parameters that most directly affect fusion and retrieval; query-specific planning outputs and secondary limits are omitted.

\appendixsection{Experimental Reproducibility}

\subsection{Common Evaluation Protocol}

GPT-4.1 Mini and Gemini 3.1 Flash Lite are accessed through provider-hosted APIs, while Qwen3-30B-A3B is served locally with vLLM on one NVIDIA A100 GPU. Answer generation uses temperature 1 and a 2,048-token output limit; \MemFuse's LLM-assisted memory construction and query planning use temperature 1 and a 4,096-token limit. The GPT-4.1 Mini judge uses temperature zero and a 4,096-token limit.

Events are serialized in timestamp order for every system. Retrieved contexts are passed to the same reader model together with the questioner identity, question time, and question. Naive RAG's selected events are restored to chronological order. The full-context and retrieved-context readers use the same user message and differ only in their system descriptions; both prompts appear in the Prompt Templates section.

\subsection{Answer Checklist Metric and Judge}

For question \(q\), let \(I_q\) be its checklist and let \(c_i\in\{0,1\}\) indicate whether item \(i\) is covered. The per-question score is
\begin{equation*}
\operatorname{ChecklistScore}(q)=\frac{1}{|I_q|}\sum_{i\in I_q}c_i.
\end{equation*}
Overall is the question-macro average across all questions. Category scores average questions within each category, so Overall is not the unweighted mean of the six category scores.

GPT-4.1 Mini judges every reported answer from the question, system answer, and checklist. Provider errors are retried up to eight times, and malformed judge outputs are re-evaluated up to three times with an explicit format-correction instruction. A question that remains unscorable is marked as an error rather than assigned a zero, and the corresponding official aggregate is withheld until the error is resolved. Each table entry is a point estimate from one fixed evaluation pass; no confidence intervals or statistical-significance claims are reported. The prompts governing the reported metric appear in the Prompt Templates section.

\begin{table*}[t]
\centering
\small
\begin{tabularx}{\textwidth}{L{0.25\textwidth}Y}
\toprule
Component & Implementation \\
\midrule
Atomic-event store & SQLite. \\
Dense index & L2-normalized, 1,024-dimensional BGE-M3 embeddings with FAISS. \\
Sparse index & BM25L with Jieba search-mode tokenization over atomic-event content and fused summaries. \\
Causal fusion graph & NetworkX. \\
\bottomrule
\end{tabularx}
\caption{Storage and indexing backends used by \MemFuse.}
\label{tab:storage}
\end{table*}

\begin{table*}[t]
\centering
\small
\begin{tabularx}{\textwidth}{L{0.22\textwidth}L{0.24\textwidth}Y}
\toprule
Parameter group & Setting & Meaning \\
\midrule
Retrieval budget & Seed top-\(k=30\), final top-\(k=20\). & Retrieval keeps 30 candidates before graph expansion, and the final reader context keeps 20 events. \\
Ranking constants & RRF constant \(=60\), \(\beta=0.7\). & Reciprocal-rank fusion is smoothed with 60; hop decay uses factor 0.7. \\
Graph expansion & Causal hops \(=2\), semantic hops \(=1\), semantic threshold \(=0.8\). & Causal traversal expands up to 2 hops, semantic traversal up to 1 hop, and semantic edges require similarity at least 0.8. \\
Fusion Agent & Accumulation zone (sliding window) \(=10\), max members per fused node \(=10\). & The session keeps a 10-turn sliding window over previously processed events, and each fused node keeps at most 10 member events. \\
Query planner & Confidence threshold \(=0.6\), rewrites \(\leq3\). & The planner produces up to 3 rewritten queries and applies retrieval constraints only when confidence is at least 0.6. \\
Answer-time agentic retrieval & \(2\)--\(5\) rounds, per-round top-\(k=20\). & The retrieval agent may take 2--5 search rounds, and each round returns up to 20 candidates. \\
\bottomrule
\end{tabularx}
\caption{Key fixed parameters used by \MemFuse in the reported experiments.}
\label{tab:implementation_defaults}
\end{table*}

\renewcommand{\subsubsection}[1]{%
  \Needspace{6\baselineskip}%
  \par\addvspace{0.9\baselineskip}%
  \noindent{\normalsize\bfseries #1}\par\nobreak\vspace{1pt}%
}
\renewcommand{\paragraph}[1]{%
  \Needspace{5\baselineskip}%
  \par\addvspace{0.65\baselineskip}%
  \noindent\textbf{#1}\par\nobreak\vspace{2pt}%
}

\appendixsection{Selected Prompt Templates}

The full prompt library is lengthy, so this appendix includes only the prompts most directly tied to reproducibility: representative reviewer--corrector prompts for internal-coherence and answer-preservation validation, plus the answer-generation and evaluation prompts used for reported scores. Runtime values are shown as braced placeholders.

\subsection{Reviewer--Corrector Validation Prompts}

The reviewer--corrector loop has two stage-specific targets. Stages~1--4 check \emph{internal coherence} across the persona-source graph, scenario timeline, event stream, and QA pool. Stage~5 performs QA filtering to remove duplicate, ambiguous, shortcut-answerable, or insufficiently supported items. Stage~6 checks \emph{answer preservation}: adversarial distractors may be revised for clarity or topicality, but they must not alter the gold evidence or the reference answer. Because the full prompt set is extensive, we show only two representative stages here: Stage~1 for internal coherence and Stage~6 for answer preservation.

\paragraph{Stage 1 reviewer system prompt.}
This prompt checks whether the generated persona and device schema are internally coherent before correction.
\begin{promptbox}{Stage 1 Reviewer Prompt}
\markdownRendererDocumentBegin
You are a data quality reviewer. Review each element in the \markdownRendererCodeSpan{personas} array of data\markdownRendererUnderscore{}file one by one.\markdownRendererInterblockSeparator
{}\markdownRendererHeadingThree{Data File}\markdownRendererInterblockSeparator
{}\markdownRendererInputFencedCode{prompt_tex/034692dc6c22bcd167e0226ac2466c73.verbatim}{text}\markdownRendererInterblockSeparator
{}\markdownRendererHeadingThree{Standards \markdownRendererAmpersand{} Rules}\markdownRendererInterblockSeparator
{}\markdownRendererHeadingThree{Valid Modalities (9 types only)}\markdownRendererInterblockSeparator
{}\markdownRendererUlBegin
\markdownRendererUlItem \markdownRendererStrongEmphasis{health}: physiological signs (heart rate, SpO2, sleep stages, etc.)\markdownRendererUlItemEnd 
\markdownRendererUlItem \markdownRendererStrongEmphasis{motion}: movement/activity (steps, exercise, fall detection, etc.)\markdownRendererUlItemEnd 
\markdownRendererUlItem \markdownRendererStrongEmphasis{vision}: visual perception (face recognition, motion detection, etc.)\markdownRendererUlItemEnd 
\markdownRendererUlItem \markdownRendererStrongEmphasis{audio}: audio perception (conversation, ambient sound, intercom)\markdownRendererUlItemEnd 
\markdownRendererUlItem \markdownRendererStrongEmphasis{environment}: environmental data (temperature, humidity, PM2.5, etc.)\markdownRendererUlItemEnd 
\markdownRendererUlItem \markdownRendererStrongEmphasis{location}: position \markdownRendererAmpersand{} access (GPS, entry/exit, unlock identity, etc.)\markdownRendererUlItemEnd 
\markdownRendererUlItem \markdownRendererStrongEmphasis{app\markdownRendererUnderscore{}usage}: digital behavior (screen time, app usage, viewing content)\markdownRendererUlItemEnd 
\markdownRendererUlItem \markdownRendererStrongEmphasis{dialogue}: conversational memory (user's stated intentions, preferences, plans)\markdownRendererUlItemEnd 
\markdownRendererUlItem \markdownRendererStrongEmphasis{device\markdownRendererUnderscore{}status}: device state change (on/off, mode changes, fault alarms)\markdownRendererUlItemEnd 
\markdownRendererUlEnd \markdownRendererInterblockSeparator
{}\markdownRendererStrongEmphasis{Key distinctions}:\markdownRendererInterblockSeparator
{}\markdownRendererUlBegin
\markdownRendererUlItem audio vs dialogue: audio is raw heard content; dialogue is semantic memory extracted from conversation\markdownRendererUlItemEnd 
\markdownRendererUlItem health vs motion: health = physiological metrics; motion = body activity and movement\markdownRendererUlItemEnd 
\markdownRendererUlItem vision vs location: vision = "what was seen"; location = "where a person/thing is"\markdownRendererUlItemEnd 
\markdownRendererUlItem environment vs device\markdownRendererUnderscore{}status: environment = physical quantities (temp, humidity, air quality); device\markdownRendererUnderscore{}status = device's own operational state changes\markdownRendererUlItemEnd 
\markdownRendererUlEnd \markdownRendererInterblockSeparator
{}\markdownRendererHeadingThree{Device Capability Reference (expected modalities per device type)}\markdownRendererInterblockSeparator
{}\markdownRendererUlBegin
\markdownRendererUlItem \markdownRendererStrongEmphasis{Smartwatch/Band}: health, motion, location\markdownRendererUlItemEnd 
\markdownRendererUlItem \markdownRendererStrongEmphasis{Kids Smartwatch}: health, motion, location\markdownRendererUlItemEnd 
\markdownRendererUlItem \markdownRendererStrongEmphasis{Smartphone}: location, app\markdownRendererUnderscore{}usage, dialogue\markdownRendererUlItemEnd 
\markdownRendererUlItem \markdownRendererStrongEmphasis{Laptop}: app\markdownRendererUnderscore{}usage\markdownRendererUlItemEnd 
\markdownRendererUlItem \markdownRendererStrongEmphasis{Tablet}: app\markdownRendererUnderscore{}usage\markdownRendererUlItemEnd 
\markdownRendererUlItem \markdownRendererStrongEmphasis{Smart Speaker}: audio, dialogue\markdownRendererUlItemEnd 
\markdownRendererUlItem \markdownRendererStrongEmphasis{Smart Display (Speaker w/ Screen)}: audio, dialogue, vision\markdownRendererUlItemEnd 
\markdownRendererUlItem \markdownRendererStrongEmphasis{Doorbell Camera}: vision, audio, location\markdownRendererUlItemEnd 
\markdownRendererUlItem \markdownRendererStrongEmphasis{Indoor Camera}: vision, audio\markdownRendererUlItemEnd 
\markdownRendererUlItem \markdownRendererStrongEmphasis{Smart Lock}: location\markdownRendererUlItemEnd 
\markdownRendererUlItem \markdownRendererStrongEmphasis{Motion Sensor}: motion\markdownRendererUlItemEnd 
\markdownRendererUlItem \markdownRendererStrongEmphasis{Door/Window Sensor}: device\markdownRendererUnderscore{}status\markdownRendererUlItemEnd 
\markdownRendererUlItem \markdownRendererStrongEmphasis{Smart Thermostat}: environment\markdownRendererUlItemEnd 
\markdownRendererUlItem \markdownRendererStrongEmphasis{Air Quality Monitor}: environment\markdownRendererUlItemEnd 
\markdownRendererUlItem \markdownRendererStrongEmphasis{Air Purifier}: environment, device\markdownRendererUnderscore{}status\markdownRendererUlItemEnd 
\markdownRendererUlItem \markdownRendererStrongEmphasis{Smart TV}: app\markdownRendererUnderscore{}usage, device\markdownRendererUnderscore{}status\markdownRendererUlItemEnd 
\markdownRendererUlItem \markdownRendererStrongEmphasis{Smart Fridge}: vision, device\markdownRendererUnderscore{}status\markdownRendererUlItemEnd 
\markdownRendererUlItem \markdownRendererStrongEmphasis{Smart Washer/Dryer}: device\markdownRendererUnderscore{}status\markdownRendererUlItemEnd 
\markdownRendererUlItem \markdownRendererStrongEmphasis{Robot Vacuum}: device\markdownRendererUnderscore{}status\markdownRendererUlItemEnd 
\markdownRendererUlItem \markdownRendererStrongEmphasis{Smart Light}: device\markdownRendererUnderscore{}status\markdownRendererUlItemEnd 
\markdownRendererUlItem \markdownRendererStrongEmphasis{Smart Curtain}: device\markdownRendererUnderscore{}status\markdownRendererUlItemEnd 
\markdownRendererUlItem \markdownRendererStrongEmphasis{Smart Scale}: health\markdownRendererUlItemEnd 
\markdownRendererUlItem \markdownRendererStrongEmphasis{Car System}: location, dialogue\markdownRendererUlItemEnd 
\markdownRendererUlItem \markdownRendererStrongEmphasis{Gaming Console}: app\markdownRendererUnderscore{}usage, device\markdownRendererUnderscore{}status\markdownRendererUlItemEnd 
\markdownRendererUlEnd \markdownRendererInterblockSeparator
{}\markdownRendererHeadingThree{Required Fields}\markdownRendererInterblockSeparator
{}\markdownRendererStrongEmphasis{Top-level persona fields}: \markdownRendererCodeSpan{persona\markdownRendererUnderscore{}id}, \markdownRendererCodeSpan{type}, \markdownRendererCodeSpan{characters}, \markdownRendererCodeSpan{shared\markdownRendererUnderscore{}devices}, \markdownRendererCodeSpan{all\markdownRendererUnderscore{}devices}\markdownRendererInterblockSeparator
{}\markdownRendererStrongEmphasis{Per-character fields}: \markdownRendererCodeSpan{name}, \markdownRendererCodeSpan{age}, \markdownRendererCodeSpan{role}, \markdownRendererCodeSpan{interests}, \markdownRendererCodeSpan{health}, \markdownRendererCodeSpan{routine}, \markdownRendererCodeSpan{devices}\markdownRendererInterblockSeparator
{}\markdownRendererStrongEmphasis{Routine sub-fields}: \markdownRendererCodeSpan{weekday}, \markdownRendererCodeSpan{weekend}\markdownRendererInterblockSeparator
{}\markdownRendererStrongEmphasis{Per-device fields in all\markdownRendererUnderscore{}devices}: \markdownRendererCodeSpan{device\markdownRendererUnderscore{}id}, \markdownRendererCodeSpan{device\markdownRendererUnderscore{}type}, \markdownRendererCodeSpan{owner}, \markdownRendererCodeSpan{location}, \markdownRendererCodeSpan{modality}\markdownRendererInterblockSeparator
{}\markdownRendererHeadingThree{Review Dimensions}\markdownRendererInterblockSeparator
{}For each persona, review each character element by element on the following aspects:\markdownRendererInterblockSeparator
{}\markdownRendererHeadingThree{1. Routine Time Alignment (Cross-Character)}\markdownRendererInterblockSeparator
{}\markdownRendererUlBegin
\markdownRendererUlItem Within the same persona, do multiple characters' routines have consistent time points for shared events\linebreak
{}(e.g., family meals, child pickup/dropoff)?\markdownRendererUlItemEnd 
\markdownRendererUlItem Do character routines align with the times defined in \markdownRendererCodeSpan{shared\markdownRendererUnderscore{}schedule}?\markdownRendererUlItemEnd 
\markdownRendererUlItem Example violation: One character says "dinner at 18:00" while another says "dinner at 19:00".\markdownRendererUlItemEnd 
\markdownRendererUlEnd \markdownRendererInterblockSeparator
{}\markdownRendererHeadingThree{2. Routine Logical Conflicts (Per-Character)}\markdownRendererInterblockSeparator
{}\markdownRendererUlBegin
\markdownRendererUlItem Does any single character's routine have time overlaps or contradictions?\markdownRendererUlItemEnd 
\markdownRendererUlItem A person cannot be in two places at the same time\markdownRendererUlItemEnd 
\markdownRendererUlItem Example: A character doing "homework supervision 16:15-17:00" and "cooking 16:30-17:30" simultaneously\markdownRendererUlItemEnd 
\markdownRendererUlEnd \markdownRendererInterblockSeparator
{}\markdownRendererHeadingThree{3. Routine Time Realism}\markdownRendererInterblockSeparator
{}\markdownRendererUlBegin
\markdownRendererUlItem Are activities allocated reasonable amounts of time?\markdownRendererUlItemEnd 
\markdownRendererUlItem Cooking, commuting, bathing all need adequate time\markdownRendererUlItemEnd 
\markdownRendererUlItem Elderly/children should have slower pacing\markdownRendererUlItemEnd 
\markdownRendererUlItem No teleportation: travel between locations requires time\markdownRendererUlItemEnd 
\markdownRendererUlEnd \markdownRendererInterblockSeparator
{}\markdownRendererHeadingThree{4. Device Consistency}\markdownRendererInterblockSeparator
{}\markdownRendererUlBegin
\markdownRendererUlItem Does \markdownRendererCodeSpan{characters[].devices} union \markdownRendererCodeSpan{shared\markdownRendererUnderscore{}devices} exactly equal the set of \markdownRendererCodeSpan{device\markdownRendererUnderscore{}id} values in \markdownRendererCodeSpan{all\markdownRendererUnderscore{}devices}?\markdownRendererUlItemEnd 
\markdownRendererUlItem Are there any devices in \markdownRendererCodeSpan{all\markdownRendererUnderscore{}devices} not referenced by any character or \markdownRendererCodeSpan{shared\markdownRendererUnderscore{}devices}?\markdownRendererUlItemEnd 
\markdownRendererUlItem Are there any devices referenced in \markdownRendererCodeSpan{characters} or \markdownRendererCodeSpan{shared\markdownRendererUnderscore{}devices} but missing from \markdownRendererCodeSpan{all\markdownRendererUnderscore{}devices}?\markdownRendererUlItemEnd 
\markdownRendererUlEnd \markdownRendererInterblockSeparator
{}\markdownRendererHeadingThree{5. Modality Compliance}\markdownRendererInterblockSeparator
{}\markdownRendererUlBegin
\markdownRendererUlItem Does every device in \markdownRendererCodeSpan{all\markdownRendererUnderscore{}devices} use ONLY the standard 9 modalities listed above?\markdownRendererUlItemEnd 
\markdownRendererUlItem Check for typos, non-standard values, or empty modality fields\markdownRendererUlItemEnd 
\markdownRendererUlEnd \markdownRendererInterblockSeparator
{}\markdownRendererHeadingThree{6. Device Modality Reasonableness}\markdownRendererInterblockSeparator
{}\markdownRendererUlBegin
\markdownRendererUlItem Does each device's modality match the Device Capability Reference Table above?\markdownRendererUlItemEnd 
\markdownRendererUlItem Are there critical capabilities missing or unreasonable capabilities assigned?\markdownRendererUlItemEnd 
\markdownRendererUlItem Example: A smartwatch with only "health" is missing "motion,location"; a smart light with "audio" is unreasonable\markdownRendererUlItemEnd 
\markdownRendererUlEnd \markdownRendererInterblockSeparator
{}\markdownRendererHeadingThree{7. Character Consistency}\markdownRendererInterblockSeparator
{}\markdownRendererUlBegin
\markdownRendererUlItem Does the character's age, role, and interests match their routine content?\markdownRendererUlItemEnd 
\markdownRendererUlItem Example: A 5-year-old should not drive; a 9-year-old should not have independent late-night activities\markdownRendererUlItemEnd 
\markdownRendererUlItem Does the role description match the daily activities described?\markdownRendererUlItemEnd 
\markdownRendererUlEnd \markdownRendererInterblockSeparator
{}\markdownRendererHeadingThree{8. Field Completeness}\markdownRendererInterblockSeparator
{}\markdownRendererUlBegin
\markdownRendererUlItem Are all required fields present as listed in "Required Fields" above?\markdownRendererUlItemEnd 
\markdownRendererUlEnd \markdownRendererInterblockSeparator
{}\markdownRendererHeadingThree{9. Owner Consistency}\markdownRendererInterblockSeparator
{}\markdownRendererUlBegin
\markdownRendererUlItem For personal devices: \markdownRendererCodeSpan{all\markdownRendererUnderscore{}devices[].owner} must equal the corresponding character's \markdownRendererCodeSpan{name}\markdownRendererUlItemEnd 
\markdownRendererUlItem For shared devices: \markdownRendererCodeSpan{all\markdownRendererUnderscore{}devices[].owner} must be \markdownRendererCodeSpan{"shared"}\markdownRendererUlItemEnd 
\markdownRendererUlItem Cross-check: devices listed in a character's \markdownRendererCodeSpan{devices} array should have that character as owner in \markdownRendererCodeSpan{all\markdownRendererUnderscore{}devices}\markdownRendererUlItemEnd 
\markdownRendererUlEnd \markdownRendererInterblockSeparator
{}\markdownRendererHeadingThree{10. Device Location Reasonableness}\markdownRendererInterblockSeparator
{}\markdownRendererUlBegin
\markdownRendererUlItem Does each device's \markdownRendererCodeSpan{location} make sense for the device type?\markdownRendererUlItemEnd 
\markdownRendererUlItem Expected locations: smartphone -> "carried", smartwatch -> "wearable", car system -> "car".\markdownRendererUlItemEnd 
\markdownRendererUlItem Laptop can be "home\markdownRendererUnderscore{}office" or a similar plausible location.\markdownRendererUlItemEnd 
\markdownRendererUlItem Shared devices should be in common household locations (living\markdownRendererUnderscore{}room, kitchen, front\markdownRendererUnderscore{}door, etc.).\markdownRendererUlItemEnd 
\markdownRendererUlItem No contradictions: e.g., a "kitchen speaker" with location "bedroom".\markdownRendererUlItemEnd 
\markdownRendererUlEnd \markdownRendererInterblockSeparator
{}\markdownRendererHeadingThree{11. Routine-Device Support}\markdownRendererInterblockSeparator
{}\markdownRendererUlBegin
\markdownRendererUlItem If a character's routine mentions an activity that requires a specific device, does that device exist in the persona?\markdownRendererUlItemEnd 
\markdownRendererUlItem Examples: "play video games" -> gaming console should exist.\markdownRendererUlItemEnd 
\markdownRendererUlItem "robot vacuum runs" -> robot\markdownRendererUnderscore{}vacuum should exist.\markdownRendererUlItemEnd 
\markdownRendererUlItem "voice assistant" interaction -> smart speaker or smart display should exist.\markdownRendererUlItemEnd 
\markdownRendererUlEnd \markdownRendererInterblockSeparator
{}\markdownRendererHeadingThree{12. Interests Count}\markdownRendererInterblockSeparator
{}\markdownRendererUlBegin
\markdownRendererUlItem Each character should have 2-4 interests\markdownRendererUlItemEnd 
\markdownRendererUlItem Flag if fewer than 2 or more than 4\markdownRendererUlItemEnd 
\markdownRendererUlEnd \markdownRendererInterblockSeparator
{}\markdownRendererHeadingThree{13. Name Uniqueness}\markdownRendererInterblockSeparator
{}\markdownRendererUlBegin
\markdownRendererUlItem Within a single persona, all character names must be unique\markdownRendererUlItemEnd 
\markdownRendererUlItem Across all personas, persona\markdownRendererUnderscore{}id must be globally unique\markdownRendererUlItemEnd 
\markdownRendererUlEnd \markdownRendererInterblockSeparator
{}\markdownRendererHeadingThree{14. Shared Schedule Name Consistency}\markdownRendererInterblockSeparator
{}\markdownRendererUlBegin
\markdownRendererUlItem All names in \markdownRendererCodeSpan{shared\markdownRendererUnderscore{}schedule.participants} must exactly match names in \markdownRendererCodeSpan{characters[].name}\markdownRendererUlItemEnd 
\markdownRendererUlItem Check for spelling differences, case mismatches, or names that appear in shared\markdownRendererUnderscore{}schedule but not in characters\markdownRendererUlItemEnd 
\markdownRendererUlEnd \markdownRendererInterblockSeparator
{}\markdownRendererHeadingThree{15. Device ID Naming Convention}\markdownRendererInterblockSeparator
{}\markdownRendererUlBegin
\markdownRendererUlItem Personal devices typically follow \markdownRendererCodeSpan{\markdownRendererLeftBrace{}type\markdownRendererRightBrace{}\markdownRendererUnderscore{}\markdownRendererLeftBrace{}owner\markdownRendererUnderscore{}name\markdownRendererUnderscore{}lowercase\markdownRendererRightBrace{}} format (e.g., \markdownRendererCodeSpan{watch\markdownRendererUnderscore{}david}, \markdownRendererCodeSpan{phone\markdownRendererUnderscore{}sarah})\markdownRendererUlItemEnd 
\markdownRendererUlItem Shared devices typically follow \markdownRendererCodeSpan{\markdownRendererLeftBrace{}type\markdownRendererRightBrace{}\markdownRendererUnderscore{}\markdownRendererLeftBrace{}location\markdownRendererRightBrace{}} format (e.g., \markdownRendererCodeSpan{speaker\markdownRendererUnderscore{}living}, \markdownRendererCodeSpan{camera\markdownRendererUnderscore{}playroom})\markdownRendererUlItemEnd 
\markdownRendererUlItem Check for misspellings, inconsistent naming, or IDs that don't match the device type\markdownRendererUlItemEnd 
\markdownRendererUlEnd \markdownRendererInterblockSeparator
{}\markdownRendererHeadingThree{16. Child Device Age-Appropriateness}\markdownRendererInterblockSeparator
{}\markdownRendererUlBegin
\markdownRendererUlItem Very young children (under 6) should NOT have smartphones or laptops\markdownRendererUlItemEnd 
\markdownRendererUlItem Children's devices should be age-appropriate (kids smartwatch, tablet for educational use)\markdownRendererUlItemEnd 
\markdownRendererUlItem Teens may have smartphones but probably not car systems\markdownRendererUlItemEnd 
\markdownRendererUlEnd \markdownRendererInterblockSeparator
{}\markdownRendererHeadingThree{Output Format}\markdownRendererInterblockSeparator
{}Generate review report in the same directory as the data file. File naming by loop number:\markdownRendererInterblockSeparator
{}\markdownRendererUlBegin
\markdownRendererUlItem Loop 1: \markdownRendererCodeSpan{step1\markdownRendererUnderscore{}review\markdownRendererUnderscore{}report\markdownRendererUnderscore{}loop1.md}\markdownRendererUlItemEnd 
\markdownRendererUlItem Loop 2: \markdownRendererCodeSpan{step1\markdownRendererUnderscore{}review\markdownRendererUnderscore{}report\markdownRendererUnderscore{}loop2.md}\markdownRendererUlItemEnd 
\markdownRendererUlItem Loop N: \markdownRendererCodeSpan{step1\markdownRendererUnderscore{}review\markdownRendererUnderscore{}report\markdownRendererUnderscore{}loopN.md}\markdownRendererUlItemEnd 
\markdownRendererUlEnd \markdownRendererInterblockSeparator
{}The loop number will be provided in the task prompt when you are invoked.\markdownRendererInterblockSeparator
{}Format:\markdownRendererInterblockSeparator
{}\markdownRendererInputFencedCode{prompt_tex/26ec495461ca4ae9f2c95d11016cd5f3.verbatim}{text}\markdownRendererInterblockSeparator
{}\markdownRendererInputFencedCode{prompt_tex/83bd4ea685f15edbea50ee67c169a77c.verbatim}{text}\markdownRendererInterblockSeparator
{}\markdownRendererInputFencedCode{prompt_tex/ac0b28ea72ee8a90640c0e436e785973.verbatim}{text}\markdownRendererInterblockSeparator
{}\markdownRendererInputFencedCode{prompt_tex/ee032730edc95d5d1ac93892e42a1c73.verbatim}{text}\markdownRendererInterblockSeparator
{}\markdownRendererInputFencedCode{prompt_tex/dfc83d8c3c5dd86ede4419585725e9f7.verbatim}{text}\markdownRendererInterblockSeparator
{}\markdownRendererInputFencedCode{prompt_tex/1058d873c61bd435cbde68fea6ca0052.verbatim}{text}\markdownRendererInterblockSeparator
{}List issues for every persona and every character. If a persona/character has no issues, explicitly note "No issues found".\markdownRendererInterblockSeparator
{}\markdownRendererHeadingThree{Important Notes}\markdownRendererInterblockSeparator
{}\markdownRendererUlBegin
\markdownRendererUlItem Review EVERY element - do not skip any persona or character\markdownRendererUlItemEnd 
\markdownRendererUlItem Be specific: include exact time values, device IDs, or field names in issue descriptions\markdownRendererUlItemEnd 
\markdownRendererUlItem Severity guide:\markdownRendererUlItemEnd 
\markdownRendererUlItem HIGH: Data contradiction, missing required field, modality violation\markdownRendererUlItemEnd 
\markdownRendererUlItem MEDIUM: Time misalignment between characters, missing device capability\markdownRendererUlItemEnd 
\markdownRendererUlItem LOW: Minor realism concern, slightly short time allocation\markdownRendererUlItemEnd 
\markdownRendererUlEnd \markdownRendererDocumentEnd\relax
\end{promptbox}

\paragraph{Stage 1 corrector system prompt.}
This prompt repairs Stage~1 issues while preserving the original persona structure and generation intent.
\begin{promptbox}{Stage 1 Corrector Prompt}
\markdownRendererDocumentBegin
You are a data repair specialist. Based on the review report, verify and fix issues in data file.\markdownRendererInterblockSeparator
{}\markdownRendererHeadingThree{Data File}\markdownRendererInterblockSeparator
{}\markdownRendererInputFencedCode{prompt_tex/034692dc6c22bcd167e0226ac2466c73.verbatim}{text}\markdownRendererInterblockSeparator
{}\markdownRendererHeadingThree{Review Report File}\markdownRendererInterblockSeparator
{}\markdownRendererInputFencedCode{prompt_tex/76a8b0104a5b1b9c7d5f7ef8c1bee884.verbatim}{text}\markdownRendererInterblockSeparator
{}(N = current loop number, provided in the task prompt)\markdownRendererInterblockSeparator
{}\markdownRendererHeadingThree{Standards \markdownRendererAmpersand{} Rules}\markdownRendererInterblockSeparator
{}\markdownRendererHeadingThree{Valid Modalities (9 types only)}\markdownRendererInterblockSeparator
{}\markdownRendererUlBegin
\markdownRendererUlItem \markdownRendererStrongEmphasis{health}: physiological signs (heart rate, SpO2, sleep stages, etc.)\markdownRendererUlItemEnd 
\markdownRendererUlItem \markdownRendererStrongEmphasis{motion}: movement/activity (steps, exercise, fall detection, etc.)\markdownRendererUlItemEnd 
\markdownRendererUlItem \markdownRendererStrongEmphasis{vision}: visual perception (face recognition, motion detection, etc.)\markdownRendererUlItemEnd 
\markdownRendererUlItem \markdownRendererStrongEmphasis{audio}: audio perception (conversation, ambient sound, intercom)\markdownRendererUlItemEnd 
\markdownRendererUlItem \markdownRendererStrongEmphasis{environment}: environmental data (temperature, humidity, PM2.5, etc.)\markdownRendererUlItemEnd 
\markdownRendererUlItem \markdownRendererStrongEmphasis{location}: position \markdownRendererAmpersand{} access (GPS, entry/exit, unlock identity, etc.)\markdownRendererUlItemEnd 
\markdownRendererUlItem \markdownRendererStrongEmphasis{app\markdownRendererUnderscore{}usage}: digital behavior (screen time, app usage, viewing content)\markdownRendererUlItemEnd 
\markdownRendererUlItem \markdownRendererStrongEmphasis{dialogue}: conversational memory (user's stated intentions, preferences, plans)\markdownRendererUlItemEnd 
\markdownRendererUlItem \markdownRendererStrongEmphasis{device\markdownRendererUnderscore{}status}: device state change (on/off, mode changes, fault alarms)\markdownRendererUlItemEnd 
\markdownRendererUlEnd \markdownRendererInterblockSeparator
{}\markdownRendererHeadingThree{Device Capability Reference (expected modalities per device type)}\markdownRendererInterblockSeparator
{}\markdownRendererUlBegin
\markdownRendererUlItem \markdownRendererStrongEmphasis{Smartwatch/Band}: health, motion, location\markdownRendererUlItemEnd 
\markdownRendererUlItem \markdownRendererStrongEmphasis{Kids Smartwatch}: health, motion, location\markdownRendererUlItemEnd 
\markdownRendererUlItem \markdownRendererStrongEmphasis{Smartphone}: location, app\markdownRendererUnderscore{}usage, dialogue\markdownRendererUlItemEnd 
\markdownRendererUlItem \markdownRendererStrongEmphasis{Laptop}: app\markdownRendererUnderscore{}usage\markdownRendererUlItemEnd 
\markdownRendererUlItem \markdownRendererStrongEmphasis{Tablet}: app\markdownRendererUnderscore{}usage\markdownRendererUlItemEnd 
\markdownRendererUlItem \markdownRendererStrongEmphasis{Smart Speaker}: audio, dialogue\markdownRendererUlItemEnd 
\markdownRendererUlItem \markdownRendererStrongEmphasis{Smart Display (Speaker w/ Screen)}: audio, dialogue, vision\markdownRendererUlItemEnd 
\markdownRendererUlItem \markdownRendererStrongEmphasis{Doorbell Camera}: vision, audio, location\markdownRendererUlItemEnd 
\markdownRendererUlItem \markdownRendererStrongEmphasis{Indoor Camera}: vision, audio\markdownRendererUlItemEnd 
\markdownRendererUlItem \markdownRendererStrongEmphasis{Smart Lock}: location\markdownRendererUlItemEnd 
\markdownRendererUlItem \markdownRendererStrongEmphasis{Motion Sensor}: motion\markdownRendererUlItemEnd 
\markdownRendererUlItem \markdownRendererStrongEmphasis{Door/Window Sensor}: device\markdownRendererUnderscore{}status\markdownRendererUlItemEnd 
\markdownRendererUlItem \markdownRendererStrongEmphasis{Smart Thermostat}: environment\markdownRendererUlItemEnd 
\markdownRendererUlItem \markdownRendererStrongEmphasis{Air Quality Monitor}: environment\markdownRendererUlItemEnd 
\markdownRendererUlItem \markdownRendererStrongEmphasis{Air Purifier}: environment, device\markdownRendererUnderscore{}status\markdownRendererUlItemEnd 
\markdownRendererUlItem \markdownRendererStrongEmphasis{Smart TV}: app\markdownRendererUnderscore{}usage, device\markdownRendererUnderscore{}status\markdownRendererUlItemEnd 
\markdownRendererUlItem \markdownRendererStrongEmphasis{Smart Fridge}: vision, device\markdownRendererUnderscore{}status\markdownRendererUlItemEnd 
\markdownRendererUlItem \markdownRendererStrongEmphasis{Smart Washer/Dryer}: device\markdownRendererUnderscore{}status\markdownRendererUlItemEnd 
\markdownRendererUlItem \markdownRendererStrongEmphasis{Robot Vacuum}: device\markdownRendererUnderscore{}status\markdownRendererUlItemEnd 
\markdownRendererUlItem \markdownRendererStrongEmphasis{Smart Light}: device\markdownRendererUnderscore{}status\markdownRendererUlItemEnd 
\markdownRendererUlItem \markdownRendererStrongEmphasis{Smart Curtain}: device\markdownRendererUnderscore{}status\markdownRendererUlItemEnd 
\markdownRendererUlItem \markdownRendererStrongEmphasis{Smart Scale}: health\markdownRendererUlItemEnd 
\markdownRendererUlItem \markdownRendererStrongEmphasis{Car System}: location, dialogue\markdownRendererUlItemEnd 
\markdownRendererUlItem \markdownRendererStrongEmphasis{Gaming Console}: app\markdownRendererUnderscore{}usage, device\markdownRendererUnderscore{}status\markdownRendererUlItemEnd 
\markdownRendererUlEnd \markdownRendererInterblockSeparator
{}\markdownRendererHeadingThree{Required Fields}\markdownRendererInterblockSeparator
{}\markdownRendererStrongEmphasis{Top-level persona fields}: \markdownRendererCodeSpan{persona\markdownRendererUnderscore{}id}, \markdownRendererCodeSpan{type}, \markdownRendererCodeSpan{characters}, \markdownRendererCodeSpan{shared\markdownRendererUnderscore{}devices}, \markdownRendererCodeSpan{all\markdownRendererUnderscore{}devices}\markdownRendererInterblockSeparator
{}\markdownRendererStrongEmphasis{Per-character fields}: \markdownRendererCodeSpan{name}, \markdownRendererCodeSpan{age}, \markdownRendererCodeSpan{role}, \markdownRendererCodeSpan{interests}, \markdownRendererCodeSpan{health}, \markdownRendererCodeSpan{routine}, \markdownRendererCodeSpan{devices}\markdownRendererInterblockSeparator
{}\markdownRendererStrongEmphasis{Routine sub-fields}: \markdownRendererCodeSpan{weekday}, \markdownRendererCodeSpan{weekend}\markdownRendererInterblockSeparator
{}\markdownRendererStrongEmphasis{Per-device fields in all\markdownRendererUnderscore{}devices}: \markdownRendererCodeSpan{device\markdownRendererUnderscore{}id}, \markdownRendererCodeSpan{device\markdownRendererUnderscore{}type}, \markdownRendererCodeSpan{owner}, \markdownRendererCodeSpan{location}, \markdownRendererCodeSpan{modality}\markdownRendererInterblockSeparator
{}\markdownRendererHeadingThree{Key Constraints}\markdownRendererInterblockSeparator
{}\markdownRendererStrongEmphasis{1.} \markdownRendererStrongEmphasis{Routine Time Alignment}: Shared events in different characters' routines must have identical time points.\markdownRendererInterblockSeparator
{}They must also align with \markdownRendererCodeSpan{shared\markdownRendererUnderscore{}schedule}.\markdownRendererInterblockSeparator
{}\markdownRendererStrongEmphasis{2.} \markdownRendererStrongEmphasis{Routine Logical Conflicts}: A single character cannot be in two places at the same time.\markdownRendererInterblockSeparator
{}No time overlaps are allowed within one person's routine.\markdownRendererInterblockSeparator
{}\markdownRendererStrongEmphasis{3.} \markdownRendererStrongEmphasis{Routine Time Realism}: Cooking, commuting, and bathing need adequate time.\markdownRendererInterblockSeparator
{}No teleportation between locations.\markdownRendererInterblockSeparator
{}\markdownRendererStrongEmphasis{4.} \markdownRendererStrongEmphasis{Device Consistency}: \markdownRendererCodeSpan{characters[].devices} union \markdownRendererCodeSpan{shared\markdownRendererUnderscore{}devices} must exactly equal the \markdownRendererCodeSpan{all\markdownRendererUnderscore{}devices} device\markdownRendererUnderscore{}id set.\markdownRendererInterblockSeparator
{}No orphans and no missing devices.\markdownRendererInterblockSeparator
{}\markdownRendererStrongEmphasis{5.} \markdownRendererStrongEmphasis{Modality Compliance}: Every device modality value must be one of the 9 standard types only.\markdownRendererInterblockSeparator
{}No typos or non-standard values.\markdownRendererInterblockSeparator
{}\markdownRendererStrongEmphasis{6.} \markdownRendererStrongEmphasis{Device Modality Reasonableness}: Each device's modality must match the Device Capability Reference Table.\markdownRendererInterblockSeparator
{}No missing critical capabilities and no unreasonable assignments.\linebreak
{}\markdownRendererStrongEmphasis{7.} \markdownRendererStrongEmphasis{Character Consistency}: Age, role, interests must match routine content (e.g., 5-year-old cannot drive)\linebreak
{}\markdownRendererStrongEmphasis{8.} \markdownRendererStrongEmphasis{Field Completeness}: All required fields must be present (see Required Fields section above)\linebreak
{}\markdownRendererStrongEmphasis{9.} \markdownRendererStrongEmphasis{Owner Consistency}: Personal devices \markdownRendererCodeSpan{owner} = character name; shared devices \markdownRendererCodeSpan{owner} = "shared".\markdownRendererInterblockSeparator
{}They must also cross-match with the \markdownRendererCodeSpan{characters[].devices} and \markdownRendererCodeSpan{shared\markdownRendererUnderscore{}devices} lists.\markdownRendererInterblockSeparator
{}\markdownRendererStrongEmphasis{10.} \markdownRendererStrongEmphasis{Device Location Reasonableness}: smartphone -> "carried", smartwatch -> "wearable", car system -> "car".\markdownRendererInterblockSeparator
{}Shared devices should be in household locations, with no contradictions to device\markdownRendererUnderscore{}id naming.\markdownRendererInterblockSeparator
{}\markdownRendererStrongEmphasis{11.} \markdownRendererStrongEmphasis{Routine-Device Support}: If a routine mentions an activity requiring a device (e.g., "play video games"), that device must exist in the persona.\linebreak
{}\markdownRendererStrongEmphasis{12.} \markdownRendererStrongEmphasis{Interests Count}: Each character must have 2-4 interests\linebreak
{}\markdownRendererStrongEmphasis{13.} \markdownRendererStrongEmphasis{Name Uniqueness}: Character names unique within persona; persona\markdownRendererUnderscore{}id globally unique across all personas\linebreak
{}\markdownRendererStrongEmphasis{14.} \markdownRendererStrongEmphasis{Shared Schedule Name Consistency}: All names in \markdownRendererCodeSpan{shared\markdownRendererUnderscore{}schedule.participants} must exactly match \markdownRendererCodeSpan{characters[].name} - no spelling differences or case mismatches\linebreak
{}\markdownRendererStrongEmphasis{15.} \markdownRendererStrongEmphasis{Device ID Naming Convention}: Personal devices follow \markdownRendererCodeSpan{\markdownRendererLeftBrace{}type\markdownRendererRightBrace{}\markdownRendererUnderscore{}\markdownRendererLeftBrace{}owner\markdownRendererUnderscore{}lowercase\markdownRendererRightBrace{}} format (e.g., \markdownRendererCodeSpan{watch\markdownRendererUnderscore{}david}); shared devices follow \markdownRendererCodeSpan{\markdownRendererLeftBrace{}type\markdownRendererRightBrace{}\markdownRendererUnderscore{}\markdownRendererLeftBrace{}location\markdownRendererRightBrace{}} format (e.g., \markdownRendererCodeSpan{speaker\markdownRendererUnderscore{}living})\linebreak
{}\markdownRendererStrongEmphasis{16.} \markdownRendererStrongEmphasis{Child Device Age-Appropriateness}: Children under 6 should not have smartphones/laptops; children's devices should be age-appropriate (kids smartwatch, tablet)\markdownRendererInterblockSeparator
{}\markdownRendererHeadingThree{Fix Workflow}\markdownRendererInterblockSeparator
{}\markdownRendererHeadingThree{Step 1: Independent Verification}\markdownRendererInterblockSeparator
{}\markdownRendererUlBegin
\markdownRendererUlItem For EACH issue in the review report, independently verify whether it actually exists by reading the data\markdownRendererUlItemEnd 
\markdownRendererUlItem Do NOT blindly trust the review report - some findings may be false positives\markdownRendererUlItemEnd 
\markdownRendererUlItem Mark each issue as CONFIRMED or REJECTED with brief reasoning\markdownRendererUlItemEnd 
\markdownRendererUlEnd \markdownRendererInterblockSeparator
{}\markdownRendererHeadingThree{Step 2: Prioritized Fixing}\markdownRendererInterblockSeparator
{}Fix confirmed issues by priority:\markdownRendererInterblockSeparator
{}\markdownRendererUlBegin
\markdownRendererUlItem \markdownRendererStrongEmphasis{HIGH}: Must fix\markdownRendererUlItemEnd 
\markdownRendererUlItem \markdownRendererStrongEmphasis{MEDIUM}: Should fix\markdownRendererUlItemEnd 
\markdownRendererUlItem \markdownRendererStrongEmphasis{LOW}: Fix if straightforward, skip if it would require major restructuring\markdownRendererUlItemEnd 
\markdownRendererUlEnd \markdownRendererInterblockSeparator
{}\markdownRendererHeadingThree{Step 3: Fix Principles}\markdownRendererInterblockSeparator
{}\markdownRendererStrongEmphasis{Routine Time Alignment Issues:}\markdownRendererInterblockSeparator
{}\markdownRendererUlBegin
\markdownRendererUlItem Use \markdownRendererCodeSpan{shared\markdownRendererUnderscore{}schedule} as the ground truth\markdownRendererUlItemEnd 
\markdownRendererUlItem Adjust inconsistent character routines to match shared\markdownRendererUnderscore{}schedule times\markdownRendererUlItemEnd 
\markdownRendererUlItem Maintain the overall flow and logic of individual routines while fixing alignment\markdownRendererUlItemEnd 
\markdownRendererUlEnd \markdownRendererInterblockSeparator
{}\markdownRendererStrongEmphasis{Device Consistency Issues:}\markdownRendererInterblockSeparator
{}\markdownRendererUlBegin
\markdownRendererUlItem If a device is in \markdownRendererCodeSpan{all\markdownRendererUnderscore{}devices} but not referenced: add to \markdownRendererCodeSpan{shared\markdownRendererUnderscore{}devices}\markdownRendererUlItemEnd 
\markdownRendererUlItem If a device is referenced but not in \markdownRendererCodeSpan{all\markdownRendererUnderscore{}devices}: remove the reference OR add the device definition (choose whichever makes more sense)\markdownRendererUlItemEnd 
\markdownRendererUlItem Ensure the union of all characters' devices + shared\markdownRendererUnderscore{}devices = all\markdownRendererUnderscore{}devices device\markdownRendererUnderscore{}ids\markdownRendererUlItemEnd 
\markdownRendererUlEnd \markdownRendererInterblockSeparator
{}\markdownRendererStrongEmphasis{Owner Consistency Issues:}\markdownRendererInterblockSeparator
{}\markdownRendererUlBegin
\markdownRendererUlItem Personal devices (in a character's \markdownRendererCodeSpan{devices} list): set owner to that character's name\markdownRendererUlItemEnd 
\markdownRendererUlItem Shared devices (in \markdownRendererCodeSpan{shared\markdownRendererUnderscore{}devices} list): set owner to "shared"\markdownRendererUlItemEnd 
\markdownRendererUlEnd \markdownRendererInterblockSeparator
{}\markdownRendererStrongEmphasis{Modality Issues:}\markdownRendererInterblockSeparator
{}\markdownRendererUlBegin
\markdownRendererUlItem Fix according to the Device Capability Reference Table above\markdownRendererUlItemEnd 
\markdownRendererUlItem If a device has an invalid modality value, replace with the correct one\markdownRendererUlItemEnd 
\markdownRendererUlItem If a device is missing a critical capability, add it\markdownRendererUlItemEnd 
\markdownRendererUlItem If a device has an unreasonable capability, remove it\markdownRendererUlItemEnd 
\markdownRendererUlEnd \markdownRendererInterblockSeparator
{}\markdownRendererStrongEmphasis{Time Conflict Issues:}\markdownRendererInterblockSeparator
{}\markdownRendererUlBegin
\markdownRendererUlItem Adjust to create a realistic, non-overlapping schedule\markdownRendererUlItemEnd 
\markdownRendererUlItem Maintain consistency with shared events\markdownRendererUlItemEnd 
\markdownRendererUlItem Allow reasonable transition time between activities\markdownRendererUlItemEnd 
\markdownRendererUlEnd \markdownRendererInterblockSeparator
{}\markdownRendererStrongEmphasis{Device ID Naming Issues:}\markdownRendererInterblockSeparator
{}\markdownRendererUlBegin
\markdownRendererUlItem Rename to follow convention: personal = \markdownRendererCodeSpan{\markdownRendererLeftBrace{}type\markdownRendererRightBrace{}\markdownRendererUnderscore{}\markdownRendererLeftBrace{}owner\markdownRendererUnderscore{}lowercase\markdownRendererRightBrace{}}, shared = \markdownRendererCodeSpan{\markdownRendererLeftBrace{}type\markdownRendererRightBrace{}\markdownRendererUnderscore{}\markdownRendererLeftBrace{}location\markdownRendererRightBrace{}}\markdownRendererUlItemEnd 
\markdownRendererUlItem Update ALL references (characters[].devices, shared\markdownRendererUnderscore{}devices, all\markdownRendererUnderscore{}devices) when renaming\markdownRendererUlItemEnd 
\markdownRendererUlEnd \markdownRendererInterblockSeparator
{}\markdownRendererStrongEmphasis{Interests Count Issues:}\markdownRendererInterblockSeparator
{}\markdownRendererUlBegin
\markdownRendererUlItem If fewer than 2: add plausible interests consistent with character's role/age\markdownRendererUlItemEnd 
\markdownRendererUlItem If more than 4: trim to the 4 most relevant\markdownRendererUlItemEnd 
\markdownRendererUlEnd \markdownRendererInterblockSeparator
{}\markdownRendererStrongEmphasis{Character Consistency Issues:}\markdownRendererInterblockSeparator
{}\markdownRendererUlBegin
\markdownRendererUlItem If routine contains activities contradicting age/role (e.g., child driving), remove or replace with age-appropriate activity\markdownRendererUlItemEnd 
\markdownRendererUlItem If interests don't match routine activities, adjust interests to align\markdownRendererUlItemEnd 
\markdownRendererUlEnd \markdownRendererInterblockSeparator
{}\markdownRendererStrongEmphasis{Field Completeness Issues:}\markdownRendererInterblockSeparator
{}\markdownRendererUlBegin
\markdownRendererUlItem Add missing required fields with reasonable default values\markdownRendererUlItemEnd 
\markdownRendererUlItem For missing \markdownRendererCodeSpan{health}: add "Healthy" or a brief condition consistent with the character\markdownRendererUlItemEnd 
\markdownRendererUlItem For missing \markdownRendererCodeSpan{interests}: infer 2-4 from role and routine content\markdownRendererUlItemEnd 
\markdownRendererUlItem For missing device fields (\markdownRendererCodeSpan{owner}, \markdownRendererCodeSpan{location}): infer from device\markdownRendererUnderscore{}id naming and context\markdownRendererUlItemEnd 
\markdownRendererUlEnd \markdownRendererInterblockSeparator
{}\markdownRendererStrongEmphasis{Device Location Reasonableness Issues:}\markdownRendererInterblockSeparator
{}\markdownRendererUlBegin
\markdownRendererUlItem Fix location to match device type: smartphone -> "carried", smartwatch -> "wearable", car system -> "car", laptop -> "home\markdownRendererUnderscore{}office"\markdownRendererUlItemEnd 
\markdownRendererUlItem Shared household devices: use actual room name (living\markdownRendererUnderscore{}room, kitchen, front\markdownRendererUnderscore{}door, etc.)\markdownRendererUlItemEnd 
\markdownRendererUlEnd \markdownRendererInterblockSeparator
{}\markdownRendererStrongEmphasis{Routine-Device Support Issues:}\markdownRendererInterblockSeparator
{}\markdownRendererUlBegin
\markdownRendererUlItem If routine mentions an activity but no supporting device exists: add the device to \markdownRendererCodeSpan{all\markdownRendererUnderscore{}devices} and \markdownRendererCodeSpan{shared\markdownRendererUnderscore{}devices} (or character's devices if personal)\markdownRendererUlItemEnd 
\markdownRendererUlItem Alternatively, if adding a device is too disruptive, rephrase the routine activity to not require the missing device\markdownRendererUlItemEnd 
\markdownRendererUlEnd \markdownRendererInterblockSeparator
{}\markdownRendererStrongEmphasis{Name Uniqueness Issues:}\markdownRendererInterblockSeparator
{}\markdownRendererUlBegin
\markdownRendererUlItem If character names are duplicated within a persona: rename one with a distinct but plausible name, update ALL references (shared\markdownRendererUnderscore{}schedule participants, device owners, device\markdownRendererUnderscore{}ids)\markdownRendererUlItemEnd 
\markdownRendererUlItem If persona\markdownRendererUnderscore{}ids are duplicated across personas: append a distinguishing suffix (e.g., \markdownRendererCodeSpan{\markdownRendererUnderscore{}002})\markdownRendererUlItemEnd 
\markdownRendererUlEnd \markdownRendererInterblockSeparator
{}\markdownRendererStrongEmphasis{Shared Schedule Name Consistency Issues:}\markdownRendererInterblockSeparator
{}\markdownRendererUlBegin
\markdownRendererUlItem If a name in shared\markdownRendererUnderscore{}schedule.participants doesn't match any character name: fix the spelling in shared\markdownRendererUnderscore{}schedule to match the character's actual name\markdownRendererUlItemEnd 
\markdownRendererUlItem Do NOT rename characters to match shared\markdownRendererUnderscore{}schedule - shared\markdownRendererUnderscore{}schedule is reference material, characters are the source of truth for names\markdownRendererUlItemEnd 
\markdownRendererUlEnd \markdownRendererInterblockSeparator
{}\markdownRendererStrongEmphasis{Child Device Age-Appropriateness Issues:}\markdownRendererInterblockSeparator
{}\markdownRendererUlBegin
\markdownRendererUlItem If a child under 6 has a smartphone/laptop: replace with age-appropriate device (kids smartwatch or tablet)\markdownRendererUlItemEnd 
\markdownRendererUlItem Update device\markdownRendererUnderscore{}id, all\markdownRendererUnderscore{}devices entry, and character's devices list accordingly\markdownRendererUlItemEnd 
\markdownRendererUlEnd \markdownRendererInterblockSeparator
{}\markdownRendererHeadingThree{Step 4: Backup \markdownRendererAmpersand{} Write Back}\markdownRendererInterblockSeparator
{}\markdownRendererUlBegin
\markdownRendererUlItem \markdownRendererStrongEmphasis{Before making any changes}, copy the original file to a backup with loop number suffix:\markdownRendererUlItemEnd 
\markdownRendererUlItem Loop 1: \markdownRendererCodeSpan{step1\markdownRendererUnderscore{}personas\markdownRendererUnderscore{}loop1.json}\markdownRendererUlItemEnd 
\markdownRendererUlItem Loop 2: \markdownRendererCodeSpan{step1\markdownRendererUnderscore{}personas\markdownRendererUnderscore{}loop2.json}\markdownRendererUlItemEnd 
\markdownRendererUlItem Loop N: \markdownRendererCodeSpan{step1\markdownRendererUnderscore{}personas\markdownRendererUnderscore{}loopN.json}\markdownRendererUlItemEnd 
\markdownRendererUlItem The loop number will be provided in the task prompt when you are invoked\markdownRendererUlItemEnd 
\markdownRendererUlItem The backup is placed in the same directory as the original file\markdownRendererUlItemEnd 
\markdownRendererUlItem Then write the fixed data back to the ORIGINAL file path (\markdownRendererCodeSpan{step1\markdownRendererUnderscore{}personas.json})\markdownRendererUlItemEnd 
\markdownRendererUlItem Ensure valid JSON format after all fixes\markdownRendererUlItemEnd 
\markdownRendererUlItem After writing, validate the JSON loads correctly\markdownRendererUlItemEnd 
\markdownRendererUlEnd \markdownRendererInterblockSeparator
{}\markdownRendererHeadingThree{Output Format}\markdownRendererInterblockSeparator
{}Generate fix report in the same directory. File naming by loop number:\markdownRendererInterblockSeparator
{}\markdownRendererUlBegin
\markdownRendererUlItem Loop 1: \markdownRendererCodeSpan{step1\markdownRendererUnderscore{}fix\markdownRendererUnderscore{}report\markdownRendererUnderscore{}loop1.md}\markdownRendererUlItemEnd 
\markdownRendererUlItem Loop 2: \markdownRendererCodeSpan{step1\markdownRendererUnderscore{}fix\markdownRendererUnderscore{}report\markdownRendererUnderscore{}loop2.md}\markdownRendererUlItemEnd 
\markdownRendererUlItem Loop N: \markdownRendererCodeSpan{step1\markdownRendererUnderscore{}fix\markdownRendererUnderscore{}report\markdownRendererUnderscore{}loopN.md}\markdownRendererUlItemEnd 
\markdownRendererUlEnd \markdownRendererInterblockSeparator
{}The loop number will be provided in the task prompt when you are invoked.\markdownRendererInterblockSeparator
{}Format:\markdownRendererInterblockSeparator
{}\markdownRendererInputFencedCode{prompt_tex/9c84d564fae707c9519eec7a2455ec23.verbatim}{text}\markdownRendererInterblockSeparator
{}\markdownRendererInputFencedCode{prompt_tex/86acfe9f29c7d5b07ece28aa4866cbd6.verbatim}{text}\markdownRendererInterblockSeparator
{}\markdownRendererInputFencedCode{prompt_tex/45e08a52077e0f4e67bc353283e7acc6.verbatim}{text}\markdownRendererInterblockSeparator
{}\markdownRendererInputFencedCode{prompt_tex/fa7692c0e7c321e2f6afb813ebbe09cb.verbatim}{text}\markdownRendererInterblockSeparator
{}\markdownRendererInputFencedCode{prompt_tex/6b8e08cf10d43183d83d9745e5315c8b.verbatim}{text}\markdownRendererInterblockSeparator
{}\markdownRendererUlBegin
\markdownRendererUlItem \markdownRendererStrongEmphasis{1}: [issue description]; Verification: CONFIRMED/REJECTED; Action: [fix description or rejection reason]; Status: FIXED/REJECTED\markdownRendererUlItemEnd 
\markdownRendererUlEnd \markdownRendererInterblockSeparator
{}\markdownRendererInputFencedCode{prompt_tex/0125f8c291549e0c0160686b0f6396de.verbatim}{text}\markdownRendererInterblockSeparator
{}\markdownRendererUlBegin
\markdownRendererUlItem \markdownRendererStrongEmphasis{1}: [issue description]; Verification: CONFIRMED/REJECTED; Action: [fix description]; Status: FIXED/REJECTED\markdownRendererUlItemEnd 
\markdownRendererUlEnd \markdownRendererInterblockSeparator
{}\markdownRendererHeadingThree{Important Notes}\markdownRendererInterblockSeparator
{}\markdownRendererUlBegin
\markdownRendererUlItem ALWAYS verify before fixing - never trust the review report blindly\markdownRendererUlItemEnd 
\markdownRendererUlItem Keep fixes minimal and targeted - do not rewrite entire routines unnecessarily\markdownRendererUlItemEnd 
\markdownRendererUlItem Maintain internal consistency after fixes (fixing one thing should not break another)\markdownRendererUlItemEnd 
\markdownRendererUlItem The file is large - use careful, targeted edits rather than full rewrites when possible\markdownRendererUlItemEnd 
\markdownRendererUlItem After all fixes, verify the JSON is still valid\markdownRendererUlItemEnd 
\markdownRendererUlEnd \markdownRendererDocumentEnd\relax
\end{promptbox}

\paragraph{Stage 6 reviewer system prompt.}
This prompt checks whether adversarial noise is realistic, non-evidence-bearing, and schema-consistent.
\begin{promptbox}{Stage 6 Reviewer Prompt}
\markdownRendererDocumentBegin
You are reviewing adversarial noise events in a benchmark dataset.\markdownRendererInterblockSeparator
{}The dataset evaluates a memory retrieval system's ability to distinguish relevant evidence from distractors.\markdownRendererInterblockSeparator
{}\markdownRendererHeadingThree{Background}\markdownRendererInterblockSeparator
{}In Stage\markdownRendererTilde{}6, adversarial noise events were generated for each target question. These events are intended to be:\linebreak
{}\markdownRendererStrongEmphasis{1.} Semantically similar to the target question's topic, while remaining non-evidence distractors\linebreak
{}\markdownRendererStrongEmphasis{2.} But must not be valid evidence for answering the target question\linebreak
{}\markdownRendererStrongEmphasis{3.} Must not serve as valid evidence for any other question\markdownRendererInterblockSeparator
{}\markdownRendererHeadingThree{What You Are Reviewing}\markdownRendererInterblockSeparator
{}Data File:\markdownRendererInterblockSeparator
{}\markdownRendererInputFencedCode{prompt_tex/034692dc6c22bcd167e0226ac2466c73.verbatim}{text}\markdownRendererInterblockSeparator
{}Structure:\markdownRendererInterblockSeparator
{}\markdownRendererInputFencedCode{prompt_tex/c239ae7936e185e538a7fc5ca99e9359.verbatim}{json}\markdownRendererInterblockSeparator
{}\markdownRendererUlBegin
\markdownRendererUlItem Only review events where \markdownRendererCodeSpan{source} == \markdownRendererCodeSpan{"adversarial"}\markdownRendererUlItemEnd 
\markdownRendererUlItem Each adversarial event has a \markdownRendererCodeSpan{target\markdownRendererUnderscore{}question} field linking it to a question in the same scenario's \markdownRendererCodeSpan{questions[]} array.\markdownRendererUlItemEnd 
\markdownRendererUlItem Use \markdownRendererCodeSpan{persona} within the same \markdownRendererCodeSpan{scenario\markdownRendererUnderscore{}data} entry for cross-referencing device/character info\markdownRendererUlItemEnd 
\markdownRendererUlEnd \markdownRendererInterblockSeparator
{}\markdownRendererHeadingThree{Reference Information}\markdownRendererInterblockSeparator
{}\markdownRendererHeadingThree{Date-Weekday Mapping (for this dataset)}\markdownRendererInterblockSeparator
{}\markdownRendererInputFencedCode{prompt_tex/d0bfcd7b08a6ddcfd70c13ded150612f.verbatim}{text}\markdownRendererInterblockSeparator
{}\markdownRendererHeadingThree{Valid Modalities (9 types only)}\markdownRendererInterblockSeparator
{}\markdownRendererUlBegin
\markdownRendererUlItem \markdownRendererStrongEmphasis{health}: physiological signs (heart rate, SpO2, sleep stages, etc.)\markdownRendererUlItemEnd 
\markdownRendererUlItem \markdownRendererStrongEmphasis{motion}: movement/activity (steps, exercise, fall detection, etc.)\markdownRendererUlItemEnd 
\markdownRendererUlItem \markdownRendererStrongEmphasis{vision}: visual perception (face recognition, motion detection, etc.)\markdownRendererUlItemEnd 
\markdownRendererUlItem \markdownRendererStrongEmphasis{audio}: audio perception (conversation, ambient sound, intercom)\markdownRendererUlItemEnd 
\markdownRendererUlItem \markdownRendererStrongEmphasis{environment}: environmental data (temperature, humidity, PM2.5, etc.)\markdownRendererUlItemEnd 
\markdownRendererUlItem \markdownRendererStrongEmphasis{location}: position \markdownRendererAmpersand{} access (GPS, entry/exit, unlock identity, etc.)\markdownRendererUlItemEnd 
\markdownRendererUlItem \markdownRendererStrongEmphasis{app\markdownRendererUnderscore{}usage}: digital behavior (screen time, app usage, viewing content)\markdownRendererUlItemEnd 
\markdownRendererUlItem \markdownRendererStrongEmphasis{dialogue}: conversational memory (user's stated intentions, preferences, plans)\markdownRendererUlItemEnd 
\markdownRendererUlItem \markdownRendererStrongEmphasis{device\markdownRendererUnderscore{}status}: device state change (on/off, mode changes, fault alarms)\markdownRendererUlItemEnd 
\markdownRendererUlEnd \markdownRendererInterblockSeparator
{}\markdownRendererStrongEmphasis{Key distinctions}:\markdownRendererInterblockSeparator
{}\markdownRendererUlBegin
\markdownRendererUlItem audio vs dialogue: audio is raw heard content; dialogue is semantic memory extracted from conversation\markdownRendererUlItemEnd 
\markdownRendererUlItem health vs motion: health = physiological metrics; motion = body activity and movement\markdownRendererUlItemEnd 
\markdownRendererUlItem vision vs location: vision = "what was seen"; location = "where a person/thing is"\markdownRendererUlItemEnd 
\markdownRendererUlItem environment vs device\markdownRendererUnderscore{}status:\markdownRendererInterblockSeparator
{}\markdownRendererUlBegin
\markdownRendererUlItem environment = physical quantities (temp, humidity, air quality)\markdownRendererUlItemEnd 
\markdownRendererUlItem device\markdownRendererUnderscore{}status = device's own operational state changes\markdownRendererUlItemEnd 
\markdownRendererUlEnd \markdownRendererUlItemEnd 
\markdownRendererUlEnd \markdownRendererInterblockSeparator
{}\markdownRendererHeadingThree{Device Capability Reference (expected modalities per device type)}\markdownRendererInterblockSeparator
{}\markdownRendererUlBegin
\markdownRendererUlItem \markdownRendererStrongEmphasis{Smartwatch/Band}: health, motion, location\markdownRendererUlItemEnd 
\markdownRendererUlItem \markdownRendererStrongEmphasis{Kids Smartwatch}: health, motion, location\markdownRendererUlItemEnd 
\markdownRendererUlItem \markdownRendererStrongEmphasis{Smartphone}: location, app\markdownRendererUnderscore{}usage, dialogue\markdownRendererUlItemEnd 
\markdownRendererUlItem \markdownRendererStrongEmphasis{Laptop}: app\markdownRendererUnderscore{}usage\markdownRendererUlItemEnd 
\markdownRendererUlItem \markdownRendererStrongEmphasis{Tablet}: app\markdownRendererUnderscore{}usage\markdownRendererUlItemEnd 
\markdownRendererUlItem \markdownRendererStrongEmphasis{Smart Speaker}: audio, dialogue\markdownRendererUlItemEnd 
\markdownRendererUlItem \markdownRendererStrongEmphasis{Smart Display (Speaker w/ Screen)}: audio, dialogue, vision\markdownRendererUlItemEnd 
\markdownRendererUlItem \markdownRendererStrongEmphasis{Doorbell Camera}: vision, audio, location\markdownRendererUlItemEnd 
\markdownRendererUlItem \markdownRendererStrongEmphasis{Indoor Camera}: vision, audio\markdownRendererUlItemEnd 
\markdownRendererUlItem \markdownRendererStrongEmphasis{Smart Lock}: location\markdownRendererUlItemEnd 
\markdownRendererUlItem \markdownRendererStrongEmphasis{Motion Sensor}: motion\markdownRendererUlItemEnd 
\markdownRendererUlItem \markdownRendererStrongEmphasis{Door/Window Sensor}: device\markdownRendererUnderscore{}status\markdownRendererUlItemEnd 
\markdownRendererUlItem \markdownRendererStrongEmphasis{Smart Thermostat}: environment\markdownRendererUlItemEnd 
\markdownRendererUlItem \markdownRendererStrongEmphasis{Air Quality Monitor}: environment\markdownRendererUlItemEnd 
\markdownRendererUlItem \markdownRendererStrongEmphasis{Air Purifier}: environment, device\markdownRendererUnderscore{}status\markdownRendererUlItemEnd 
\markdownRendererUlItem \markdownRendererStrongEmphasis{Smart TV}: app\markdownRendererUnderscore{}usage, device\markdownRendererUnderscore{}status\markdownRendererUlItemEnd 
\markdownRendererUlItem \markdownRendererStrongEmphasis{Smart Fridge}: vision, device\markdownRendererUnderscore{}status\markdownRendererUlItemEnd 
\markdownRendererUlItem \markdownRendererStrongEmphasis{Smart Washer/Dryer}: device\markdownRendererUnderscore{}status\markdownRendererUlItemEnd 
\markdownRendererUlItem \markdownRendererStrongEmphasis{Robot Vacuum}: device\markdownRendererUnderscore{}status\markdownRendererUlItemEnd 
\markdownRendererUlItem \markdownRendererStrongEmphasis{Smart Light}: device\markdownRendererUnderscore{}status\markdownRendererUlItemEnd 
\markdownRendererUlItem \markdownRendererStrongEmphasis{Smart Curtain}: device\markdownRendererUnderscore{}status\markdownRendererUlItemEnd 
\markdownRendererUlItem \markdownRendererStrongEmphasis{Smart Scale}: health\markdownRendererUlItemEnd 
\markdownRendererUlItem \markdownRendererStrongEmphasis{Car System}: location, dialogue\markdownRendererUlItemEnd 
\markdownRendererUlItem \markdownRendererStrongEmphasis{Gaming Console}: app\markdownRendererUnderscore{}usage, device\markdownRendererUnderscore{}status\markdownRendererUlItemEnd 
\markdownRendererUlEnd \markdownRendererInterblockSeparator
{}\markdownRendererHeadingThree{Adversarial Event Fields}\markdownRendererInterblockSeparator
{}Each adversarial event has exactly 8 fields:\markdownRendererInterblockSeparator
{}\markdownRendererUlBegin
\markdownRendererUlItem \markdownRendererCodeSpan{event\markdownRendererUnderscore{}id}: do not review; assigned by pipeline with format \markdownRendererCodeSpan{\markdownRendererLeftBrace{}episode\markdownRendererUnderscore{}id\markdownRendererRightBrace{}\markdownRendererUnderscore{}e\markdownRendererLeftBrace{}N\markdownRendererRightBrace{}}.\markdownRendererUlItemEnd 
\markdownRendererUlItem \markdownRendererCodeSpan{device}: review; must exist in persona's \markdownRendererCodeSpan{all\markdownRendererUnderscore{}devices[].device\markdownRendererUnderscore{}id}.\markdownRendererUlItemEnd 
\markdownRendererUlItem \markdownRendererCodeSpan{modality}: review; must be one of 9 valid types and compatible with the device.\markdownRendererUlItemEnd 
\markdownRendererUlItem \markdownRendererCodeSpan{description}: review; primary target for evidence leakage, content validity, and adversarial quality.\markdownRendererUlItemEnd 
\markdownRendererUlItem \markdownRendererCodeSpan{timestamp}: review; must use ISO format \markdownRendererCodeSpan{YYYY-MM-DDTHH:mm:ss}, valid date, and plausible time-of-day.\markdownRendererUlItemEnd 
\markdownRendererUlItem \markdownRendererCodeSpan{location}: review; must be consistent with the device's installation location.\markdownRendererUlItemEnd 
\markdownRendererUlItem \markdownRendererCodeSpan{source}: do not review; always \markdownRendererCodeSpan{"adversarial"}.\markdownRendererUlItemEnd 
\markdownRendererUlItem \markdownRendererCodeSpan{target\markdownRendererUnderscore{}question}: do not review; fixed assignment used to look up the target question.\markdownRendererUlItemEnd 
\markdownRendererUlEnd \markdownRendererInterblockSeparator
{}\markdownRendererStrongEmphasis{Note}: Adversarial events intentionally do NOT have a \markdownRendererCodeSpan{characters} field (unlike other event sources).\markdownRendererInterblockSeparator
{}Do not flag its absence.\markdownRendererInterblockSeparator
{}\markdownRendererHeadingThree{Review Criteria}\markdownRendererInterblockSeparator
{}For each adversarial event, you must check FOUR dimensions:\markdownRendererInterblockSeparator
{}\markdownRendererUlBegin
\markdownRendererUlItem (A) Whether it would \markdownRendererStrongEmphasis{actually constitute valid evidence} for answering its \markdownRendererCodeSpan{target\markdownRendererUnderscore{}question}\markdownRendererUlItemEnd 
\markdownRendererUlItem (B) Whether it would \markdownRendererStrongEmphasis{actually constitute valid evidence} for answering \markdownRendererStrongEmphasis{any other question} in the same scenario\markdownRendererUlItemEnd 
\markdownRendererUlItem (C) Whether the event itself is \markdownRendererStrongEmphasis{internally valid} (content, logic, device-modality consistency)\markdownRendererUlItemEnd 
\markdownRendererUlItem (D) Whether it has \markdownRendererStrongEmphasis{sufficient adversarial quality} (actually tests discrimination ability)\markdownRendererUlItemEnd 
\markdownRendererUlEnd \markdownRendererInterblockSeparator
{}An adversarial event FAILS the review if:\markdownRendererInterblockSeparator
{}\markdownRendererHeadingThree{Dimension A \markdownRendererAmpersand{} B: Evidence Leakage}\markdownRendererInterblockSeparator
{}\markdownRendererStrongEmphasis{1.} \markdownRendererStrongEmphasis{Direct Answer Leakage (target)}: The event description directly reveals or strongly implies part of the answer to the target question.\markdownRendererInterblockSeparator
{}\markdownRendererUlBegin
\markdownRendererUlItem Example: If the question asks "Why did Sarah turn on the air purifier?" and the adversarial event says "Sarah turned on the air purifier because of high pollen", that's a direct leak.\markdownRendererUlItemEnd 
\markdownRendererUlEnd \markdownRendererInterblockSeparator
{}\markdownRendererStrongEmphasis{2.} \markdownRendererStrongEmphasis{Evidence Equivalence (target)}: The event provides the same type and quality of information as the actual evidence events for the target question.\markdownRendererInterblockSeparator
{}\markdownRendererUlBegin
\markdownRendererUlItem Example: If a question asks about health readings progression and the adversarial event contains specific health readings (HR, SpO2, etc.) at a time that would fill a gap in the progression, it becomes valid evidence.\markdownRendererUlItemEnd 
\markdownRendererUlEnd \markdownRendererInterblockSeparator
{}\markdownRendererStrongEmphasis{3.} \markdownRendererStrongEmphasis{Causal Chain Completion (target)}: The event completes a causal chain needed to answer the target question.\markdownRendererInterblockSeparator
{}\markdownRendererUlBegin
\markdownRendererUlItem Example: If the question asks "What happened after X?" and the adversarial event describes a consequence of X that matches the expected answer.\markdownRendererUlItemEnd 
\markdownRendererUlEnd \markdownRendererInterblockSeparator
{}\markdownRendererStrongEmphasis{4.} \markdownRendererStrongEmphasis{Temporal/Factual Contradiction with Evidence}: The event contradicts the actual evidence events or the ground-truth answer in a way that would confuse the system about the correct answer.\markdownRendererInterblockSeparator
{}\markdownRendererUlBegin
\markdownRendererUlItem Example: It may provide contradictory readings at the same time as real evidence events.\markdownRendererUlItemEnd 
\markdownRendererUlEnd \markdownRendererInterblockSeparator
{}\markdownRendererStrongEmphasis{5.} \markdownRendererStrongEmphasis{Cross-Question Evidence Leakage}: The event serves as valid evidence for a DIFFERENT question in the same scenario, not its target question.\markdownRendererInterblockSeparator
{}\markdownRendererUlBegin
\markdownRendererUlItem Example: If an adversarial event targeted at q12 describes "David cancelled the soccer reminder" and q17 asks about what David did after Ethan's injury, this event would be valid evidence for q17.\markdownRendererUlItemEnd 
\markdownRendererUlItem Check the event against ALL other questions in the scenario.\markdownRendererUlItemEnd 
\markdownRendererUlEnd \markdownRendererInterblockSeparator
{}\markdownRendererHeadingThree{Dimension C: Content/Logic Validity}\markdownRendererInterblockSeparator
{}\markdownRendererStrongEmphasis{6.} \markdownRendererStrongEmphasis{Device-Modality Incompatibility}: The event's \markdownRendererCodeSpan{modality} is not supported by the \markdownRendererCodeSpan{device}.\markdownRendererInterblockSeparator
{}\markdownRendererUlBegin
\markdownRendererUlItem Cross-reference the device in persona's \markdownRendererCodeSpan{all\markdownRendererUnderscore{}devices} and check its \markdownRendererCodeSpan{modality} field (comma-separated list).\markdownRendererUlItemEnd 
\markdownRendererUlItem Example: \markdownRendererCodeSpan{device: "watch\markdownRendererUnderscore{}david"} with \markdownRendererCodeSpan{modality: "audio"} - a smartwatch cannot produce audio events.\markdownRendererUlItemEnd 
\markdownRendererUlEnd \markdownRendererInterblockSeparator
{}\markdownRendererStrongEmphasis{7.} \markdownRendererStrongEmphasis{Device Non-Existence}: The \markdownRendererCodeSpan{device} value doesn't exist in the persona's \markdownRendererCodeSpan{all\markdownRendererUnderscore{}devices[].device\markdownRendererUnderscore{}id} list. Example: event uses \markdownRendererCodeSpan{"device": "camera\markdownRendererUnderscore{}garage"} but no such device exists in persona.\markdownRendererInterblockSeparator
{}\markdownRendererStrongEmphasis{8.} \markdownRendererStrongEmphasis{Modality-Description Mismatch}: The description content doesn't match what the stated modality can perceive.\markdownRendererInterblockSeparator
{}\markdownRendererUlBegin
\markdownRendererUlItem Refer to the "perceivable information" column in the modality table.\markdownRendererUlItemEnd 
\markdownRendererUlItem Example: \markdownRendererCodeSpan{modality: "health"} but description says "camera detected movement".\markdownRendererUlItemEnd 
\markdownRendererUlEnd \markdownRendererInterblockSeparator
{}\markdownRendererStrongEmphasis{9.} \markdownRendererStrongEmphasis{Impossible/Unrealistic Readings}: Physiological or environmental readings that are physically impossible. Examples:\markdownRendererInterblockSeparator
{}\markdownRendererUlBegin
\markdownRendererUlItem Heart rate: normal range 40-180 bpm (exercise up to 200)\markdownRendererUlItemEnd 
\markdownRendererUlItem SpO2: normal range 88-100\markdownRendererPercentSign{}\markdownRendererUlItemEnd 
\markdownRendererUlItem Body temperature: 35-40 deg C\markdownRendererUlItemEnd 
\markdownRendererUlItem Room temperature: 15-35 deg C\markdownRendererUlItemEnd 
\markdownRendererUlItem Humidity: 20-80\markdownRendererPercentSign{}\markdownRendererUlItemEnd 
\markdownRendererUlItem PM2.5: 0-500 ug/m\markdownRendererBackslash{}textasciicircum\markdownRendererLeftBrace{}\markdownRendererRightBrace{}3\markdownRendererUlItemEnd 
\markdownRendererUlItem Step count per hour: 0-10000\markdownRendererUlItemEnd 
\markdownRendererUlEnd \markdownRendererInterblockSeparator
{}\markdownRendererStrongEmphasis{10.} \markdownRendererStrongEmphasis{Location-Device Inconsistency}: The event's \markdownRendererCodeSpan{location} conflicts with where the device should be.\markdownRendererInterblockSeparator
{}\markdownRendererUlBegin
\markdownRendererUlItem Fixed devices (cameras, sensors) must match their installation location.\markdownRendererUlItemEnd 
\markdownRendererUlItem Wearable/carried devices follow the person.\markdownRendererUlItemEnd 
\markdownRendererUlEnd \markdownRendererInterblockSeparator
{}\markdownRendererStrongEmphasis{11.} \markdownRendererStrongEmphasis{Timestamp Implausibility}: Activity described at an impossible time. Examples:\markdownRendererInterblockSeparator
{}\markdownRendererUlBegin
\markdownRendererUlItem A child at school at 2 AM\markdownRendererUlItemEnd 
\markdownRendererUlItem Breakfast at 23:00\markdownRendererUlItemEnd 
\markdownRendererUlItem A person sleeping at noon without storyline justification\markdownRendererUlItemEnd 
\markdownRendererUlEnd \markdownRendererInterblockSeparator
{}\markdownRendererStrongEmphasis{12.} \markdownRendererStrongEmphasis{Character-Device Ownership Mismatch}: A wearable/personal device reports data about someone other than its owner. Example: \markdownRendererCodeSpan{watch\markdownRendererUnderscore{}david} reports Sarah's health data.\markdownRendererInterblockSeparator
{}\markdownRendererStrongEmphasis{13.} \markdownRendererStrongEmphasis{Language Error}: Description contains non-English text (this is an EN dataset) or garbled/incomplete text.\markdownRendererInterblockSeparator
{}\markdownRendererStrongEmphasis{14.} \markdownRendererStrongEmphasis{Non-Atomic Event}: The description spans multiple time points or combines multiple distinct activities into one event.\markdownRendererInterblockSeparator
{}Each event should describe a SINGLE instantaneous observation at ONE timestamp.\markdownRendererInterblockSeparator
{}\markdownRendererUlBegin
\markdownRendererUlItem Example violation: "Smart watch recorded that David ran 5km between 7:00 and 7:45, then had breakfast at 8:00" - combines two activities at different times\markdownRendererUlItemEnd 
\markdownRendererUlItem Example PASS: "Smart watch morning summary: total sleep 6.5h, deep sleep 2.1h, 3 awakenings" at 07:00 - single summary report at one timestamp is acceptable\markdownRendererUlItemEnd 
\markdownRendererUlItem Severity: MEDIUM\markdownRendererUlItemEnd 
\markdownRendererUlEnd \markdownRendererInterblockSeparator
{}\markdownRendererStrongEmphasis{15.} \markdownRendererStrongEmphasis{Omniscient Narrator Perspective}: The description uses an omniscient narrator viewpoint instead of the device's sensing perspective. Devices can only report what they can physically sense.\markdownRendererInterblockSeparator
{}\markdownRendererUlBegin
\markdownRendererUlItem Camera -> visual observations (what it captured/detected/recorded)\markdownRendererUlItemEnd 
\markdownRendererUlItem Watch -> physiological/motion data (what it measured/recorded)\markdownRendererUlItemEnd 
\markdownRendererUlItem Speaker -> audio content (what it heard/detected)\markdownRendererUlItemEnd 
\markdownRendererUlItem Sensor -> environmental readings (what it measured)\markdownRendererUlItemEnd 
\markdownRendererUlItem Example violation: "David felt anxious about the meeting" - a device cannot know feelings unless expressed\markdownRendererUlItemEnd 
\markdownRendererUlItem Example violation: "Sarah decided to skip yoga today" - internal decisions are not observable\markdownRendererUlItemEnd 
\markdownRendererUlItem Example PASS: "Smart watch recorded elevated heart rate of 95bpm, increased perspiration" - device-observable data\markdownRendererUlItemEnd 
\markdownRendererUlItem Severity: MEDIUM\markdownRendererUlItemEnd 
\markdownRendererUlEnd \markdownRendererInterblockSeparator
{}\markdownRendererStrongEmphasis{16.} \markdownRendererStrongEmphasis{Generic/Anonymous Name Usage}: The description uses generic terms like "user", "owner", "the person", "he/she" instead of character names.\markdownRendererInterblockSeparator
{}In a multi-person household, it must be clear WHO is being described.\markdownRendererInterblockSeparator
{}\markdownRendererUlBegin
\markdownRendererUlItem Example violation: "The user returned home at 6pm"\markdownRendererUlItemEnd 
\markdownRendererUlItem Example PASS: "David returned home at 6pm"\markdownRendererUlItemEnd 
\markdownRendererUlItem Severity: MEDIUM (LOW if it's a single-person device where the owner is obvious)\markdownRendererUlItemEnd 
\markdownRendererUlEnd \markdownRendererInterblockSeparator
{}\markdownRendererStrongEmphasis{17.} \markdownRendererStrongEmphasis{Timestamp Format Invalid}: The timestamp does not conform to \markdownRendererCodeSpan{YYYY-MM-DDTHH:mm:ss} format.\markdownRendererInterblockSeparator
{}It may also contain an invalid date/time (e.g., month 13, hour 25), or the date may fall outside the scenario's time span (2026-05-11 \markdownRendererBackslash{}textasciitilde\markdownRendererLeftBrace{}\markdownRendererRightBrace{} 2026-05-17).\markdownRendererInterblockSeparator
{}\markdownRendererUlBegin
\markdownRendererUlItem Severity: HIGH\markdownRendererUlItemEnd 
\markdownRendererUlEnd \markdownRendererInterblockSeparator
{}\markdownRendererStrongEmphasis{18.} \markdownRendererStrongEmphasis{Weekday-Activity Inconsistency}: The event describes an activity that contradicts the day of the week. Check against the date-weekday mapping above.\markdownRendererInterblockSeparator
{}\markdownRendererUlBegin
\markdownRendererUlItem Example violation: Event on 2026-05-16 (Saturday) describes "Ethan attending math class at school"\markdownRendererUlItemEnd 
\markdownRendererUlItem Example violation: Event on 2026-05-11 (Monday) describes "family weekend outing to the zoo"\markdownRendererUlItemEnd 
\markdownRendererUlItem Example PASS: Activities that are plausible on any day (cooking, watching TV, health readings) don't need weekday alignment\markdownRendererUlItemEnd 
\markdownRendererUlItem Severity: MEDIUM\markdownRendererUlItemEnd 
\markdownRendererUlEnd \markdownRendererInterblockSeparator
{}\markdownRendererStrongEmphasis{19.} \markdownRendererStrongEmphasis{Character-Action Age/Role Inconsistency}: The event describes a character performing an action that is impossible given their age, role, or health status as defined in the persona.\markdownRendererInterblockSeparator
{}\markdownRendererUlBegin
\markdownRendererUlItem Cross-reference: check the character's \markdownRendererCodeSpan{age}, \markdownRendererCodeSpan{role}, and \markdownRendererCodeSpan{health} fields in persona\markdownRendererUlItemEnd 
\markdownRendererUlItem Example violation: A 5-year-old child "drives to the pharmacy" - children cannot drive\markdownRendererUlItemEnd 
\markdownRendererUlItem Example violation: A 9-year-old "logs into their work laptop for a video conference" - children don't have work meetings\markdownRendererUlItemEnd 
\markdownRendererUlItem Example PASS: A 9-year-old "plays video games on the tablet" - age-appropriate\markdownRendererUlItemEnd 
\markdownRendererUlItem Severity: MEDIUM\markdownRendererUlItemEnd 
\markdownRendererUlEnd \markdownRendererInterblockSeparator
{}\markdownRendererStrongEmphasis{20.} \markdownRendererStrongEmphasis{Duplicate Device-Timestamp}: The adversarial event has the same \markdownRendererCodeSpan{device} AND \markdownRendererCodeSpan{timestamp} as another event (adversarial or non-adversarial) in the same episode.\markdownRendererInterblockSeparator
{}This creates an ambiguous data point.\markdownRendererInterblockSeparator
{}\markdownRendererUlBegin
\markdownRendererUlItem Severity: MEDIUM\markdownRendererUlItemEnd 
\markdownRendererUlEnd \markdownRendererInterblockSeparator
{}\markdownRendererHeadingThree{Dimension D: Adversarial Quality (Distractor Effectiveness)}\markdownRendererInterblockSeparator
{}\markdownRendererStrongEmphasis{21.} \markdownRendererStrongEmphasis{Topic Irrelevance}: The event has NO semantic connection to the target question's topic.\markdownRendererInterblockSeparator
{}It would be trivially easy for a retrieval system to exclude it, so it would not test discrimination ability at all.\markdownRendererInterblockSeparator
{}\markdownRendererUlBegin
\markdownRendererUlItem Check: Does the event share at least some topical overlap (same domain: health/location/activity/device type) with the target question?\markdownRendererUlItemEnd 
\markdownRendererUlItem Example violation: Target question asks "Why did Ethan's heart rate spike?" but adversarial event describes "Living room thermostat measured 23.5 deg C" - completely unrelated topic\markdownRendererUlItemEnd 
\markdownRendererUlItem Example PASS: Target question asks about health readings, adversarial event describes health readings from a different context - this IS topically related even though it's not evidence\markdownRendererUlItemEnd 
\markdownRendererUlItem Severity: MEDIUM\markdownRendererUlItemEnd 
\markdownRendererUlEnd \markdownRendererInterblockSeparator
{}\markdownRendererStrongEmphasis{22.} \markdownRendererStrongEmphasis{Too Vague/Generic}: The event description is so generic that it provides no specific information and would never confuse a retrieval system.\markdownRendererInterblockSeparator
{}Examples: "Normal readings detected", "Activity observed", "Everything is fine", "Device is working properly".\markdownRendererInterblockSeparator
{}\markdownRendererUlBegin
\markdownRendererUlItem Adversarial events must be SPECIFIC - they should contain concrete details (specific readings, named people, particular activities) to actually test discrimination ability\markdownRendererUlItemEnd 
\markdownRendererUlItem Severity: MEDIUM\markdownRendererUlItemEnd 
\markdownRendererUlEnd \markdownRendererInterblockSeparator
{}\markdownRendererHeadingThree{Review Passing Criteria}\markdownRendererInterblockSeparator
{}An adversarial event PASSES if:\markdownRendererInterblockSeparator
{}\markdownRendererUlBegin
\markdownRendererUlItem It is topically related but does NOT provide information that helps answer the target question OR any other question\markdownRendererUlItemEnd 
\markdownRendererUlItem It describes similar-type activities/readings but at irrelevant times or for irrelevant purposes\markdownRendererUlItemEnd 
\markdownRendererUlItem It is a genuine distractor that tests discrimination ability without leaking answer information to any question\markdownRendererUlItemEnd 
\markdownRendererUlItem Its content is internally consistent, realistic, and logically sound given the scenario context\markdownRendererUlItemEnd 
\markdownRendererUlItem Device, modality, and description are mutually consistent\markdownRendererUlItemEnd 
\markdownRendererUlEnd \markdownRendererInterblockSeparator
{}\markdownRendererHeadingThree{Review Process}\markdownRendererInterblockSeparator
{}\markdownRendererStrongEmphasis{1.} Read the full data file\linebreak
{}\markdownRendererStrongEmphasis{2.} For each scenario in \markdownRendererCodeSpan{scenario\markdownRendererUnderscore{}data}:\linebreak
{}a. Build a map of \markdownRendererCodeSpan{question\markdownRendererUnderscore{}id} -> question (with \markdownRendererCodeSpan{question}, \markdownRendererCodeSpan{answer}, \markdownRendererCodeSpan{evidence\markdownRendererUnderscore{}event\markdownRendererUnderscore{}ids})\linebreak
{}b. Build the full list of ALL questions in the scenario for cross-checking\linebreak
{}c. Extract the persona's \markdownRendererCodeSpan{all\markdownRendererUnderscore{}devices} for device/modality validation\linebreak
{}d. For each adversarial event (where \markdownRendererCodeSpan{source} == \markdownRendererCodeSpan{"adversarial"}):\markdownRendererInterblockSeparator
{}\markdownRendererUlBegin
\markdownRendererUlItem \markdownRendererStrongEmphasis{Content/Logic check (Dimension C)}:\markdownRendererInterblockSeparator
{}\markdownRendererUlBegin
\markdownRendererUlItem Validate device existence\markdownRendererUlItemEnd 
\markdownRendererUlItem Validate modality-device compatibility\markdownRendererUlItemEnd 
\markdownRendererUlItem Validate modality-description consistency\markdownRendererUlItemEnd 
\markdownRendererUlItem Validate reading plausibility\markdownRendererUlItemEnd 
\markdownRendererUlItem Validate location-device match\markdownRendererUlItemEnd 
\markdownRendererUlItem Validate timestamp plausibility\markdownRendererUlItemEnd 
\markdownRendererUlItem Validate character-device ownership\markdownRendererUlItemEnd 
\markdownRendererUlEnd \markdownRendererUlItemEnd 
\markdownRendererUlItem \markdownRendererStrongEmphasis{Target question check (Dimension A)}: Look up its \markdownRendererCodeSpan{target\markdownRendererUnderscore{}question}, then compare the event description against the question's \markdownRendererCodeSpan{question} and \markdownRendererCodeSpan{answer} text.\markdownRendererUlItemEnd 
\markdownRendererUlItem \markdownRendererStrongEmphasis{Cross-question check (Dimension B)}: Compare the event against ALL other questions in the same scenario.\markdownRendererUlItemEnd 
\markdownRendererUlEnd \markdownRendererInterblockSeparator
{}Check whether it could serve as valid evidence for any of them.\linebreak
{}\markdownRendererStrongEmphasis{3.} Document all failures with specific reasoning\markdownRendererInterblockSeparator
{}\markdownRendererHeadingThree{Output Format}\markdownRendererInterblockSeparator
{}Generate a review report as a markdown file. File naming by loop number:\markdownRendererInterblockSeparator
{}\markdownRendererUlBegin
\markdownRendererUlItem Loop 1: \markdownRendererCodeSpan{step6\markdownRendererUnderscore{}review\markdownRendererUnderscore{}report\markdownRendererUnderscore{}loop1.md}\markdownRendererUlItemEnd 
\markdownRendererUlItem Loop 2: \markdownRendererCodeSpan{step6\markdownRendererUnderscore{}review\markdownRendererUnderscore{}report\markdownRendererUnderscore{}loop2.md}\markdownRendererUlItemEnd 
\markdownRendererUlItem Loop N: \markdownRendererCodeSpan{step6\markdownRendererUnderscore{}review\markdownRendererUnderscore{}report\markdownRendererUnderscore{}loopN.md}\markdownRendererUlItemEnd 
\markdownRendererUlEnd \markdownRendererInterblockSeparator
{}The loop number will be provided in the task prompt when you are invoked.\markdownRendererInterblockSeparator
{}Format:\markdownRendererInterblockSeparator
{}\markdownRendererInputFencedCode{prompt_tex/097afdc98e0ba691d7535cef4fa07309.verbatim}{text}\markdownRendererInterblockSeparator
{}\markdownRendererInputFencedCode{prompt_tex/77727a630320e1ecc5d9f10edeccc65d.verbatim}{text}\markdownRendererInterblockSeparator
{}\markdownRendererInputFencedCode{prompt_tex/a9e0c713eee395c8f704d2370bb3ee44.verbatim}{text}\markdownRendererInterblockSeparator
{}\markdownRendererInputFencedCode{prompt_tex/371b475a7114a93c21d641900e77dd2d.verbatim}{text}\markdownRendererInterblockSeparator
{}\markdownRendererInputFencedCode{prompt_tex/40a0b34ff01bcbc67c8d63e0e3065b18.verbatim}{text}\markdownRendererInterblockSeparator
{}\markdownRendererInputFencedCode{prompt_tex/07efe2a712b6a8791f665bf907ae4b2f.verbatim}{text}\markdownRendererInterblockSeparator
{}\markdownRendererInputFencedCode{prompt_tex/00e1500a893bd6197611a5e1c0fbde5d.verbatim}{text}\markdownRendererInterblockSeparator
{}\markdownRendererInputFencedCode{prompt_tex/f8323807601287692aadc1e8373ae9e1.verbatim}{text}\markdownRendererInterblockSeparator
{}\markdownRendererInputFencedCode{prompt_tex/c9dd9e29094c2babae13e0f6967ba041.verbatim}{text}\markdownRendererInterblockSeparator
{}Each issue should be a separate section with all relevant fields clearly listed.\markdownRendererInterblockSeparator
{}\markdownRendererHeadingThree{Severity Guide}\markdownRendererInterblockSeparator
{}\markdownRendererUlBegin
\markdownRendererUlItem \markdownRendererStrongEmphasis{HIGH}:\markdownRendererInterblockSeparator
{}\markdownRendererUlBegin
\markdownRendererUlItem Direct answer leakage\markdownRendererUlItemEnd 
\markdownRendererUlItem Evidence equivalence\markdownRendererUlItemEnd 
\markdownRendererUlItem Device non-existence\markdownRendererUlItemEnd 
\markdownRendererUlItem Device-modality incompatibility\markdownRendererUlItemEnd 
\markdownRendererUlItem Modality-description mismatch\markdownRendererUlItemEnd 
\markdownRendererUlItem Language error\markdownRendererUlItemEnd 
\markdownRendererUlItem Timestamp format invalid\markdownRendererUlItemEnd 
\markdownRendererUlEnd \markdownRendererUlItemEnd 
\markdownRendererUlItem \markdownRendererStrongEmphasis{MEDIUM}:\markdownRendererInterblockSeparator
{}\markdownRendererUlBegin
\markdownRendererUlItem Causal chain completion\markdownRendererUlItemEnd 
\markdownRendererUlItem Cross-question evidence leakage\markdownRendererUlItemEnd 
\markdownRendererUlItem Temporal/factual contradiction\markdownRendererUlItemEnd 
\markdownRendererUlItem Impossible readings\markdownRendererUlItemEnd 
\markdownRendererUlItem Location-device inconsistency\markdownRendererUlItemEnd 
\markdownRendererUlItem Timestamp implausibility\markdownRendererUlItemEnd 
\markdownRendererUlItem Character-device ownership mismatch\markdownRendererUlItemEnd 
\markdownRendererUlItem Non-atomic event\markdownRendererUlItemEnd 
\markdownRendererUlItem Omniscient narrator perspective\markdownRendererUlItemEnd 
\markdownRendererUlItem Generic/anonymous name usage\markdownRendererUlItemEnd 
\markdownRendererUlItem Weekday-activity inconsistency\markdownRendererUlItemEnd 
\markdownRendererUlItem Character-action age/role inconsistency\markdownRendererUlItemEnd 
\markdownRendererUlItem Duplicate device-timestamp\markdownRendererUlItemEnd 
\markdownRendererUlItem Topic irrelevance\markdownRendererUlItemEnd 
\markdownRendererUlItem Too vague/generic\markdownRendererUlItemEnd 
\markdownRendererUlEnd \markdownRendererUlItemEnd 
\markdownRendererUlEnd \markdownRendererInterblockSeparator
{}\markdownRendererHeadingThree{Important Notes}\markdownRendererInterblockSeparator
{}\markdownRendererUlBegin
\markdownRendererUlItem Only flag REAL issues. An adversarial event being "somewhat related" to the topic is by design - that's what makes it adversarial. Only flag events that would genuinely serve as valid evidence or leak answer information.\markdownRendererUlItemEnd 
\markdownRendererUlItem For content/logic checks: look up the persona's \markdownRendererCodeSpan{all\markdownRendererUnderscore{}devices} to get each device's \markdownRendererCodeSpan{modality} and \markdownRendererCodeSpan{location} fields, then validate against the event.\markdownRendererUlItemEnd 
\markdownRendererUlItem Be thorough: check every adversarial event in every scenario.\markdownRendererUlItemEnd 
\markdownRendererUlItem Read the full answer text carefully - adversarial events might leak specific details mentioned in the answer.\markdownRendererUlItemEnd 
\markdownRendererUlItem When checking cross-question leakage, focus on questions whose topic clearly overlaps with the event content - you don't need to exhaustively compare against every single question if the topics are unrelated.\markdownRendererUlItemEnd 
\markdownRendererUlItem For device validation, use the \markdownRendererCodeSpan{persona} field within the same \markdownRendererCodeSpan{scenario\markdownRendererUnderscore{}data} entry (not the top-level \markdownRendererCodeSpan{personas} array).\markdownRendererUlItemEnd 
\markdownRendererUlEnd \markdownRendererDocumentEnd\relax
\end{promptbox}

\paragraph{Stage 6 corrector system prompt.}
This prompt repairs Stage~6 adversarial events while keeping them useful as distractors.
\begin{promptbox}{Stage 6 Corrector Prompt}
\markdownRendererDocumentBegin
You are correcting adversarial noise events in a benchmark dataset based on a review report.\markdownRendererInterblockSeparator
{}The dataset evaluates a memory retrieval system's ability to distinguish relevant evidence from distractors.\markdownRendererInterblockSeparator
{}\markdownRendererHeadingThree{Background}\markdownRendererInterblockSeparator
{}In Stage\markdownRendererTilde{}6, adversarial noise events were generated for each target question. These events are intended to be:\linebreak
{}\markdownRendererStrongEmphasis{1.} Semantically similar to the target question's topic, while remaining non-evidence distractors\linebreak
{}\markdownRendererStrongEmphasis{2.} \markdownRendererStrongEmphasis{But must not be valid evidence for answering the target question}\linebreak
{}\markdownRendererStrongEmphasis{3.} Must not serve as valid evidence for any other question\markdownRendererInterblockSeparator
{}A reviewer has identified events that FAIL these criteria.\markdownRendererInterblockSeparator
{}They may leak answer information, serve as valid evidence, have content/logic errors, or lack adversarial quality.\markdownRendererInterblockSeparator
{}\markdownRendererHeadingThree{Input Files}\markdownRendererInterblockSeparator
{}\markdownRendererUlBegin
\markdownRendererUlItem Data File\markdownRendererInterblockSeparator
{}\markdownRendererInputFencedCode{prompt_tex/034692dc6c22bcd167e0226ac2466c73.verbatim}{text}\markdownRendererUlItemEnd 
\markdownRendererUlItem Review Report File\markdownRendererInterblockSeparator
{}\markdownRendererInputFencedCode{prompt_tex/76a8b0104a5b1b9c7d5f7ef8c1bee884.verbatim}{text}\markdownRendererUlItemEnd 
\markdownRendererUlEnd \markdownRendererInterblockSeparator
{}The loop number will be provided in the task prompt when you are invoked.\markdownRendererInterblockSeparator
{}\markdownRendererHeadingThree{Reference Information}\markdownRendererInterblockSeparator
{}\markdownRendererHeadingThree{Date-Weekday Mapping (for this dataset)}\markdownRendererInterblockSeparator
{}\markdownRendererInputFencedCode{prompt_tex/d0bfcd7b08a6ddcfd70c13ded150612f.verbatim}{text}\markdownRendererInterblockSeparator
{}\markdownRendererHeadingThree{Valid Modalities (9 types only)}\markdownRendererInterblockSeparator
{}\markdownRendererUlBegin
\markdownRendererUlItem \markdownRendererStrongEmphasis{health}: physiological signs (heart rate, SpO2, sleep stages, etc.)\markdownRendererUlItemEnd 
\markdownRendererUlItem \markdownRendererStrongEmphasis{motion}: movement/activity (steps, exercise, fall detection, etc.)\markdownRendererUlItemEnd 
\markdownRendererUlItem \markdownRendererStrongEmphasis{vision}: visual perception (face recognition, motion detection, etc.)\markdownRendererUlItemEnd 
\markdownRendererUlItem \markdownRendererStrongEmphasis{audio}: audio perception (conversation, ambient sound, intercom)\markdownRendererUlItemEnd 
\markdownRendererUlItem \markdownRendererStrongEmphasis{environment}: environmental data (temperature, humidity, PM2.5, etc.)\markdownRendererUlItemEnd 
\markdownRendererUlItem \markdownRendererStrongEmphasis{location}: position \markdownRendererAmpersand{} access (GPS, entry/exit, unlock identity, etc.)\markdownRendererUlItemEnd 
\markdownRendererUlItem \markdownRendererStrongEmphasis{app\markdownRendererUnderscore{}usage}: digital behavior (screen time, app usage, viewing content)\markdownRendererUlItemEnd 
\markdownRendererUlItem \markdownRendererStrongEmphasis{dialogue}: conversational memory (user's stated intentions, preferences, plans)\markdownRendererUlItemEnd 
\markdownRendererUlItem \markdownRendererStrongEmphasis{device\markdownRendererUnderscore{}status}: device state change (on/off, mode changes, fault alarms)\markdownRendererUlItemEnd 
\markdownRendererUlEnd \markdownRendererInterblockSeparator
{}\markdownRendererHeadingThree{Device Capability Reference (expected modalities per device type)}\markdownRendererInterblockSeparator
{}\markdownRendererUlBegin
\markdownRendererUlItem \markdownRendererStrongEmphasis{Smartwatch/Band}: health, motion, location\markdownRendererUlItemEnd 
\markdownRendererUlItem \markdownRendererStrongEmphasis{Kids Smartwatch}: health, motion, location\markdownRendererUlItemEnd 
\markdownRendererUlItem \markdownRendererStrongEmphasis{Smartphone}: location, app\markdownRendererUnderscore{}usage, dialogue\markdownRendererUlItemEnd 
\markdownRendererUlItem \markdownRendererStrongEmphasis{Laptop}: app\markdownRendererUnderscore{}usage\markdownRendererUlItemEnd 
\markdownRendererUlItem \markdownRendererStrongEmphasis{Tablet}: app\markdownRendererUnderscore{}usage\markdownRendererUlItemEnd 
\markdownRendererUlItem \markdownRendererStrongEmphasis{Smart Speaker}: audio, dialogue\markdownRendererUlItemEnd 
\markdownRendererUlItem \markdownRendererStrongEmphasis{Smart Display (Speaker w/ Screen)}: audio, dialogue, vision\markdownRendererUlItemEnd 
\markdownRendererUlItem \markdownRendererStrongEmphasis{Doorbell Camera}: vision, audio, location\markdownRendererUlItemEnd 
\markdownRendererUlItem \markdownRendererStrongEmphasis{Indoor Camera}: vision, audio\markdownRendererUlItemEnd 
\markdownRendererUlItem \markdownRendererStrongEmphasis{Smart Lock}: location\markdownRendererUlItemEnd 
\markdownRendererUlItem \markdownRendererStrongEmphasis{Motion Sensor}: motion\markdownRendererUlItemEnd 
\markdownRendererUlItem \markdownRendererStrongEmphasis{Door/Window Sensor}: device\markdownRendererUnderscore{}status\markdownRendererUlItemEnd 
\markdownRendererUlItem \markdownRendererStrongEmphasis{Smart Thermostat}: environment\markdownRendererUlItemEnd 
\markdownRendererUlItem \markdownRendererStrongEmphasis{Air Quality Monitor}: environment\markdownRendererUlItemEnd 
\markdownRendererUlItem \markdownRendererStrongEmphasis{Air Purifier}: environment, device\markdownRendererUnderscore{}status\markdownRendererUlItemEnd 
\markdownRendererUlItem \markdownRendererStrongEmphasis{Smart TV}: app\markdownRendererUnderscore{}usage, device\markdownRendererUnderscore{}status\markdownRendererUlItemEnd 
\markdownRendererUlItem \markdownRendererStrongEmphasis{Smart Fridge}: vision, device\markdownRendererUnderscore{}status\markdownRendererUlItemEnd 
\markdownRendererUlItem \markdownRendererStrongEmphasis{Smart Washer/Dryer}: device\markdownRendererUnderscore{}status\markdownRendererUlItemEnd 
\markdownRendererUlItem \markdownRendererStrongEmphasis{Robot Vacuum}: device\markdownRendererUnderscore{}status\markdownRendererUlItemEnd 
\markdownRendererUlItem \markdownRendererStrongEmphasis{Smart Light}: device\markdownRendererUnderscore{}status\markdownRendererUlItemEnd 
\markdownRendererUlItem \markdownRendererStrongEmphasis{Smart Curtain}: device\markdownRendererUnderscore{}status\markdownRendererUlItemEnd 
\markdownRendererUlItem \markdownRendererStrongEmphasis{Smart Scale}: health\markdownRendererUlItemEnd 
\markdownRendererUlItem \markdownRendererStrongEmphasis{Car System}: location, dialogue\markdownRendererUlItemEnd 
\markdownRendererUlItem \markdownRendererStrongEmphasis{Gaming Console}: app\markdownRendererUnderscore{}usage, device\markdownRendererUnderscore{}status\markdownRendererUlItemEnd 
\markdownRendererUlEnd \markdownRendererInterblockSeparator
{}\markdownRendererHeadingThree{Event Description Quality Rules (must follow when writing/rewriting descriptions)}\markdownRendererInterblockSeparator
{}Any description you write or rewrite MUST comply with these rules. Violating them while fixing another issue would introduce a new defect.\markdownRendererInterblockSeparator
{}\markdownRendererHeadingThree{1. Atomic Event Principle}\markdownRendererInterblockSeparator
{}\markdownRendererUlBegin
\markdownRendererUlItem Each event describes a SINGLE instantaneous observation at ONE time point\markdownRendererUlItemEnd 
\markdownRendererUlItem Must NOT describe a span of time (e.g., "from 7:00 to 8:00 David exercised")\markdownRendererUlItemEnd 
\markdownRendererUlItem Must NOT combine multiple distinct activities into one event\markdownRendererUlItemEnd 
\markdownRendererUlItem \markdownRendererStrongEmphasis{Exception}: A summary reading reported at a single timestamp is valid.\markdownRendererUlItemEnd 
\markdownRendererUlEnd \markdownRendererInterblockSeparator
{}Example: "Smart watch recorded: total sleep 6.5h, deep sleep 2.1h" at 07:00 is a single summary observation at wake-up.\markdownRendererInterblockSeparator
{}\markdownRendererHeadingThree{2. Device Perspective}\markdownRendererInterblockSeparator
{}\markdownRendererUlBegin
\markdownRendererUlItem Descriptions MUST be from the device's sensing perspective, not an omniscient narrator\markdownRendererUlItemEnd 
\markdownRendererUlItem Camera: "captured/detected/recorded [visual observation]"\markdownRendererUlItemEnd 
\markdownRendererUlItem Watch: "measured/recorded [physiological/motion data]"\markdownRendererUlItemEnd 
\markdownRendererUlItem Speaker: "heard/detected [audio content]"\markdownRendererUlItemEnd 
\markdownRendererUlItem Phone: "recorded [app activity/location]"\markdownRendererUlItemEnd 
\markdownRendererUlItem Sensor: "measured/detected [environmental readings]"\markdownRendererUlItemEnd 
\markdownRendererUlItem \markdownRendererStrongEmphasis{Violation example}: "David felt tired" - a device cannot know feelings unless expressed aloud\markdownRendererUlItemEnd 
\markdownRendererUlItem \markdownRendererStrongEmphasis{Acceptable}: "Smart watch recorded elevated heart rate of 95bpm and reduced step count"\markdownRendererUlItemEnd 
\markdownRendererUlEnd \markdownRendererInterblockSeparator
{}\markdownRendererHeadingThree{3. Name Usage}\markdownRendererInterblockSeparator
{}\markdownRendererUlBegin
\markdownRendererUlItem Use character NAMES (e.g., "David", "Sarah"), not generic terms ("user", "owner", "the person", "he/she")\markdownRendererUlItemEnd 
\markdownRendererUlItem In multi-person households, must be clear WHO is being described\markdownRendererUlItemEnd 
\markdownRendererUlEnd \markdownRendererInterblockSeparator
{}\markdownRendererHeadingThree{4. Modality-Description Match}\markdownRendererInterblockSeparator
{}\markdownRendererUlBegin
\markdownRendererUlItem Description content must match what the stated modality can perceive.\linebreak
{}See the modality table in Reference Information.\markdownRendererUlItemEnd 
\markdownRendererUlItem health -> physiological metrics only\markdownRendererUlItemEnd 
\markdownRendererUlItem motion -> movement/activity data only\markdownRendererUlItemEnd 
\markdownRendererUlItem vision -> visual observations only\markdownRendererUlItemEnd 
\markdownRendererUlItem audio -> heard sounds/speech only\markdownRendererUlItemEnd 
\markdownRendererUlItem environment -> physical measurements only\markdownRendererUlItemEnd 
\markdownRendererUlItem location -> position/access data only\markdownRendererUlItemEnd 
\markdownRendererUlItem app\markdownRendererUnderscore{}usage -> digital behavior only\markdownRendererUlItemEnd 
\markdownRendererUlItem dialogue -> spoken intentions/preferences only\markdownRendererUlItemEnd 
\markdownRendererUlItem device\markdownRendererUnderscore{}status -> device operational state changes only\markdownRendererUlItemEnd 
\markdownRendererUlEnd \markdownRendererInterblockSeparator
{}\markdownRendererHeadingThree{5. Realistic Readings}\markdownRendererInterblockSeparator
{}\markdownRendererUlBegin
\markdownRendererUlItem Physiological: HR 40-200 bpm, SpO2 88-100\markdownRendererPercentSign{}, temp 35-40 deg C, respiration 10-30/min\markdownRendererUlItemEnd 
\markdownRendererUlItem Environmental: room temp 15-35 deg C, humidity 20-80\markdownRendererPercentSign{}, PM2.5 0-500 ug/m\markdownRendererBackslash{}textasciicircum\markdownRendererLeftBrace{}\markdownRendererRightBrace{}3\markdownRendererUlItemEnd 
\markdownRendererUlItem All numerical values must be physically plausible\markdownRendererUlItemEnd 
\markdownRendererUlEnd \markdownRendererInterblockSeparator
{}\markdownRendererHeadingThree{6. Specificity}\markdownRendererInterblockSeparator
{}\markdownRendererUlBegin
\markdownRendererUlItem Descriptions must be SPECIFIC with concrete details (exact readings, named activities, particular observations)\markdownRendererUlItemEnd 
\markdownRendererUlItem Avoid vague/generic: "normal readings", "activity detected", "everything fine"\markdownRendererUlItemEnd 
\markdownRendererUlEnd \markdownRendererInterblockSeparator
{}\markdownRendererHeadingThree{7. Language}\markdownRendererInterblockSeparator
{}\markdownRendererUlBegin
\markdownRendererUlItem All text must be in English (this is an EN dataset)\markdownRendererUlItemEnd 
\markdownRendererUlItem No Chinese, no garbled/incomplete text\markdownRendererUlItemEnd 
\markdownRendererUlEnd \markdownRendererInterblockSeparator
{}\markdownRendererHeadingThree{Correction Process}\markdownRendererInterblockSeparator
{}\markdownRendererStrongEmphasis{1.} \markdownRendererStrongEmphasis{Backup}: Create backup of current data file (see "Backup \markdownRendererAmpersand{} Working Copy" above).\markdownRendererInterblockSeparator
{}\markdownRendererStrongEmphasis{2.} \markdownRendererStrongEmphasis{Read} the review report to understand all identified issues.\linebreak
{}\markdownRendererStrongEmphasis{3.} For each issue in the report:\linebreak
{}a. \markdownRendererStrongEmphasis{Verify the issue is valid}: Read the actual event and the target question/answer in the data file.\markdownRendererInterblockSeparator
{}Confirm the reviewer's assessment is correct. If you disagree, document why and skip the fix.\markdownRendererInterblockSeparator
{}b. \markdownRendererStrongEmphasis{Fix the event}: Apply the appropriate fix strategy (see below).\markdownRendererInterblockSeparator
{}c. \markdownRendererStrongEmphasis{Post-fix verification} (mandatory for evidence leakage fixes): After writing the new description, verify it doesn't become evidence for ANY question in the scenario.\markdownRendererInterblockSeparator
{}See the "Post-Fix Global Verification" section.\markdownRendererInterblockSeparator
{}\markdownRendererStrongEmphasis{4.} \markdownRendererStrongEmphasis{Rules for fixes}:\markdownRendererInterblockSeparator
{}\markdownRendererUlBegin
\markdownRendererUlItem ONLY modify events with \markdownRendererCodeSpan{source} == \markdownRendererCodeSpan{"adversarial"} that are flagged in the report\markdownRendererUlItemEnd 
\markdownRendererUlItem Do NOT modify any non-adversarial events, questions, answers, or evidence\markdownRendererUnderscore{}event\markdownRendererUnderscore{}ids\markdownRendererUlItemEnd 
\markdownRendererUlItem The fixed event should still be topically similar enough to be a meaningful distractor\markdownRendererUlItemEnd 
\markdownRendererUlEnd \markdownRendererInterblockSeparator
{}\markdownRendererStrongEmphasis{5.} \markdownRendererStrongEmphasis{Modifiable fields} (for adversarial events only):\markdownRendererInterblockSeparator
{}\markdownRendererUlBegin
\markdownRendererUlItem \markdownRendererCodeSpan{description}: can modify; primary fix target for most issues.\markdownRendererUlItemEnd 
\markdownRendererUlItem \markdownRendererCodeSpan{device}: can modify only for device non-existence or device-modality incompatibility.\markdownRendererUlItemEnd 
\markdownRendererUlItem \markdownRendererCodeSpan{modality}: can modify only for device-modality incompatibility or modality-description mismatch.\markdownRendererUlItemEnd 
\markdownRendererUlItem \markdownRendererCodeSpan{timestamp}: can modify only for invalid timestamp format or duplicate device-timestamp.\markdownRendererUlItemEnd 
\markdownRendererUlItem \markdownRendererCodeSpan{location}: can modify only for location-device inconsistency or after changing device.\markdownRendererUlItemEnd 
\markdownRendererUlItem \markdownRendererCodeSpan{event\markdownRendererUnderscore{}id}: never modify; assigned by pipeline.\markdownRendererUlItemEnd 
\markdownRendererUlItem \markdownRendererCodeSpan{source}: never modify; always "adversarial".\markdownRendererUlItemEnd 
\markdownRendererUlItem \markdownRendererCodeSpan{target\markdownRendererUnderscore{}question}: never modify; fixed assignment from generation.\markdownRendererUlItemEnd 
\markdownRendererUlEnd \markdownRendererInterblockSeparator
{}\markdownRendererStrongEmphasis{Principle}: Prefer fixing \markdownRendererCodeSpan{description} alone. Only change metadata fields when the issue is specifically about that field being invalid.\markdownRendererInterblockSeparator
{}After changing any metadata field, always verify the triple constraint:\markdownRendererInterblockSeparator
{}\markdownRendererUlBegin
\markdownRendererUlItem \markdownRendererCodeSpan{device} exists in the persona\markdownRendererUlItemEnd 
\markdownRendererUlItem \markdownRendererCodeSpan{modality} is compatible with that device\markdownRendererUlItemEnd 
\markdownRendererUlItem \markdownRendererCodeSpan{location} matches the device\markdownRendererUlItemEnd 
\markdownRendererUlEnd \markdownRendererInterblockSeparator
{}\markdownRendererHeadingThree{Fix Strategy by Issue Type}\markdownRendererInterblockSeparator
{}\markdownRendererHeadingThree{Evidence Leakage Fixes (Dimension A \markdownRendererAmpersand{} B)}\markdownRendererInterblockSeparator
{}\markdownRendererUlBegin
\markdownRendererUlItem \markdownRendererStrongEmphasis{Direct Answer Leakage}: Change specific details that match the answer.\linebreak
{}Example: if the answer mentions "SpO2 91\markdownRendererPercentSign{}" and the adversarial event also mentions "SpO2 91\markdownRendererPercentSign{}", change to a different reading that doesn't match any evidence detail.\markdownRendererUlItemEnd 
\markdownRendererUlItem \markdownRendererStrongEmphasis{Evidence Equivalence}: Shift the content to be related but not evidentially useful.\linebreak
{}Example: if the question asks about health progression and the event gives readings at a critical time point, change to readings at a non-critical parameter or make them routine/unremarkable.\markdownRendererUlItemEnd 
\markdownRendererUlItem \markdownRendererStrongEmphasis{Causal Chain Completion}: Break the causal link by describing a similar but unrelated activity.\markdownRendererUlItemEnd 
\markdownRendererUlItem \markdownRendererStrongEmphasis{Temporal/Factual Contradiction}: Ensure the event doesn't contradict evidence at the same timestamp.\markdownRendererUlItemEnd 
\markdownRendererUlItem \markdownRendererStrongEmphasis{Cross-Question Evidence Leakage}: Remove the specific details that answer that other question, while keeping the event topically related to its original target\markdownRendererUnderscore{}question.\markdownRendererUlItemEnd 
\markdownRendererUlEnd \markdownRendererInterblockSeparator
{}\markdownRendererHeadingThree{Content/Logic Fixes (Dimension C)}\markdownRendererInterblockSeparator
{}\markdownRendererUlBegin
\markdownRendererUlItem \markdownRendererStrongEmphasis{Device-Modality Incompatibility / Device Non-Existence}: Fix the \markdownRendererCodeSpan{device} or \markdownRendererCodeSpan{modality} field to use a valid device from the persona's \markdownRendererCodeSpan{all\markdownRendererUnderscore{}devices} list with a compatible modality.\linebreak
{}Then adjust the description to match:\markdownRendererInterblockSeparator
{}\markdownRendererUlBegin
\markdownRendererUlItem \markdownRendererCodeSpan{device} exists in \markdownRendererCodeSpan{all\markdownRendererUnderscore{}devices[].device\markdownRendererUnderscore{}id}\markdownRendererUlItemEnd 
\markdownRendererUlItem \markdownRendererCodeSpan{modality} is listed in that device's \markdownRendererCodeSpan{modality} field\markdownRendererUlItemEnd 
\markdownRendererUlItem \markdownRendererCodeSpan{location} matches the device's \markdownRendererCodeSpan{location} for fixed devices or follows the character for wearable/carried devices\markdownRendererUlItemEnd 
\markdownRendererUlEnd \markdownRendererUlItemEnd 
\markdownRendererUlItem \markdownRendererStrongEmphasis{Modality-Description Mismatch}: Change the description to describe information that the stated modality can actually perceive.\linebreak
{}Or change the modality to match what the description actually describes, if a compatible modality exists for that device.\markdownRendererUlItemEnd 
\markdownRendererUlItem \markdownRendererStrongEmphasis{Impossible Readings}: Replace with realistic values within normal physiological/environmental ranges.\markdownRendererUlItemEnd 
\markdownRendererUlItem \markdownRendererStrongEmphasis{Location-Device Inconsistency}: Fix the location to match where the device is installed.\linebreak
{}Or adjust the description to be consistent with the device's actual location.\markdownRendererUlItemEnd 
\markdownRendererUlItem \markdownRendererStrongEmphasis{Character-Device Ownership Mismatch}: Fix the description to refer to the device's actual owner.\markdownRendererUlItemEnd 
\markdownRendererUlItem \markdownRendererStrongEmphasis{Language Error}: Rewrite in proper English.\markdownRendererUlItemEnd 
\markdownRendererUlItem \markdownRendererStrongEmphasis{Timestamp Format Invalid}: Fix to valid \markdownRendererCodeSpan{YYYY-MM-DDTHH:mm:ss} within time span 2026-05-11 \markdownRendererBackslash{}textasciitilde\markdownRendererLeftBrace{}\markdownRendererRightBrace{} 2026-05-17.\linebreak
{}Keep the time portion plausible for the described activity.\markdownRendererUlItemEnd 
\markdownRendererUlItem \markdownRendererStrongEmphasis{Weekday-Activity Inconsistency}: Rewrite the description to match the day of the week (use date-weekday mapping: Mon=05-11 through Sun=05-17).\linebreak
{}Example: if the event is on Saturday but describes school attendance, change to a weekend-appropriate activity.\markdownRendererUlItemEnd 
\markdownRendererUlItem \markdownRendererStrongEmphasis{Character-Action Age/Role Inconsistency}: Rewrite the description so the action is appropriate for the character's age, role, and health status.\linebreak
{}Cross-reference the persona's character definitions.\markdownRendererUlItemEnd 
\markdownRendererUlItem \markdownRendererStrongEmphasis{Duplicate Device-Timestamp}: Shift the adversarial event's timestamp by 1-5 minutes to avoid collision with the other event on the same device.\linebreak
{}Ensure the description still makes sense at the new time.\markdownRendererUlItemEnd 
\markdownRendererUlItem \markdownRendererStrongEmphasis{Non-Atomic Event}: Rewrite to describe a single instantaneous observation at the event's timestamp.\linebreak
{}Remove any time spans or multiple combined activities.\markdownRendererUlItemEnd 
\markdownRendererUlItem \markdownRendererStrongEmphasis{Omniscient Narrator Perspective}: Rewrite from the device's sensing perspective.\linebreak
{}Replace internal states/feelings with device-observable data.\markdownRendererUlItemEnd 
\markdownRendererUlItem \markdownRendererStrongEmphasis{Generic/Anonymous Name Usage}: Replace "user"/"owner"/"the person" with the appropriate character name.\linebreak
{}The character name is typically the device owner.\markdownRendererUlItemEnd 
\markdownRendererUlEnd \markdownRendererInterblockSeparator
{}\markdownRendererHeadingThree{Adversarial Quality Fixes (Dimension D)}\markdownRendererInterblockSeparator
{}\markdownRendererUlBegin
\markdownRendererUlItem \markdownRendererStrongEmphasis{Topic Irrelevance}: Rewrite the description to be semantically related to the target question's topic while still not being valid evidence.\linebreak
{}Keep the same device/modality/timestamp but change the content to share topical overlap with the target question's domain.\markdownRendererUlItemEnd 
\markdownRendererUlItem \markdownRendererStrongEmphasis{Too Vague/Generic}: Add concrete, specific details to the description.\linebreak
{}Use specific readings, named activities, and particular observations.\linebreak
{}Make sure these details don't constitute evidence for any question.\markdownRendererUlItemEnd 
\markdownRendererUlEnd \markdownRendererInterblockSeparator
{}\markdownRendererHeadingThree{Post-Fix Global Verification}\markdownRendererInterblockSeparator
{}\markdownRendererStrongEmphasis{After fixing any evidence leakage issue (Dimension A or B), you MUST perform this verification:}\markdownRendererInterblockSeparator
{}\markdownRendererStrongEmphasis{1.} Read ALL questions in the same scenario (not just the target or affected question)\linebreak
{}\markdownRendererStrongEmphasis{2.} For the new/fixed description, check: does it contain information that directly helps answer ANY question?\linebreak
{}\markdownRendererStrongEmphasis{3.} Specifically focus on questions whose topic overlaps with the new description's content\linebreak
{}\markdownRendererStrongEmphasis{4.} If the fix creates a new leakage to another question: iterate the fix until no leakage remains to any question\markdownRendererInterblockSeparator
{}This step is critical because fixing a leak to one question can accidentally create a leak to another.\markdownRendererInterblockSeparator
{}\markdownRendererHeadingThree{Output}\markdownRendererInterblockSeparator
{}Generate fix report in the same directory. File naming by loop number:\markdownRendererInterblockSeparator
{}\markdownRendererUlBegin
\markdownRendererUlItem Loop 1: \markdownRendererCodeSpan{step1\markdownRendererUnderscore{}fix\markdownRendererUnderscore{}report\markdownRendererUnderscore{}loop1.md}\markdownRendererUlItemEnd 
\markdownRendererUlItem Loop 2: \markdownRendererCodeSpan{step1\markdownRendererUnderscore{}fix\markdownRendererUnderscore{}report\markdownRendererUnderscore{}loop2.md}\markdownRendererUlItemEnd 
\markdownRendererUlItem Loop N: \markdownRendererCodeSpan{step1\markdownRendererUnderscore{}fix\markdownRendererUnderscore{}report\markdownRendererUnderscore{}loopN.md}\markdownRendererUlItemEnd 
\markdownRendererUlEnd \markdownRendererInterblockSeparator
{}The loop number will be provided in the task prompt when you are invoked.\markdownRendererInterblockSeparator
{}Fix report format:\markdownRendererInterblockSeparator
{}\markdownRendererInputFencedCode{prompt_tex/a182dba903578ded45eaee172f023791.verbatim}{text}\markdownRendererInterblockSeparator
{}\markdownRendererInputFencedCode{prompt_tex/40d3f1d072674c984d4156dca82959bb.verbatim}{text}\markdownRendererInterblockSeparator
{}\markdownRendererInputFencedCode{prompt_tex/45e08a52077e0f4e67bc353283e7acc6.verbatim}{text}\markdownRendererInterblockSeparator
{}\markdownRendererInputFencedCode{prompt_tex/4e3fa80d71284c9c22e0a475f6af85e9.verbatim}{text}\markdownRendererInterblockSeparator
{}\markdownRendererInputFencedCode{prompt_tex/dc3e3215640db4e51c035d2d69daf545.verbatim}{text}\markdownRendererInterblockSeparator
{}\markdownRendererInputFencedCode{prompt_tex/3d673980cc440bd70bca9b1d072b9a47.verbatim}{text}\markdownRendererInterblockSeparator
{}\markdownRendererInputFencedCode{prompt_tex/619753e60bf2a3ddd825596e6a274c78.verbatim}{text}\markdownRendererInterblockSeparator
{}\markdownRendererInputFencedCode{prompt_tex/30c3b14326c074e7a0cc176f695d2df4.verbatim}{text}\markdownRendererInterblockSeparator
{}\markdownRendererInputFencedCode{prompt_tex/0907d6d60a3487d0cb44d2485183b99b.verbatim}{text}\markdownRendererInterblockSeparator
{}\markdownRendererHeadingThree{Important Notes}\markdownRendererInterblockSeparator
{}\markdownRendererUlBegin
\markdownRendererUlItem Verify each issue before fixing - reviewers can have false positives\markdownRendererUlItemEnd 
\markdownRendererUlItem Fixed descriptions must still be realistic and specific (not generic/vague)\markdownRendererUlItemEnd 
\markdownRendererUlItem The goal is to maintain distractor quality while removing evidence leakage\markdownRendererUlItemEnd 
\markdownRendererUlItem Do not over-correct: the event should still be topically related enough to test discrimination ability\markdownRendererUlItemEnd 
\markdownRendererUlItem For device/modality fixes, always cross-reference the persona's \markdownRendererCodeSpan{all\markdownRendererUnderscore{}devices} to ensure the fix uses valid devices and compatible modalities.\linebreak
{}After any device/modality fix, verify the triple constraint: device exists, modality is compatible with device, location matches device's installation.\markdownRendererUlItemEnd 
\markdownRendererUlItem After fixing, double-check that the new description doesn't accidentally become evidence for ANY other question in the scenario.\linebreak
{}See Post-Fix Global Verification.\markdownRendererUlItemEnd 
\markdownRendererUlItem Adversarial events intentionally do NOT have a \markdownRendererCodeSpan{characters} field - do not add one\markdownRendererUlItemEnd 
\markdownRendererUlEnd \markdownRendererDocumentEnd\relax
\end{promptbox}

\subsection{MemFuse Method Prompts}

These templates govern the LLM-controlled parts of memory construction and retrieval.

\subsubsection{Fusion Agent}

\paragraph{System prompt.}
This prompt defines the fusion agent's role, available tools, validity constraints, output schema, and fusion criteria.
\begin{promptlisting}{Fusion Agent System Prompt}
\markdownRendererDocumentBegin
You are a tool-driven memory fusion agent.\markdownRendererInterblockSeparator
{}Decide whether the current new event should create causal edges with existing memories, create a fused node, update an existing fused node, or repair a noisy fused pack.\markdownRendererInterblockSeparator
{}You may use three tools:\linebreak
{}1. search\markdownRendererUnderscore{}memory: Search candidate memories with a retrieval query.\linebreak
{}The system has already called search\markdownRendererUnderscore{}memory once with the current new event.\linebreak
{}If candidates are insufficient, you may call search\markdownRendererUnderscore{}memory a limited number of additional times.\linebreak
{}2. get\markdownRendererUnderscore{}pack\markdownRendererUnderscore{}members: Inspect the member details of a candidate fused node only when member-level evidence is required.\linebreak
{}Use it when you are seriously considering update\markdownRendererUnderscore{}fusion\markdownRendererUnderscore{}node into a full pack, or when visible evidence gives a specific reason to suspect clear noisy members.\linebreak
{}Each event has only a small lookup budget.\linebreak
{}Repeated lookups of the same pack waste a turn and should be avoided.\linebreak
{}3. submit\markdownRendererUnderscore{}fusion\markdownRendererUnderscore{}plan: Submit the final fusion plan. After submission, processing for this event ends.\markdownRendererInterblockSeparator
{}Important constraints:\markdownRendererInterblockSeparator
{}\markdownRendererUlBegin
\markdownRendererUlItem All JSON string values must be written in English. If input events or candidate summaries are in another language, translate or paraphrase them into English in your summaries and reasons.\markdownRendererUlItemEnd 
\markdownRendererUlItem search\markdownRendererUnderscore{}memory returns the same format as Candidate memories, including fused node summaries, retrieved member events, and standalone events.\markdownRendererUlItemEnd 
\markdownRendererUlItem get\markdownRendererUnderscore{}pack\markdownRendererUnderscore{}members returns formatted text with the fused node summary, metadata, valid removable member IDs, and member event list.\markdownRendererUlItemEnd 
\markdownRendererUlItem Do not invent chunk\markdownRendererUnderscore{}ids or fused node IDs that are not present in candidates.\markdownRendererUlItemEnd 
\markdownRendererUlItem Never output placeholder literals such as "Fused node [ID]" or "candidate fused node ID". Always output a real fused\markdownRendererUnderscore{}xxx ID.\markdownRendererUlItemEnd 
\markdownRendererUlItem target\markdownRendererUnderscore{}chunk\markdownRendererUnderscore{}id and fusion\markdownRendererUnderscore{}node\markdownRendererUnderscore{}id can only use fused node IDs shown as "Fused node [ID]" in candidates, not ordinary event IDs.\markdownRendererUlItemEnd 
\markdownRendererUlItem involved\markdownRendererUnderscore{}events should contain only the current new event ID.\markdownRendererUlItemEnd 
\markdownRendererUlItem Except for remove\markdownRendererUnderscore{}member, every operation must directly involve the current new event.\markdownRendererUlItemEnd 
\markdownRendererUlItem For create\markdownRendererUnderscore{}edge, one endpoint must be the current new event ID. The other endpoint must be a real chunk\markdownRendererUnderscore{}id or fused\markdownRendererUnderscore{}xxx ID visible in Candidate memories or a get\markdownRendererUnderscore{}pack\markdownRendererUnderscore{}members response.\linebreak
{}Never create edges between two old memories, and never use an event ID that only appears inside your own reasoning or summary text.\markdownRendererUlItemEnd 
\markdownRendererUlItem Candidate fused nodes show member count, time range, subjects, and devices. When deciding whether to update\markdownRendererUnderscore{}existing, use these structured ranges: the current new event should naturally belong to the same concrete event segment, not merely share a topic.\markdownRendererUlItemEnd 
\markdownRendererUlItem A fused pack has a maximum member count. When a candidate fused node reaches the configured maximum, the pack is full.\markdownRendererUlItemEnd 
\markdownRendererUlItem A fused pack with one member is a singleton.\linebreak
{}Do not call get\markdownRendererUnderscore{}pack\markdownRendererUnderscore{}members merely to inspect a singleton pack, and never output remove\markdownRendererUnderscore{}member for a singleton pack because it would empty the pack.\linebreak
{}For singleton packs, either update\markdownRendererUnderscore{}fusion\markdownRendererUnderscore{}node if the current event belongs with that member, create\markdownRendererUnderscore{}edge if it is only causally related, or create\markdownRendererUnderscore{}fusion\markdownRendererUnderscore{}node/no\markdownRendererUnderscore{}op if it is separate.\markdownRendererUlItemEnd 
\markdownRendererUlItem Do not call get\markdownRendererUnderscore{}pack\markdownRendererUnderscore{}members for every full pack.\linebreak
{}If the candidate summary, metadata, and retrieved member events already make the decision clear, submit a plan directly.\markdownRendererUlItemEnd 
\markdownRendererUlItem If you plan to update\markdownRendererUnderscore{}fusion\markdownRendererUnderscore{}node into a full pack and the visible evidence is insufficient to know whether the current event belongs, first call get\markdownRendererUnderscore{}pack\markdownRendererUnderscore{}members for that pack.\linebreak
{}After inspection, remove clearly noisy members if they exist.\linebreak
{}If there is no clear noise, do not remove members just to make room; you may still output update\markdownRendererUnderscore{}fusion\markdownRendererUnderscore{}node and the system will apply its full-pack logic.\markdownRendererUlItemEnd 
\markdownRendererUlItem remove\markdownRendererUnderscore{}member only removes BELONG edges between members and the fused node; it does not delete atomic events.\linebreak
{}Remove only members that clearly do not belong to the concrete event segment.\linebreak
{}When uncertain, do not remove.\linebreak
{}Never remove members just to make room.\markdownRendererUlItemEnd 
\markdownRendererUlItem remove\markdownRendererUnderscore{}member is also a candidate pack repair operation: if get\markdownRendererUnderscore{}pack\markdownRendererUnderscore{}members reveals clear noisy members, output remove\markdownRendererUnderscore{}member even if the current event ultimately does not join that pack.\linebreak
{}Do not skip obvious pack repair just because the final action is create\markdownRendererUnderscore{}new or no\markdownRendererUnderscore{}op.\markdownRendererUlItemEnd 
\markdownRendererUlItem If you inspect a pack and do not output remove\markdownRendererUnderscore{}member, that means you confirm all inspected members belong to the same concrete event segment and provide useful evidence for the pack summary.\linebreak
{}Do not replace this judgment with weak similarity such as related topic, same day, same household/dormitory, or close timestamps.\markdownRendererUlItemEnd 
\markdownRendererUlItem If you output remove\markdownRendererUnderscore{}member, you must also output summary.\linebreak
{}That summary must describe the remaining pack after removal and must not include information from removed members.\markdownRendererUlItemEnd 
\markdownRendererUlItem remove\markdownRendererUnderscore{}member.remove\markdownRendererUnderscore{}member\markdownRendererUnderscore{}ids must use exact member IDs shown in the latest get\markdownRendererUnderscore{}pack\markdownRendererUnderscore{}members response for that fused node.\linebreak
{}Do not remove IDs copied from candidates, summaries, placeholders, or the current new event.\markdownRendererUlItemEnd 
\markdownRendererUlItem If the same plan first removes members and then updates the fused node, remove\markdownRendererUnderscore{}member.summary describes the pack after removal, and update\markdownRendererUnderscore{}fusion\markdownRendererUnderscore{}node.summary describes the final pack after removal plus the current event.\markdownRendererUlItemEnd 
\markdownRendererUlItem Use remove\markdownRendererUnderscore{}member only for a few clear noisy members, within the configured per-operation removal limit.\linebreak
{}Do not use it as a bulk pack restructuring tool.\markdownRendererUlItemEnd 
\markdownRendererUlItem remove\markdownRendererUnderscore{}member must leave at least one original pack member.\linebreak
{}If no core member can be kept, the pack cannot be repaired by deletion; use create\markdownRendererUnderscore{}fusion\markdownRendererUnderscore{}node, create\markdownRendererUnderscore{}edge, or no\markdownRendererUnderscore{}op instead.\markdownRendererUlItemEnd 
\markdownRendererUlItem General remove\markdownRendererUnderscore{}member rule: remove a member only when it clearly breaks the pack's "same concrete event segment" boundary.\linebreak
{}Examples: a different activity instance, an unrelated time segment, a separate person storyline without interaction/evidence support, a different location/device scene that does not support the same event, a nearby system/environment reading with no event contribution, or a member that would force the summary into a broad daily log/topic bucket.\markdownRendererUlItemEnd 
\markdownRendererUlItem Do not remove members that are before/after steps of the same event, family care/game/meal/return-home follow-ups in a continuous time segment, mutually supportive multi-device observations, a dialogue and direct follow-up action, an evidence chain for the same problem/conflict, complementary observations in the same household/dormitory scene, or cases where you are uncertain.\markdownRendererUlItemEnd 
\markdownRendererUlItem After calling get\markdownRendererUnderscore{}pack\markdownRendererUnderscore{}members, choose one of three paths:\markdownRendererInterblockSeparator
{}\markdownRendererOlBegin
\markdownRendererOlItemWithNumber{1}members are clean, so update\markdownRendererUnderscore{}fusion\markdownRendererUnderscore{}node with a clean summary if appropriate\markdownRendererOlItemEnd 
\markdownRendererOlItemWithNumber{2}a few clear noisy members exist, so remove\markdownRendererUnderscore{}member first and then update\markdownRendererUnderscore{}fusion\markdownRendererUnderscore{}node if appropriate\markdownRendererOlItemEnd 
\markdownRendererOlItemWithNumber{3}the pack is too mixed or needs many removals, so do not update that pack and use create\markdownRendererUnderscore{}fusion\markdownRendererUnderscore{}node, create\markdownRendererUnderscore{}edge, or no\markdownRendererUnderscore{}op\markdownRendererOlItemEnd 
\markdownRendererOlEnd \markdownRendererUlItemEnd 
\markdownRendererUlItem For full packs, the key question is not whether you can make room, but whether the pack is still one concrete event segment.\linebreak
{}If there are obvious off-topic members, remove them first.\linebreak
{}If most members are mixed, keep only the coherent core when possible; otherwise do not merge the current event into the original pack.\markdownRendererUlItemEnd 
\markdownRendererUlItem Do not repeatedly call get\markdownRendererUnderscore{}pack\markdownRendererUnderscore{}members for the same pack. Use the member details already shown to submit a plan.\markdownRendererUlItemEnd 
\markdownRendererUlItem If member relationships are clean but the old summary is polluted or too broad, do not remove members just to rewrite the summary; use update\markdownRendererUnderscore{}fusion\markdownRendererUnderscore{}node.summary to rewrite a concrete summary covering the current pack members and the current new event.\markdownRendererUlItemEnd 
\markdownRendererUlItem Keep fused summaries compact but evidence-rich.\linebreak
{}Preserve details that may matter in future QA: participants, source device, time order, concrete actions, objects, colors, counts, rounds/attempts, health readings, emotional reactions, care/concern, conflicts, viewpoint differences, and outcomes.\markdownRendererUlItemEnd 
\markdownRendererUlItem Prefer create\markdownRendererUnderscore{}edge without fusion when events are causally related but belong to different concrete activity instances. Prefer fusion only when events describe the same activity instance, short scene, or tightly connected evidence segment.\markdownRendererUlItemEnd 
\markdownRendererUlItem If a causal relationship is plausible but the other endpoint is not visible as an exact candidate/member ID, do not output create\markdownRendererUnderscore{}edge. Use no\markdownRendererUnderscore{}op or search\markdownRendererUnderscore{}memory with a targeted query if search budget remains.\markdownRendererUlItemEnd 
\markdownRendererUlItem Prefer create\markdownRendererUnderscore{}fusion\markdownRendererUnderscore{}node over update\markdownRendererUnderscore{}fusion\markdownRendererUnderscore{}node when the existing pack has a different core activity, broad daily-log scope, mixed subject lines, or would require the summary to become generic.\markdownRendererUlItemEnd 
\markdownRendererUlItem Relatedness is not fusion. If events only share the same topic, person, day, or device but lack concrete scene/activity continuity, prefer create\markdownRendererUnderscore{}edge or no\markdownRendererUnderscore{}op.\markdownRendererUlItemEnd 
\markdownRendererUlItem Do not force noisy or weakly related events into a fused node. If the current event is causally related but not part of the same concrete event segment, prefer create\markdownRendererUnderscore{}edge without fusion.\markdownRendererUlItemEnd 
\markdownRendererUlItem Fusion is appropriate for multi-device evidence in the same short scene, consecutive steps of the same activity instance, evidence before and after the same problem/conflict, a dialogue and its direct follow-up action, or complementary observations in one household/dormitory event segment.\markdownRendererUlItemEnd 
\markdownRendererUlItem When the default candidates are sufficient, prefer submit\markdownRendererUnderscore{}fusion\markdownRendererUnderscore{}plan directly. Call search\markdownRendererUnderscore{}memory only when candidates are clearly insufficient and you have a clear new retrieval query.\markdownRendererUlItemEnd 
\markdownRendererUlItem When the search\markdownRendererUnderscore{}memory budget is exhausted, submit a fusion plan using existing candidates; do not request more search\markdownRendererUnderscore{}memory calls.\markdownRendererUlItemEnd 
\markdownRendererUlItem remove\markdownRendererUnderscore{}member is not a top-level tool. Do not output \markdownRendererLeftBrace{}"tool":"remove\markdownRendererUnderscore{}member"\markdownRendererRightBrace{}; it can only appear as an operation inside submit\markdownRendererUnderscore{}fusion\markdownRendererUnderscore{}plan.plan.operations.\markdownRendererUlItemEnd 
\markdownRendererUlItem When uncertain, choose none. It is better not to fuse than to fuse incorrectly.\markdownRendererUlItemEnd 
\markdownRendererUlItem You cannot write to the store directly; you can only output a JSON plan through submit\markdownRendererUnderscore{}fusion\markdownRendererUnderscore{}plan.\markdownRendererUlItemEnd 
\markdownRendererUlEnd \markdownRendererInterblockSeparator
{}Each response must be exactly one JSON object in one of the following formats.\markdownRendererInterblockSeparator
{}Call search\markdownRendererUnderscore{}memory:\markdownRendererInterblockSeparator
{}\markdownRendererInputFencedCode{prompt_tex/b9d4811ef74300892a31d10a99b2a1f5.verbatim}{json}\markdownRendererInterblockSeparator
{}Call get\markdownRendererUnderscore{}pack\markdownRendererUnderscore{}members only when member-level inspection is necessary:\markdownRendererInterblockSeparator
{}\markdownRendererInputFencedCode{prompt_tex/c43fc46a51c64b04526f1048b451bddc.verbatim}{json}\markdownRendererInterblockSeparator
{}Submit fusion plan:\markdownRendererInterblockSeparator
{}\markdownRendererInputFencedCode{prompt_tex/beb388bb8d7302670a89af9bb64792f0.verbatim}{json}\markdownRendererInterblockSeparator
{}Fusion standard: the events should belong to the same concrete activity instance, the same short-time scene, the same event segment, or a mutually supportive evidence chain around the same problem/conflict.\markdownRendererInterblockSeparator
{}Do not fuse merely because they share topic, person, day, or device.\markdownRendererInterblockSeparator
{}create\markdownRendererUnderscore{}new is for multiple events that form the same concrete activity, same scene, short causal chain, or complementary observations in one event segment.\markdownRendererInterblockSeparator
{}update\markdownRendererUnderscore{}existing is for a current new event that is a direct next step of an existing fused node, same-scene complementary evidence, or a key detail that completes that event segment.\markdownRendererInterblockSeparator
{}none is for cases without enough evidence for fusion or edge creation.\markdownRendererInterblockSeparator
{}Summary requirements: write 1-2 concise English sentences, ideally 40-100 English words, including who, when, what specifically happened, and the key result.\markdownRendererInterblockSeparator
{}Do not write a broad topic.\markdownRendererInterblockSeparator
{}For update\markdownRendererUnderscore{}fusion\markdownRendererUnderscore{}node, summary must be the updated fused node summary, not the current event summary.\markdownRendererInterblockSeparator
{}Preserve key people, time, actions, results, conflicts, and viewpoint differences from the old summary when they remain relevant to the same activity/scene.\markdownRendererInterblockSeparator
{}You may compress minor background, but do not drop key information that changes the original meaning or affects future QA.\markdownRendererInterblockSeparator
{}remove\markdownRendererUnderscore{}member.summary must be the new pack summary after removing noisy members.\markdownRendererInterblockSeparator
{}If update\markdownRendererUnderscore{}fusion\markdownRendererUnderscore{}node follows it, update\markdownRendererUnderscore{}fusion\markdownRendererUnderscore{}node.summary must be the final pack summary.\markdownRendererDocumentEnd\relax
\end{promptlisting}

\subsubsection{Fusion-Aware Retrieval}

This agentic retrieval loop uses two prompt templates: one to plan each round and one to judge accumulated evidence. Each box below is one complete call containing its System and User messages; there are no separate calls for the planning rules or JSON schema. The bracketed role labels and cross-reference are presentation annotations rather than literal prompt text. The controller may stop only after two rounds and is capped at five rounds.

\paragraph{Round Retrieval Planning Prompt.}
This prompt asks the controller to propose the next structured retrieval plan for the current round.
\begin{promptlisting}{Round Retrieval Planning Prompt}
\markdownRendererDocumentBegin
\markdownRendererHeadingThree{System}\markdownRendererInterblockSeparator
{}You are the planning phase of an evidence-aware memory retrieval controller. Do not answer the question. Emit one bounded, executable retrieval plan as strict JSON.\markdownRendererInterblockSeparator
{}\markdownRendererHeadingThree{User}\markdownRendererInterblockSeparator
{}Create the first retrieval plan for this question.\markdownRendererInterblockSeparator
{}\markdownRendererStrongEmphasis{Original question:} \markdownRendererLeftBrace{}question\markdownRendererRightBrace{}\markdownRendererInterblockSeparator
{}\markdownRendererStrongEmphasis{Frozen question time:} \markdownRendererLeftBrace{}query\markdownRendererUnderscore{}time\markdownRendererRightBrace{}\markdownRendererInterblockSeparator
{}\markdownRendererStrongEmphasis{Question weekday/calendar reference:} \markdownRendererLeftBrace{}weekday\markdownRendererUnderscore{}calendar\markdownRendererUnderscore{}text\markdownRendererRightBrace{}\markdownRendererInterblockSeparator
{}\markdownRendererStrongEmphasis{Requester:} \markdownRendererLeftBrace{}requester\markdownRendererUnderscore{}id\markdownRendererRightBrace{}\markdownRendererInterblockSeparator
{}\markdownRendererStrongEmphasis{Planning-field mapping for the unified Wrapper Controller:}\markdownRendererInterblockSeparator
{}\markdownRendererUlBegin
\markdownRendererUlItem \markdownRendererCodeSpan{focus} is the primary standalone query and must preserve all relevant constraints from the original question.\markdownRendererUlItemEnd 
\markdownRendererUlItem \markdownRendererCodeSpan{semantic\markdownRendererUnderscore{}queries}, \markdownRendererCodeSpan{bm25\markdownRendererUnderscore{}query}, \markdownRendererCodeSpan{keywords}, and \markdownRendererCodeSpan{entities} correspond to the legacy \markdownRendererCodeSpan{query\markdownRendererUnderscore{}rewrite} fields.\markdownRendererUlItemEnd 
\markdownRendererUlItem \markdownRendererCodeSpan{coverage} in the legacy rules means \markdownRendererCodeSpan{coverage\markdownRendererUnderscore{}diversification} here; it is distinct from evidence-facet coverage.\markdownRendererUlItemEnd 
\markdownRendererUlItem The strings in the schema are descriptions, not default values.\markdownRendererUlItemEnd 
\markdownRendererUlEnd \markdownRendererInterblockSeparator
{}\markdownRendererStrongEmphasis{Planning decision rules:}\markdownRendererInterblockSeparator
{}\markdownRendererUlBegin
\markdownRendererUlItem \markdownRendererStrongEmphasis{Query rewriting.} \markdownRendererCodeSpan{query\markdownRendererUnderscore{}rewrite} has the highest priority and must not be distorted by later control strategies:\markdownRendererInterblockSeparator
{}\markdownRendererUlBegin
\markdownRendererUlItem Every semantic query must preserve the main answer shape of the original question.\markdownRendererUlItemEnd 
\markdownRendererUlItem Semantic queries should retrieve evidence events, not draft the final answer.\markdownRendererUlItemEnd 
\markdownRendererUlItem If the original question asks for a full sequence, what exactly happened, a timeline, a process, or an experience, semantic\markdownRendererUnderscore{}queries must keep that process-reconstruction intent. Do not rewrite it into a why/cause query.\markdownRendererUlItemEnd 
\markdownRendererUlItem Only generate cause-oriented rewrites when the original question explicitly asks why, for the reason, what caused it, or how it happened.\markdownRendererUlItemEnd 
\markdownRendererUlItem The first semantic\markdownRendererUnderscore{}query should stay close to the core wording of the original question; only compress, remove redundancy, and highlight keywords.\markdownRendererUlItemEnd 
\markdownRendererUlItem Later semantic queries should cover complementary evidence facets, such as participant actions, device observations, location/object/state changes, before/after context, or source-specific viewpoints. Do not make all rewrites near-duplicates.\markdownRendererUlItemEnd 
\markdownRendererUlItem Preserve detail requests such as colors, quantities, durations, counts, game rounds, health readings, emotional reactions, care/concern, conflicts, viewpoints, and explicit objects; these details often decide checklist credit.\markdownRendererUlItemEnd 
\markdownRendererUlItem For contradiction or record-arbitration questions, preserve every competing value and source in both semantic\markdownRendererUnderscore{}queries and bm25\markdownRendererUnderscore{}query. Include nearby evidence facets such as timestamps, pulse/score/count/unit fields, manual entry vs automated device source, and transcription/recording context when the original question compares two records.\markdownRendererUlItemEnd 
\markdownRendererUlItem bm25\markdownRendererUnderscore{}query should remain close to the original question and keep exact lexical anchors. Do not replace names/devices/objects with generic paraphrases in bm25\markdownRendererUnderscore{}query.\markdownRendererUlItemEnd 
\markdownRendererUlEnd \markdownRendererUlItemEnd 
\markdownRendererUlItem \markdownRendererStrongEmphasis{Time-window selection.} Set \markdownRendererCodeSpan{time\markdownRendererUnderscore{}window} only when one bounded time span is useful for retrieval. A calendar clue is not enough by itself.\markdownRendererInterblockSeparator
{}\markdownRendererUlBegin
\markdownRendererUlItem Disable time\markdownRendererUnderscore{}window for broad or longitudinal questions spanning multiple days, such as "this week", "past week", "over the week", "eventually", "from X to Y", "lead to ... later", "commitments/plans this week", "everything bought/ordered this week", or questions mentioning multiple distinct dates. These need semantic/BM25 coverage across days rather than one day crowding out other evidence.\markdownRendererUlItemEnd 
\markdownRendererUlItem Disable time\markdownRendererUnderscore{}window when the question has no explicit date and the only time clue is a broad period like "this week" or "recently".\markdownRendererUlItemEnd 
\markdownRendererUlItem If a question contains both an initial event date and a later outcome date, do not choose only the initial date. Disable time\markdownRendererUnderscore{}window unless the question asks specifically about one bounded episode.\markdownRendererUlItemEnd 
\markdownRendererUlEnd \markdownRendererUlItemEnd 
\markdownRendererUlItem \markdownRendererStrongEmphasis{Enabled time windows.} When \markdownRendererCodeSpan{time\markdownRendererUnderscore{}window.enabled=true}, always output both \markdownRendererCodeSpan{start\markdownRendererUnderscore{}time} and \markdownRendererCodeSpan{end\markdownRendererUnderscore{}time}. The retrieval code cannot use a date-only window.\markdownRendererInterblockSeparator
{}\markdownRendererUlBegin
\markdownRendererUlItem For date-only questions such as an explicit calendar date, a named weekday, "that day", "all day", or "throughout the day", use start\markdownRendererUnderscore{}time="00:00:00" and end\markdownRendererUnderscore{}time="23:59:59".\markdownRendererUlItemEnd 
\markdownRendererUlItem Do not treat a question as date-only if it contains a part-of-day or activity/session anchor. Words such as morning, afternoon, evening, night, appointment, commute, clinic visit, or practice imply a narrower window.\markdownRendererUlItemEnd 
\markdownRendererUlItem Do not infer a clock range from meal/session labels alone, such as dinner, breakfast, lunch, movie night, board game session, yoga session, or conversation, unless the question also gives an explicit time, part-of-day, or the answer asks only about that bounded session. If the session label may be colloquial or the key anchors are topical (for example treatment plans, purchases, bookings, reminders), prefer semantic/BM25 retrieval with time\markdownRendererUnderscore{}window disabled or with a low quota.\markdownRendererUlItemEnd 
\markdownRendererUlItem Sleep/night questions often refer to evidence recorded around wake-up, not the whole daytime. If the question asks about "that night", sleep quality, sleep score, sleep tracker/bed sensor records, or an earlier-morning vs later-morning sleep summary, do not use a full-day daytime window. Use the referenced morning window, usually 00:00:00-09:59:59, and rely on exact sleep-score/duration/device anchors for prior-night evidence.\markdownRendererUlItemEnd 
\markdownRendererUlItem When a question names a weekday/date but asks about sleep "that night" or "earlier that morning", the relevant evidence is the early-morning sleep records for that resolved date, not the entire daytime. Keep sleep-score numbers, durations, and device names as lexical anchors.\markdownRendererUlItemEnd 
\markdownRendererUlItem Explicit part-of-day words are stronger than a bare calendar date. If the question says morning, afternoon, evening, night, or a similar part-of-day phrase, use that narrower range unless the question clearly spans multiple days.\markdownRendererUlItemEnd 
\markdownRendererUlItem When both a date and a part-of-day/session anchor are present, keep the narrower part-of-day/session window instead of expanding to the full day.\markdownRendererUlItemEnd 
\markdownRendererUlItem Use null start\markdownRendererUnderscore{}time/end\markdownRendererUnderscore{}time only when time\markdownRendererUnderscore{}window.enabled=false.\markdownRendererUlItemEnd 
\markdownRendererUlEnd \markdownRendererUlItemEnd 
\markdownRendererUlItem \markdownRendererStrongEmphasis{Relative weekday/date normalization.}\markdownRendererInterblockSeparator
{}\markdownRendererUlBegin
\markdownRendererUlItem Use Question time as the reference point for relative dates and weekday names.\markdownRendererUlItemEnd 
\markdownRendererUlItem If the question mentions a weekday such as Monday/Tuesday/Wednesday/Thursday/Friday/Saturday/Sunday without an explicit calendar date, choose the most recent matching weekday at or before Question time, not an arbitrary weekday from the same week.\markdownRendererUlItemEnd 
\markdownRendererUlItem If the weekday in the question is the same weekday as Question time and the question describes an event that could already have happened, use Question time's date.\markdownRendererUlItemEnd 
\markdownRendererUlItem If the question says "last", "yesterday", "earlier", "this morning", "tonight", or similar relative wording, resolve it strictly relative to Question time.\markdownRendererUlItemEnd 
\markdownRendererUlItem If the question names a weekday night and the evidence may continue after midnight, anchor the window to the named weekday and use evening/night hours; do not shift the whole window to the next day.\markdownRendererUlItemEnd 
\markdownRendererUlItem If you cannot confidently resolve a weekday/date from Question time, set confidence below 0.6 rather than inventing a high-confidence date.\markdownRendererUlItemEnd 
\markdownRendererUlEnd \markdownRendererUlItemEnd 
\markdownRendererUlItem \markdownRendererStrongEmphasis{Vague-time normalization.} Normalize vague times with these ranges; do not expand morning/evening into the whole day:\markdownRendererInterblockSeparator
{}\markdownRendererUlBegin
\markdownRendererUlItem pre-dawn: 00:00:00-05:59:00\markdownRendererUlItemEnd 
\markdownRendererUlItem early morning: 06:00:00-09:59:00\markdownRendererUlItemEnd 
\markdownRendererUlItem morning: 05:00:00-11:59:00\markdownRendererUlItemEnd 
\markdownRendererUlItem noon: 11:00:00-13:59:00\markdownRendererUlItemEnd 
\markdownRendererUlItem afternoon: 13:00:00-17:59:00\markdownRendererUlItemEnd 
\markdownRendererUlItem dusk: 17:00:00-19:59:00\markdownRendererUlItemEnd 
\markdownRendererUlItem evening/night: 18:00:00-23:59:00\markdownRendererUlItemEnd 
\markdownRendererUlEnd \markdownRendererUlItemEnd 
\markdownRendererUlItem \markdownRendererStrongEmphasis{Coverage diversification.} Enable coverage when the question requires complete coverage across multiple people, devices, places, sources, or viewpoints, such as comparing parties, explaining each one separately, all related people/device states, who was present, or what each source observed. Do not enable coverage for a single subject's cause, state, or fact query.\markdownRendererUlItemEnd 
\markdownRendererUlItem \markdownRendererStrongEmphasis{Timeline mode.} Enable \markdownRendererCodeSpan{timeline\markdownRendererUnderscore{}mode} when the answer needs multiple ordered steps, such as a full sequence, route/itinerary, key milestones, phase changes, repeated attempts, multiple rounds, a morning/day reconstruction, or the process from one time to another. Do not enable it for a single-point reason or single fact query.\markdownRendererUlItemEnd 
\markdownRendererUlItem \markdownRendererStrongEmphasis{Temporal-neighbor decision table.}\markdownRendererInterblockSeparator
{}\markdownRendererUlBegin
\markdownRendererUlItem Enable when the question explicitly asks for a full sequence, timeline, total occurrences, each occurrence, a continuous period, a before/after progression, dialogue followed by actions, or a multi-step activity where adjacent snippets around hit events are needed in context.\markdownRendererUlItemEnd 
\markdownRendererUlItem Disable for ordinary why/reason/state/existence/how questions. Do not enable merely because a question is causal; multi-subject coverage also does not imply temporal\markdownRendererUnderscore{}neighbors.\markdownRendererUlItemEnd 
\markdownRendererUlItem radius defaults to 3; use 1-2 only for very short point-context questions.\markdownRendererUlItemEnd 
\markdownRendererUlEnd \markdownRendererUlItemEnd 
\markdownRendererUlItem \markdownRendererStrongEmphasis{Seed time-window quota decision table.}\markdownRendererInterblockSeparator
{}\markdownRendererUlBegin
\markdownRendererUlItem Default is 15.\markdownRendererUlItemEnd 
\markdownRendererUlItem Use 20 when coverage.enabled=true and timeline\markdownRendererUnderscore{}mode.enabled=false.\markdownRendererUlItemEnd 
\markdownRendererUlItem Use 25 when timeline\markdownRendererUnderscore{}mode.enabled=true and time\markdownRendererUnderscore{}window.enabled=true only if the time window is narrower than a full day.\markdownRendererUlItemEnd 
\markdownRendererUlItem Use 25 for explicit claim-verification questions with a date/time anchor, especially when the question asks whether a statement was true, accurate, supported by evidence, contradicted by records, or what actually happened. These questions need same-window corroborating and contradicting events, not just semantically similar older events.\markdownRendererUlItemEnd 
\markdownRendererUlItem Use 25 for narrow activity/session windows when the question asks what happened during a conversation, dinner, movie night, board game session, yoga session, appointment, commute, clinic visit, or practice; the answer often depends on several adjacent snippets inside that session.\markdownRendererUlItemEnd 
\markdownRendererUlItem For full-day/date-only windows such as a named weekday, an explicit calendar date, "that day", or 00:00:00-23:59:59, do not let the time window crowd out semantic/BM25 seeds:\markdownRendererInterblockSeparator
{}\markdownRendererUlBegin
\markdownRendererUlItem Use 15 for full-day timeline, full-day reconstruction, complete sequence, throughout-the-day progression, or multi-person/multi-device coverage.\markdownRendererUlItemEnd 
\markdownRendererUlItem Use 10-12 for full-day questions about a specific subject, object, appointment, decision, purchase, message, or incident.\markdownRendererUlItemEnd 
\markdownRendererUlItem Use 20 only when the question explicitly asks for all events across the whole day and lexical anchors are weak.\markdownRendererUlItemEnd 
\markdownRendererUlEnd \markdownRendererUlItemEnd 
\markdownRendererUlItem Do not go below 15 unless there is no time\markdownRendererUnderscore{}window.\markdownRendererUlItemEnd 
\markdownRendererUlItem Never use a full-day quota merely because a calendar date is present. First decide whether the question has a narrower activity/session anchor; if it does, use the narrower window and its normal quota.\markdownRendererUlItemEnd 
\markdownRendererUlEnd \markdownRendererUlItemEnd 
\markdownRendererUlItem \markdownRendererStrongEmphasis{Seed temporal-neighbor keep decision table.}\markdownRendererInterblockSeparator
{}\markdownRendererUlBegin
\markdownRendererUlItem Default is 24.\markdownRendererUlItemEnd 
\markdownRendererUlItem When temporal\markdownRendererUnderscore{}neighbors.enabled=true, still default to 24: first keep 24 original seeds, then use the remaining slots for adjacent events.\markdownRendererUlItemEnd 
\markdownRendererUlItem Do not go below 24 unless the question clearly requires continuous adjacent-snippet coverage and original semantic recall is not important.\markdownRendererUlItemEnd 
\markdownRendererUlEnd \markdownRendererUlItemEnd 
\markdownRendererUlItem \markdownRendererStrongEmphasis{Confidence.} Control confidence means confidence in the enabled decision: clear enable -> \markdownRendererCodeSpan{enabled=true} and 0.7-1.0; clear disable -> \markdownRendererCodeSpan{enabled=false} and 0.7-1.0; ambiguous -> 0.4-0.5.\markdownRendererUlItemEnd 
\markdownRendererUlItem \markdownRendererStrongEmphasis{Semantic queries.} \markdownRendererCodeSpan{semantic\markdownRendererUnderscore{}queries} should cover different retrieval intents of the question without adding unknown facts. Prefer 2-3 queries when the question asks for multi-hop, multi-detail, multi-source, timeline, comparison, or checklist-like evidence.\markdownRendererUlItemEnd 
\markdownRendererUlItem \markdownRendererStrongEmphasis{Minimum query coverage.} Provide at least one semantic query unless the original question is already very short and cannot be usefully rewritten.\markdownRendererUlItemEnd 
\markdownRendererUlItem \markdownRendererStrongEmphasis{Keywords.} \markdownRendererCodeSpan{keywords} must be short terms extracted from the original question for lexical matching within time windows. Prefer people, places, devices, objects, actions, numbers, colors, game/activity words, health readings, and state words. Do not output long phrases, full sentences, generic words, or negated/intensified compound phrases.\markdownRendererUlItemEnd 
\markdownRendererUlItem \markdownRendererStrongEmphasis{Entities.} \markdownRendererCodeSpan{entities} must be people, places, devices, or objects extracted from the original question. Do not leave it empty when such entities exist.\markdownRendererUlItemEnd 
\markdownRendererUlItem \markdownRendererStrongEmphasis{Output format.} Output JSON only, no markdown.\markdownRendererUlItemEnd 
\markdownRendererUlEnd \markdownRendererInterblockSeparator
{}\markdownRendererStrongEmphasis{Output exactly one plan object using this schema:}\markdownRendererInterblockSeparator
{}\markdownRendererInputFencedCode{prompt_tex/a45411f12ac1834103759202fca48065.verbatim}{json}\markdownRendererDocumentEnd\relax
\end{promptlisting}

\paragraph{Round Evidence Controller Prompt.}
This prompt asks the controller to judge accumulated evidence and decide whether another retrieval round is needed.
\begin{promptlisting}{Round Evidence Controller Prompt}
\markdownRendererDocumentBegin
\markdownRendererHeadingThree{System}\markdownRendererInterblockSeparator
{}You are an evidence-aware memory retrieval controller and planner.\markdownRendererInterblockSeparator
{}Judge accumulated evidence, select direct evidence IDs, and when another round is needed emit one executable structured retrieval plan.\markdownRendererInterblockSeparator
{}Do not answer the question. Return strict JSON.\markdownRendererInterblockSeparator
{}\markdownRendererHeadingThree{User}\markdownRendererInterblockSeparator
{}\markdownRendererStrongEmphasis{Original question:} \markdownRendererLeftBrace{}question\markdownRendererRightBrace{}\markdownRendererInterblockSeparator
{}\markdownRendererStrongEmphasis{Current search query:} \markdownRendererLeftBrace{}current\markdownRendererUnderscore{}query\markdownRendererRightBrace{}\markdownRendererInterblockSeparator
{}\markdownRendererStrongEmphasis{Frozen question time:} \markdownRendererLeftBrace{}query\markdownRendererUnderscore{}time\markdownRendererRightBrace{}\markdownRendererInterblockSeparator
{}\markdownRendererStrongEmphasis{Current executed round plan:}\linebreak
{}\markdownRendererLeftBrace{}current\markdownRendererUnderscore{}plan\markdownRendererRightBrace{}\markdownRendererInterblockSeparator
{}\markdownRendererStrongEmphasis{Queries already searched:} \markdownRendererLeftBrace{}searched\markdownRendererUnderscore{}queries\markdownRendererRightBrace{}\markdownRendererInterblockSeparator
{}\markdownRendererStrongEmphasis{Executed plan history:}\linebreak
{}\markdownRendererLeftBrace{}plan\markdownRendererUnderscore{}history\markdownRendererRightBrace{}\markdownRendererInterblockSeparator
{}\markdownRendererStrongEmphasis{Completed round:} \markdownRendererLeftBrace{}current\markdownRendererUnderscore{}round\markdownRendererRightBrace{}; \markdownRendererStrongEmphasis{minimum rounds before stopping:} \markdownRendererLeftBrace{}minimum\markdownRendererUnderscore{}rounds\markdownRendererRightBrace{}\markdownRendererInterblockSeparator
{}\markdownRendererStrongEmphasis{This is the final allowed round:} \markdownRendererLeftBrace{}final\markdownRendererUnderscore{}round\markdownRendererRightBrace{}\markdownRendererInterblockSeparator
{}\markdownRendererStrongEmphasis{Accumulated evidence candidates:}\markdownRendererInterblockSeparator
{}\markdownRendererInputFencedCode{prompt_tex/242b8c8aa0c64ac7349fa4e0a545458e.verbatim}{text}\markdownRendererInterblockSeparator
{}Set \markdownRendererCodeSpan{sufficient=true} only when every material facet of the question is directly supported by the retrieved evidence.\markdownRendererInterblockSeparator
{}Do not stop merely because the evidence suggests a plausible answer.\markdownRendererInterblockSeparator
{}Treat required people, events, time constraints, causal links, comparisons, and cross-device facts as separate facets when they matter.\markdownRendererInterblockSeparator
{}For every facet, cite one or more candidate IDs. A fused summary is valid evidence only when it directly contains the needed fact.\markdownRendererInterblockSeparator
{}Before round \markdownRendererCodeSpan{\markdownRendererLeftBrace{}minimum\markdownRendererUnderscore{}rounds\markdownRendererRightBrace{}}, \markdownRendererCodeSpan{sufficient} must remain false even when coverage appears complete.\markdownRendererInterblockSeparator
{}Emit a focused verification \markdownRendererCodeSpan{next\markdownRendererUnderscore{}plan} that seeks independent direct evidence for the covered facets.\markdownRendererInterblockSeparator
{}This prevents a single retrieval pass from authorizing its own early stop.\markdownRendererInterblockSeparator
{}If any required facet lacks direct evidence, set \markdownRendererCodeSpan{sufficient=false}, list it in \markdownRendererCodeSpan{missing\markdownRendererUnderscore{}facets}, and emit one \markdownRendererCodeSpan{next\markdownRendererUnderscore{}plan} targeting the most important missing facts.\markdownRendererInterblockSeparator
{}Its focus and rewrites must preserve all relevant people, dates, time ranges, devices, answer shape, and other constraints from the original question.\markdownRendererInterblockSeparator
{}Do not repeat a plan already present in the executed plan history.\markdownRendererInterblockSeparator
{}On the final allowed round, do not claim \markdownRendererCodeSpan{sufficient} unless all facets are actually covered; still return the best available selection when coverage is incomplete.\markdownRendererInterblockSeparator
{}\markdownRendererCodeSpan{selected\markdownRendererUnderscore{}ids} is an ordered priority shortlist and does not need to contain \markdownRendererCodeSpan{\markdownRendererLeftBrace{}top\markdownRendererUnderscore{}k\markdownRendererRightBrace{}} IDs.\markdownRendererInterblockSeparator
{}The retrieval system will fill any remaining slots from the accumulated candidate pool.\markdownRendererInterblockSeparator
{}When another round is allowed and \markdownRendererCodeSpan{sufficient=false}, \markdownRendererCodeSpan{next\markdownRendererUnderscore{}plan} must be a plan object following the rules and schema below. When \markdownRendererCodeSpan{sufficient=true} or this is the final round, \markdownRendererCodeSpan{next\markdownRendererUnderscore{}plan} must be null.\markdownRendererInterblockSeparator
{}[The same planning-field mapping and complete Query Planner decision rules shown in the initial-round call are inserted here verbatim.]\markdownRendererInterblockSeparator
{}\markdownRendererStrongEmphasis{next\markdownRendererUnderscore{}plan schema:}\markdownRendererInterblockSeparator
{}[The same round-plan JSON schema shown in the initial-round call is inserted here verbatim.]\markdownRendererInterblockSeparator
{}\markdownRendererStrongEmphasis{Output exactly:}\markdownRendererInterblockSeparator
{}\markdownRendererInputFencedCode{prompt_tex/aacbd31f13f286b1f8428756bf141c86.verbatim}{json}\markdownRendererDocumentEnd\relax
\end{promptlisting}

\subsection{Evaluation Prompts}

\subsubsection{Reader}

\paragraph{Full-context system prompt.}
This prompt tells the reader how to answer when the entire event stream is available as context.
\begin{promptbox}{Full-Context System Prompt}
\markdownRendererDocumentBegin
You are a household smart assistant. The following is an event stream recorded by smart devices in the home, including cameras, watches, speakers, and sensors.\linebreak
{}Each event record is formatted as: [time] [device] [location] event description.\linebreak
{}Answer the user's question accurately and completely based on these event records.\linebreak
{}When useful for answering, include relevant concrete details from the records, especially times, dates, numbers, device/app names, locations, people involved, and actions taken.\linebreak
{}Use only information from the event records. Do not fabricate or infer unsupported content.\linebreak
{}If the event records are insufficient for a complete answer, answer as well as possible based on the available information.\markdownRendererDocumentEnd\relax
\end{promptbox}

\paragraph{Retrieved-context system prompt.}
This prompt tells the reader how to answer when only retrieved evidence is available.
\begin{promptbox}{Retrieved-Context System Prompt}
\markdownRendererDocumentBegin
You are a household smart assistant. The following event records may be relevant to the user's question and were retrieved from smart devices in the home.\linebreak
{}Each event record is formatted as: [time] [device] [location] event description.\linebreak
{}Answer the user's question accurately and completely based on these event records.\linebreak
{}When useful for answering, include relevant concrete details from the records, especially times, dates, numbers, device/app names, locations, people involved, and actions taken.\linebreak
{}Use only information from the event records. Do not fabricate or infer unsupported content.\linebreak
{}If the provided information is insufficient for a complete answer, answer as well as possible based on the available information.\markdownRendererDocumentEnd\relax
\end{promptbox}

\paragraph{Shared reader user prompt.}
This prompt provides the retrieved context, questioner identity, question time, and question to the shared reader.
\begin{promptbox}{Shared Reader User Prompt}
\markdownRendererDocumentBegin
Event Records:\markdownRendererInterblockSeparator
{}\markdownRendererInputFencedCode{prompt_tex/5ee41b508fb41980a949cf968e06047d.verbatim}{text}\markdownRendererInterblockSeparator
{}Questioner: \markdownRendererLeftBrace{}question\markdownRendererUnderscore{}user\markdownRendererRightBrace{}\linebreak
{}Question Time: \markdownRendererLeftBrace{}question\markdownRendererUnderscore{}time\markdownRendererRightBrace{}\markdownRendererInterblockSeparator
{}Question: \markdownRendererLeftBrace{}question\markdownRendererRightBrace{}\markdownRendererDocumentEnd\relax
\end{promptbox}

\subsubsection{LLM-as-judge}

\paragraph{Judge system prompt.}
This prompt instructs the judge to evaluate checklist coverage and output JSON only.
\begin{promptbox}{Judge System Prompt}
\markdownRendererDocumentBegin
You are a rigorous evaluation expert responsible for judging answer checklist coverage. Always output JSON.\markdownRendererDocumentEnd\relax
\end{promptbox}

\paragraph{Judge user prompt.}
This prompt supplies the question, system answer, and checklist items to the judge.
\begin{promptbox}{Judge User Prompt}
\markdownRendererDocumentBegin
You are an evaluation expert. Determine whether the system answer covers each key information point.\markdownRendererInterblockSeparator
{}Question: \markdownRendererLeftBrace{}question\markdownRendererRightBrace{}\markdownRendererInterblockSeparator
{}System Answer:\linebreak
{}\markdownRendererLeftBrace{}system\markdownRendererUnderscore{}answer\markdownRendererRightBrace{}\markdownRendererInterblockSeparator
{}Judge each item below for whether it is covered by the system answer (directly stated or clearly implied both count as covered).\linebreak
{}Evaluation criteria: be tolerant of wording differences, but strict about factual correctness. If the system answer contains the core meaning of the item, it counts as covered even with different wording.\markdownRendererInterblockSeparator
{}Checklist Items:\linebreak
{}\markdownRendererLeftBrace{}checklist\markdownRendererUnderscore{}text\markdownRendererRightBrace{}\markdownRendererInterblockSeparator
{}Output strictly as a JSON array and do not output anything else:\markdownRendererInterblockSeparator
{}\markdownRendererInputFencedCode{prompt_tex/f7c1bc5dcf754612c4c0e624a5127c51.verbatim}{json}\markdownRendererDocumentEnd\relax
\end{promptbox}

\end{document}